\documentclass{article}

\usepackage[numbers]{natbib}
\usepackage{PRIMEarxiv}

\usepackage[hidelinks]{hyperref}
\usepackage{multirow}
\usepackage[utf8]{inputenc}
\usepackage[ruled, lined, longend, linesnumbered]{algorithm2e}
\usepackage{amsmath,multicol}
\usepackage{subcaption,textcomp,comment}
\usepackage{url}

\usepackage{rotating}

\usepackage{booktabs}
\usepackage{multirow}
\usepackage{enumitem}
\usepackage{amssymb}
\usepackage{graphicx}

\usepackage{array}
\newcolumntype{L}[1]{>{\raggedright\let\newline\\\arraybackslash\hspace{0pt}}m{#1}}
\newcolumntype{C}[1]{>{\centering\let\newline\\\arraybackslash\hspace{0pt}}m{#1}}
\newcolumntype{R}[1]{>{\raggedleft\let\newline\\\arraybackslash\hspace{0pt}}m{#1}}

\newcommand{\update}[1]{#1}

\DeclareSymbolFont{extraup}{U}{zavm}{m}{n}
\DeclareMathSymbol{\varheart}{\mathalpha}{extraup}{86}
\DeclareMathSymbol{\vardiamond}{\mathalpha}{extraup}{87}

\usepackage{tikz}
\usetikzlibrary{external}
\usepackage[edges]{forest}
\definecolor{hidden-draw}{RGB}{20,68,106}
\definecolor{hidden-pink}{RGB}{255,245,247}
\definecolor{hidden-red}{RGB}{180,0,0}

	\title{A Survey of Resource-efficient\\ LLM and Multimodal Foundation Models}
	
	\author{
		\bf Mengwei Xu$^{\clubsuit}$,  Wangsong Yin$^\vardiamond$, Dongqi Cai$^\clubsuit$, Rongjie Yi$^\clubsuit$, Daliang Xu$^\vardiamond$, Qipeng Wang$^\vardiamond$, Bingyang Wu$^\vardiamond$,\\
		\bf  Yihao Zhao$^\vardiamond$, Chen Yang$^\clubsuit$, Shihe Wang$^\clubsuit$, Qiyang Zhang$^\clubsuit$, Zhenyan Lu$^\clubsuit$, Li Zhang$^\clubsuit$, \\
		\bf Shangguang Wang$^\clubsuit$, Yuanchun Li$^\varheart$, Yunxin Liu$^\varheart$, Xin Jin$^\vardiamond$, Xuanzhe Liu$^\vardiamond$\\ \\
		$\clubsuit$ Beijing University of Posts and Telecommunications (BUPT)\\ $\vardiamond$ Peking University (PKU)\\
		$\varheart$ Tsinghua University (THU)\\ \\
		Contact: \texttt{mwx@bupt.edu.cn}\\
		Website: \url{https:github.com/UbiquitousLearning/Efficient_Foundation_Model_Survey}
	}

\begin{document}
	\maketitle
	
	\begin{abstract}
Large foundation models, including large language models (LLMs), vision transformers (ViTs), diffusion, and LLM-based multimodal models, are revolutionizing the entire machine learning lifecycle, from training to deployment. However, the substantial advancements in versatility and performance these models offer come at a significant cost in terms of hardware resources. To support the growth of these large models in a scalable and environmentally sustainable way, there has been a considerable focus on developing resource-efficient strategies. This survey delves into the critical importance of such research, examining both algorithmic and systemic aspects. It offers a comprehensive analysis and valuable insights gleaned from existing literature, encompassing a broad array of topics from cutting-edge model architectures and training/serving algorithms to practical system designs and implementations. The goal of this survey is to provide an overarching understanding of how current approaches are tackling the resource challenges posed by large foundation models and to potentially inspire future breakthroughs in this field.

\end{abstract}
	
	
	 \section{INTRODUCTION}\label{sec:intro}

In the rapidly evolving field of artificial intelligence (AI), a paradigm shift is underway.
We are witnessing the transition from specialized, fragmented deep learning models to versatile, one-size-fits-all foundation models.
These advanced AI systems are capable of operating in an open-world context, interacting with open vocabularies and image pixels for unseen AI tasks, i.e., zero-shot abilities.
They are exemplified by
(1) Large Language Models (LLMs) such as GPTs~\cite{brown2020language} that can ingest almost every NLP task in the form as a prompt;
(2) Vision Transformers Models (ViTs) such as Masked Autoencoder~\cite{he2022masked} that can handle various downstream vision tasks;
(3) Latent Diffusion Models (LDMs) such as Stable Diffusion~\cite{rombach2022high} that generate high-quality images with arbitrary text-based prompts;
(4) Multimodal models such as CLIP~\cite{radford2021learning} and ImageBind~\cite{girdhar2023imagebind} that map different modal data into the same latent space and are widely used as backbone for cross-modality tasks like image retrieval/search and visual-question answering.
Such flexibility and generality marks a significant departure from the earlier era of AI, setting a new standard for how AI interfaces with the world.

The success of these foundation models is deeply rooted in their scalability: unlike their predecessors, these models' accuracy and generalization ability can continuously expand with more data or parameters, without altering the underlying simple algorithms and architectures.
An impressive evidence is the scaling law~\cite{kaplan2020scaling}: it describes how the performance of transformer-based models can predictably improve with more model size and data volume;
until today, the scaling law stands still.
This scalability is not just a matter of model size; it extends to their ability to tackle increasingly complex tasks, making them a cornerstone in the journey towards artificial general intelligence (AGI).

\begin{figure*}
	\centering
	\includegraphics[width=0.8\textwidth]{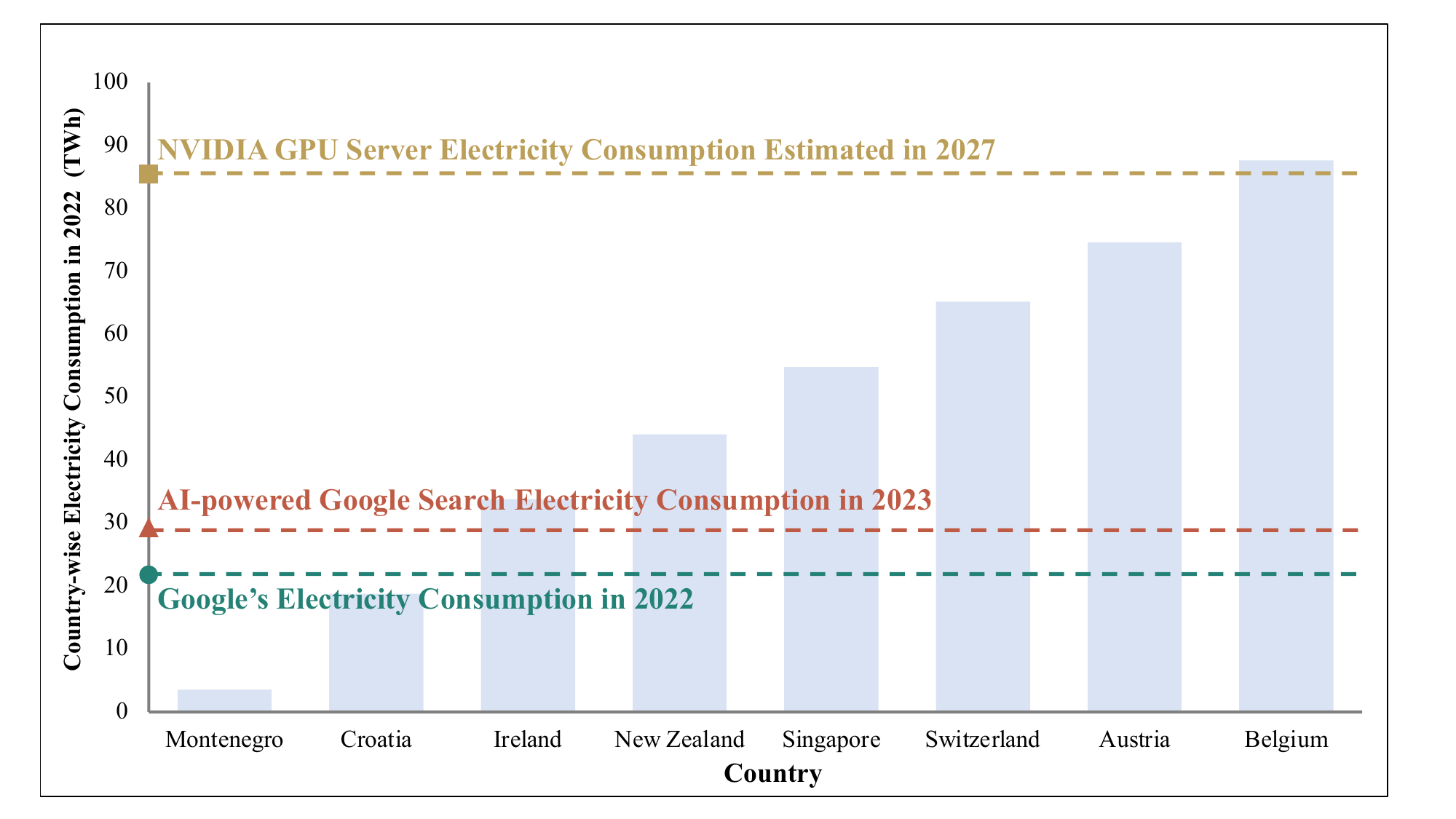}
	\caption{
		The electricity consumption comparison between countries and AI. Data source: \cite{de2023growing}.
		}
	\label{fig:intro-AI-elec}
\end{figure*}

However, the scalability comes at a cost of huge resource demand.
Foundation models, by their very nature, are resource-hungry for training and deployment.
These resources encompass not only the computing processors like GPUs and TPUs, but also the memory, energy, and network bandwidth.
For example, the pre-training of LLaMa-2-70B takes $1.7\times$ millions of GPU hours and consumes $2.5\times10^{12}$ Joules of energy.
The estimated total emissions were 291 tons of CO2 equivalent.
Beyond training, the data processing, experimentation, and inference stages consume comparable or even more electricity according to Meta AI~\cite{wu2022sustainable}.
A recent analysis~\cite{de2023growing} reveals that, to satisfy the continuation of the current trends in AI capacity and adoption, NVIDIA needs to ship 1.5 million AI server units per year by 2027.
These servers, running at full capacity, would consume at least 85.4 terawatt-hours of electricity annuall -- more than what many countries like New Zealand and Austria use in a whole year, as illustrated in Figure~\ref{fig:intro-AI-elec}.
Since foundation models proceed growth in size and complexity, their resource requirements escalate, often exponentially, posing a significant challenge in their development and deployment.

The huge resource footprint of large foundation model also hinders its democratization.
Till the end of 2023, there are only a few major players capable of training and deploying the state-of-the-art foundation models, who thereby have powerful control over the public and can potentially manipulate them in a way they prefer.
The models are served on clouds instead of devices as many lightweight DNNs do~\cite{xu2019first, zhang2023comprehensive}; it makes data privacy preservation almost impossible.
Though recently, smartphone vendors have been boasting about running large foundation models locally and some pioneering engines are developed for on-device LLMs~\cite{llamacpp,mnn-llm,mllm}, the models demonstrated are limited to relatively small scale (e.g., $<$10B)~\cite{slmsurvey} and have not yet seen real-world deployment.

Thereby, a significant amount of research has been dedicated to enhance the efficiency of these foundation models.
These efforts span a wide range of approaches, from optimizing algorithms to system-level innovations, focusing on reducing the resource footprint of these models without compromising their performance.
This survey aims to delve into these research efforts, exploring the diverse strategies employed to make foundation models more resource-efficient.
We will examine advancements in algorithmic efficiency, system optimizations, data management techniques, and the development of novel architectures that are less resource-intensive.
The survey also spans from clouds to edge and devices, where the large foundation models gain dramatic attentions as well.
Through this exploration, we aim to provide a comprehensive understanding of the current state and future directions of resource-efficient algorithms and systems in the realm of foundation models.

\tikzstyle{subsubsection}=[font=\small, align=center, text width=13em]

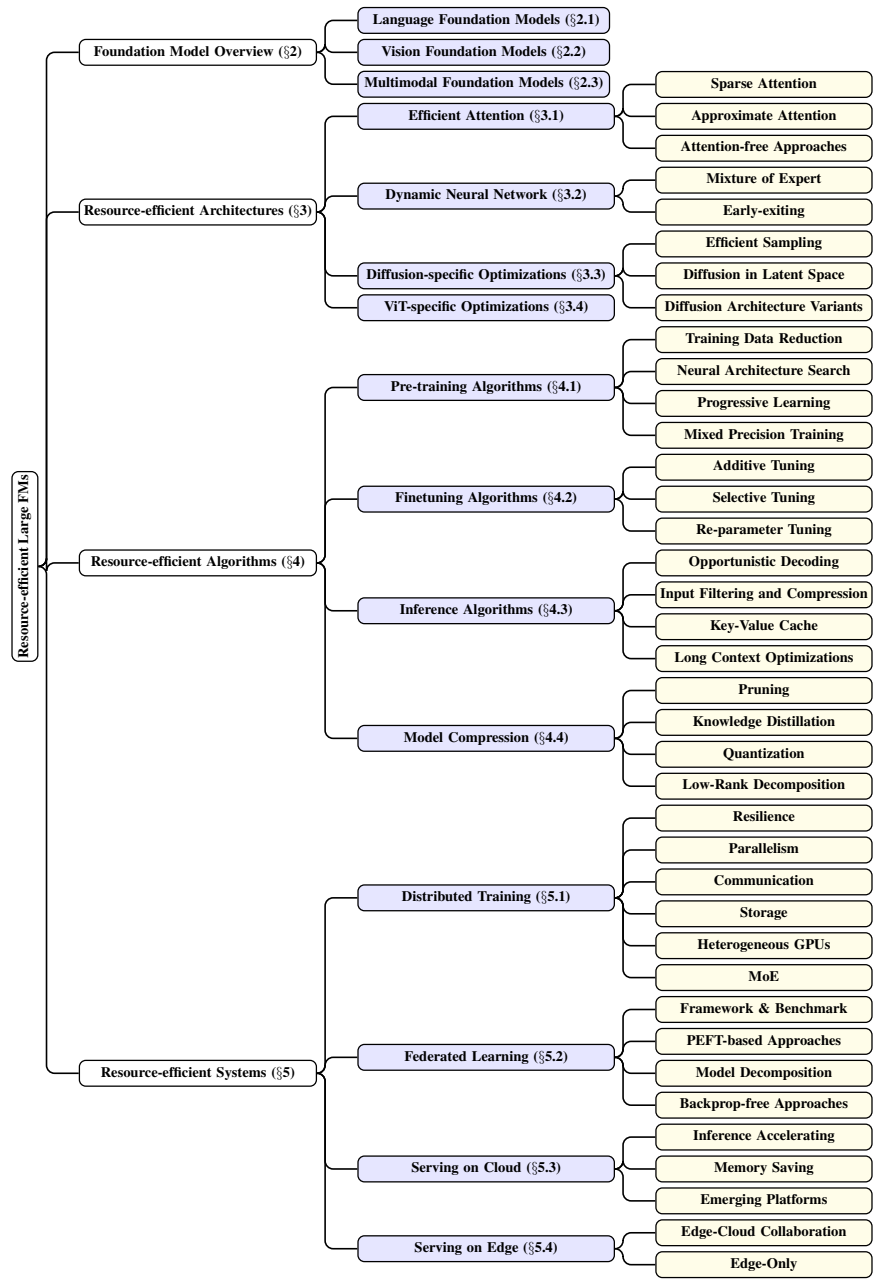
\begin{figure*}[t]
    \centering
    \resizebox{0.7\textwidth}{!}{
        \begin{forest}
            forked edges,
            for tree={
                grow=east,
                reversed=true,
                anchor=base west,
                parent anchor=east,
                child anchor=west,
                base=center,
                font=\large,
                rectangle,
                draw=black,
                rounded corners,
                align=left,
                text centered,
                minimum width=4em,
                edge+={black, line width=1pt, rounded corners=5pt},
                l sep=25pt,
                s sep=3pt,
                inner xsep=2pt,
                inner ysep=3pt,
                line width=0.8pt,
                ver/.style={rotate=90, child anchor=north, parent anchor=south, anchor=center},
            },
            where level=1{text width=12em,font=\small,}{},
            where level=2{text width=12em,font=\small,}{},
            [
              \textbf{Resource-efficient Large FMs}, font=\small, align=center, ver
              [
                \textbf{Foundation Model Overview ($\S$\ref{sec:bkgnd})}, text width=14.3em
                [
                  \textbf{ Language Foundation Models ($\S$\ref{subsec:bkgnd-LLM})}, fill=blue!10, text width=15.2em
                ]
                [
                  \textbf{Vision Foundation Models ($\S$\ref{subsec:bkgnd-vit})}, fill=blue!10, text width=15.2em
                ]
                [ \textbf{Multimodal Foundation Models ($\S$\ref{subsec:bkgnd-mfm})}, fill=blue!10, text width=15.2em
                ]
              ]
              [
                \textbf{Resource-eﬀicient Architectures ($\S$\ref{sec:architecture})}, text width=14.3em
                [
                  \textbf{Eﬀicient Attention ($\S$\ref{subsec:efficient-attention})}, fill=blue!10, text width=15.5em
                  [
                    \textbf{Sparse Attention}, subsubsection, fill=yellow!10
                  ]
                  [
                    \textbf{Approximate Attention}, subsubsection, fill=yellow!10
                  ]
                  [
                    \textbf{Attention-free Approaches}, subsubsection, fill=yellow!10
                  ]
                ]
                [
                  \textbf{Dynamic Neural Network ($\S$\ref{subsec:dynamic-neural-network})}, fill=blue!10, text width=15.5em
                  [
                    \textbf{Mixture of Expert}, subsubsection, fill=yellow!10
                  ]
                  [
                    \textbf{Early-exiting}, subsubsection, fill=yellow!10
                  ]
                ]
                [ 
                  \textbf{Diffusion-specific Optimizations ($\S$\ref{subsec:diffusion-optim})}, fill=blue!10, text width=15.5em
                  [
                    \textbf{Efficient Sampling}, subsubsection, fill=yellow!10
                  ]
                  [
                    \textbf{Diffusion in Latent Space}, subsubsection, fill=yellow!10
                  ]
                  [
                    \textbf{Diffusion Architecture Variants}, subsubsection, fill=yellow!10
                  ]
                ]
                [ 
                  \textbf{ViT-specific Optimizations ($\S$\ref{subsec:ViT-specific Optimization})}, fill=blue!10, text width=15.5em
                ]
              ]
            [
                \textbf{Resource-efficient Algorithms ($\S$\ref{sec:algorithms})}, text width=14.3em
                [
                  \textbf{Pre-training Algorithms ($\S$\ref{subsec:Pre-training-algorithms})}, fill=blue!10, text width=15.5em
                            [
                            \textbf{Training Data Reduction}, subsubsection, fill=yellow!10
                            ]
                            [
                            \textbf{Neural Architecture Search}, subsubsection, fill=yellow!10
                            ]
                            [
                            \textbf{Progressive Learning}, subsubsection, fill=yellow!10
                            ]
                            [
                            \textbf{Mixed Precision Training}, subsubsection, fill=yellow!10
                            ]
                ]
                [
                  \textbf{Finetuning Algorithms ($\S$\ref{subsec:finetuning-alg})}, fill=blue!10, text width=15.5em
                    [
                    \textbf{Additive Tuning}, subsubsection, fill=yellow!10
                    ]
                    [
                    \textbf{Selective Tuning}, subsubsection, fill=yellow!10
                    ]
                    [
                    \textbf{Re-parameter Tuning}, subsubsection, fill=yellow!10
                    ]
                ]
                [
					        \textbf{Inference Algorithms ($\S$\ref{subsec:inference-alg})}, fill=blue!10, text width=15.5em
                    [
                    \textbf{Opportunistic Decoding}, subsubsection, fill=yellow!10
                    ]
                    [
                    \textbf{Input Filtering and Compression}, subsubsection, fill=yellow!10
                    ]
                    [
                    \textbf{Key-Value Cache}, subsubsection, fill=yellow!10
                    ]
                    [
                    \textbf{Long Context Optimizations}, subsubsection, fill=yellow!10
                    ]
                ]
                [
                  \textbf{Model Compression ($\S$\ref{subsec:model-compression})}, fill=blue!10, text width=15.5em 
                    [
                    \textbf{Pruning}, subsubsection, fill=yellow!10
                    ]
                    [
                    \textbf{Knowledge Distillation}, subsubsection, fill=yellow!10
                    ]
                    [
                    \textbf{Quantization}, subsubsection, fill=yellow!10
                    ]
                    [
                    \textbf{Low-Rank Decomposition}, subsubsection, fill=yellow!10
                    ]
                ]
            ]
            [
              \textbf{Resource-efficient Systems ($\S$\ref{sec:systems})}, text width=14.3em
              [
                  \textbf{Distributed Training ($\S$\ref{subsec:distributed-training-systems})}, fill=blue!10, text width=15.5em
                            [
                            \textbf{Resilience}, subsubsection, fill=yellow!10
                            ]
                            [
                            \textbf{Parallelism}, subsubsection, fill=yellow!10
                            ]
                            [
                            \textbf{Communication}, subsubsection, fill=yellow!10
                            ]
                            [
                            \textbf{Storage}, subsubsection, fill=yellow!10
                            ]
                            [
                            \textbf{Heterogeneous GPUs}, subsubsection, fill=yellow!10
                            ]
                            [
                            \textbf{MoE}, subsubsection, fill=yellow!10
                            ]
              ]
              [
                \textbf{Federated Learning ($\S$\ref{subsec:federated-fine-tuning-systems})}, fill=blue!10, text width=15.5em
                [
                \textbf{Framework \& Benchmark}, subsubsection, fill=yellow!10
                ] 
                [
                \textbf{PEFT-based Approaches}, subsubsection, fill=yellow!10
                ]  
                [
                \textbf{Model Decomposition}, subsubsection, fill=yellow!10
                ]  
                [
                \textbf{Backprop-free Approaches}, subsubsection, fill=yellow!10
                ]               
              ]
              [
                \textbf{Serving on Cloud ($\S$\ref{subsec:cloud-serving-sys})}, fill=blue!10, text width=15.5em
                  [
                  \textbf{Inference Accelerating}, subsubsection, fill=yellow!10
                  ] 
                  [
                  \textbf{Memory Saving}, subsubsection, fill=yellow!10
                  ]  
                  [
                  \textbf{Emerging Platforms}, subsubsection, fill=yellow!10
                  ]  
              ]
              [
                \textbf{Serving on Edge ($\S$\ref{subsec:edge-serving-sys})}, fill=blue!10, text width=15.5em
                  [
                  \textbf{Edge-Cloud Collaboration}, subsubsection, fill=yellow!10
                  ]  
                  [
                  \textbf{Edge-Only}, subsubsection, fill=yellow!10
                  ]
              ]
            ]
          ]     
        \end{forest}
    }
    \caption{The organization of this survey.}
    \label{fig:overview}
\end{figure*}

\textbf{Scope and rationales.}
The scope of this survey is mainly defined by following aspects.
(i) We survey only algorithm and system innovations;
we exclude a huge body of work at hardware design,  which is equally important but has been already wrapped up well~\cite{koilia2024hardware,kim2023full}.
(ii) The definition of resource in this survey is limited to mainly physical ones, including computing, memory, storage, bandwidth, etc;
we exclude training data (labels) and privacy that can also be regarded as resources;
(iii) We mainly survey papers published on top-tier CS conferences, i.e., those included in CSRankings.
We also manually pick related and potentially high-impact papers from arXiv.
(iv) We mainly survey papers published after the year of 2020, since the innovation of AI is going fast with old knowledge and methods being overturned frequently.


\textbf{Organization.} Figure~\ref{fig:overview} illustrates the organization of this survey.

\textbf{Full open-source.} All materials of this survey are freely available at:\\ \\
\centerline{ \url{https:github.com/UbiquitousLearning/Efficient_Foundation_Model_Survey}}
	\section{FOUNDATION MODEL OVERVIEW}
\label{sec:bkgnd}

\begin{figure*}
	\centering
	\includegraphics[width=0.98\textwidth]{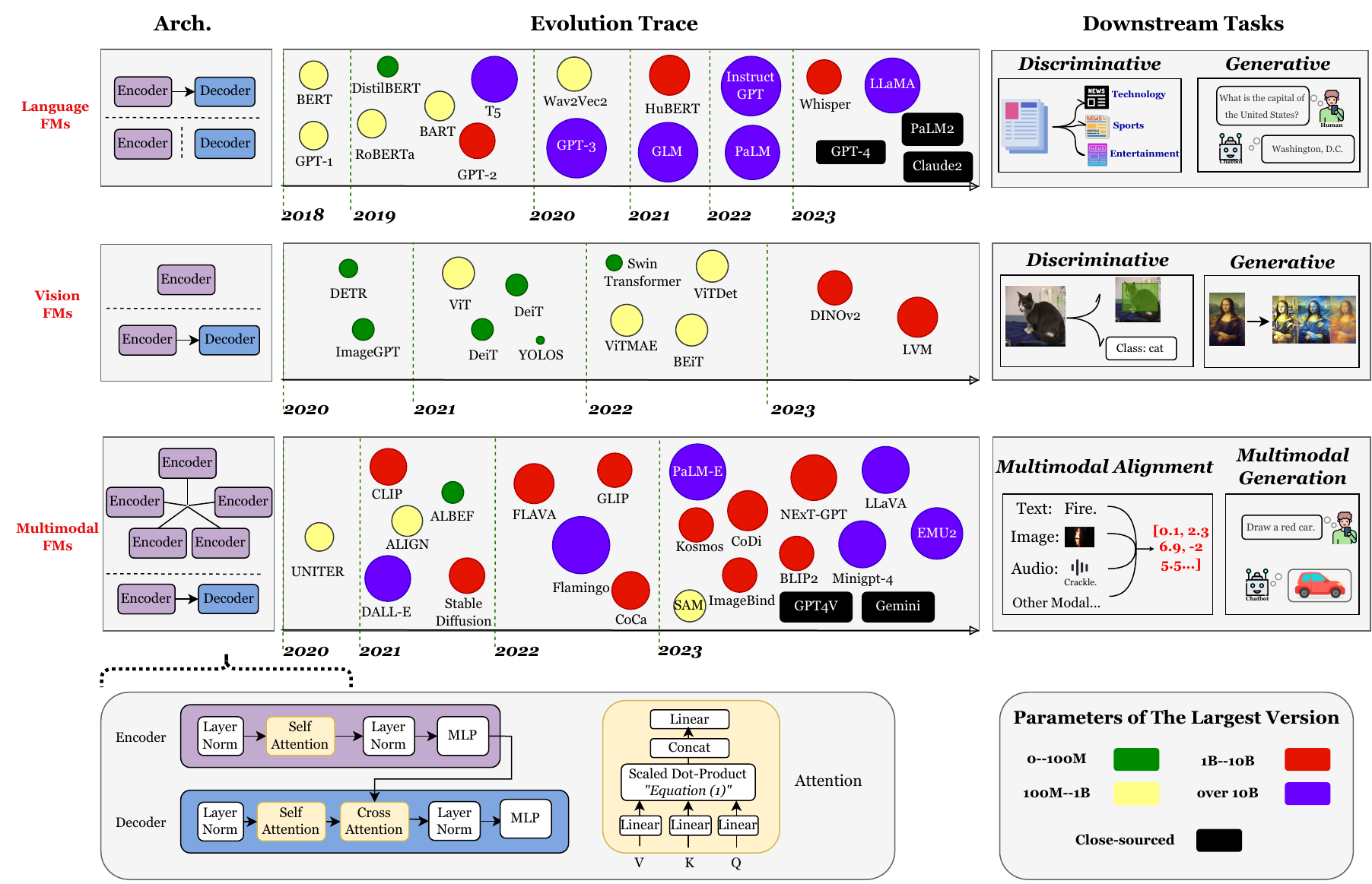}
	\caption{
    The evolutionary trace of foundation models. 
		}
	\label{fig:bkg-tree}
\end{figure*}


\subsection{Language Foundation Models}
\label{subsec:bkgnd-LLM}

This section includes a discussion of both text-based and speech-based language models, highlighting their key architecture and milestone models.


\subsubsection{Model Architectures}

\textbf{Transformer pipeline.}
Vaswani et al.~\cite{vaswani2017attention} introduced the attention-based Transformer architecture, a foundational element in the development of most Large FMs. As depicted in Figure~\ref{fig:bkg-tree}, the process initiates by converting input words into high-dimensional vectors through an embedding layer. During processing, attention mechanisms assign varying weights to different segments of these input vectors. Following attention, layer normalization is applied to the output, ensuring stabilization and standardization of the activations. Subsequently, each position-wise vector undergoes transformation through a feedforward network, introducing non-linearity and enabling the model to capture complex data patterns. Through multiple layers that incorporate these components, the Transformer learns hierarchical representations of the input data. In the final stage, the output from the last Transformer layer is directed into a linear layer, culminating in the final prediction.
We briefly outlines the key components of Large FMs as follows:

\textbf{Embedding.} 
Initially, the input word is transformed into a sequence of tokens by a tokenizer. Commonly used tokenizers, such as wordpiece and byte-pair encoding, are frequently employed in this process~\cite{toraman2023impact}. Following tokenization, a learned embedding layer converts these tokens into a sequence of vectors. In such sequences, the order of words is essential for meaning. To address this, position encoding is incorporated into the embeddings, infusing them with positional information. This addition is critical for capturing the sequential nature of the input, ensuring that the model accurately interprets word order and context.

\textbf{Attention.}
Attention mechanisms play a crucial role in capturing the relationships between words in a sequence. The calculation of attention can be represented as:
\begin{equation}
    A(Q, K, V) = \text{softmax}\left(\frac{QK^T}{\sqrt{d_k}}\right) V
    \label{eq:attention}
\end{equation}
where $Q$, $K$, and $V$ represent the query, key, and value, respectively; each derived by multiplying the input vector with a distinct weight matrix, and $d_k$ denotes the dimension of these vectors. 
Self-attention, a specific form of attention where queries, keys, and values all originate from the same input sequence, enables the model to focus on different segments of the input for each position. In contrast, multi-head attention, a variation of self-attention, permits simultaneous attention to information from diverse representation subspaces at different positions. Other variants, such as sparse attention~\cite{beltagy2020longformer} and multi-query attention~\cite{shazeer2019fast}, are tailored for efficiency or various downstream tasks. 
These variants are further detailed in $\S$\ref{subsec:efficient-attention}, $\S$\ref{subsec:kv-cache} and $\S$\ref{subsec:recurrent-structure}.

\textbf{Encoder-decoder architecture.}
The standard Transformer architecture consists of two main components: an encoder and a decoder. Encoder processes the input sequence through self-attention mechanisms, allowing the model to assign varying weights to different segments of the input sequence based on their relative importance. This feature is crucial for discerning complex patterns and dependencies within the input data.
In contrast, the decoder is responsible for generating the output sequence. Decoder utilizes self-attention mechanisms to understand the relationships within the generated output so far. Additionally, the decoder incorporates cross-attention mechanisms, focusing on the input sequence to extract relevant information for each token in the output sequence. This part of the architecture is autoregressive, generating tokens sequentially. The production of each token depends on the tokens generated previously, unlike the parallel processing approach of the encoder.

\textbf{Auto-regressive decoding and KV cache.}
In auto-regressive decoding, the decoder function $F_{Decoder}$ infers a new token $x_{i+1}$ based on the input token sequence $X = \{x_{1}, x_{2}, ..., x_{i}\}$. Subsequently, $x_{i+1}$ is added to $X$ for the next inference step, constituting auto-regressive decoding. Key-Value (KV) cache~\cite{dao2022flashattention} enhances efficiency by storing intermediate states of the attention mechanism at each step. This approach prevents recalculation of tokens processed in earlier steps. While mitigating re-computation, KV cache introduces additional storage or memory overhead, detailed in $\S$\ref{subsec: llm-cost-analysis}.

\subsubsection{Representative Models and Downstream Tasks}
\begin{table}[]
  \centering
  \resizebox{\columnwidth}{!}{%
  \begin{tabular}{|c|c|c|c|ccc|c|}
  \hline
  \textbf{Year} & \textbf{Model Name} & \textbf{Model Arch.}           & \textbf{Oriented Tasks}                                                                               & \multicolumn{1}{c|}{\textbf{Parameters}} & \multicolumn{1}{c|}{\textbf{Pre-training Method}}                                                   & \multicolumn{1}{c|}{\textbf{Pre-training Datasets}}                                                                                 & \textbf{Testing Datasets}                                                                 \\ \hline
  2018          & BERT~\cite{devlin2018bert}                & Encoder-Only                   & \begin{tabular}[c]{@{}c@{}}Text-CLS, \\ Token-CLS, \\ Fill-Mask, QA,\\ Translation, etc.\end{tabular} & \multicolumn{1}{c|}{110-340M}            & \multicolumn{1}{c|}{Self-Supervised}                                                                & \multicolumn{1}{c|}{\begin{tabular}[c]{@{}c@{}}Bookscorpus, \\ Enlish Wikipedia\end{tabular}}                                       & \begin{tabular}[c]{@{}c@{}}GLUE, \\ SquAD v1.1/2.0, \\ SWAG,\\ IMDb\end{tabular}          \\ \hline
  2019          & DistilBERT~\cite{sanh2019distilbert}          & Encoder-Only                   & Same as BERT                                                                                          & \multicolumn{1}{c|}{66M}                 & \multicolumn{1}{c|}{\begin{tabular}[c]{@{}c@{}}Self-Supervised, \\ Distillation\end{tabular}}       & \multicolumn{1}{c|}{Same as BERT}                                                                                                   & \begin{tabular}[c]{@{}c@{}}GLUE, \\ SquAD, IMDb\end{tabular}                              \\ \hline
  2019          & RoBERTa~\cite{liu2019roberta}             & Encoder-Only                   & Same as BERT                                                                                          & \multicolumn{1}{c|}{125-355M}            & \multicolumn{1}{c|}{Self-Supervised}                                                                & \multicolumn{1}{c|}{\begin{tabular}[c]{@{}c@{}}Boookcorpus, \\ CC-news, \\ Openwebtext, \\ Stories\end{tabular}}                    & \begin{tabular}[c]{@{}c@{}}GLUE, \\ SQuAD, RACE\end{tabular}                              \\ \hline
  2019          & Sentence-BERT~\cite{reimers2019sentence}       &    Encoder-Only                            & Text Similarity                                                                                       & \multicolumn{1}{c|}{110M}                & \multicolumn{1}{c|}{Only Fine-tuning}                                                               & \multicolumn{1}{c|}{SNLI, Multi-Genre NLI}                                                                                          & STSb                                                                                      \\ \hline
  2019          & BART~\cite{lewis2019bart}                & Encoder-Decoder                & Same as BERT                                                                                          & \multicolumn{1}{c|}{140-400M}            & \multicolumn{1}{c|}{Self-Supervised}                                                                & \multicolumn{1}{c|}{\begin{tabular}[c]{@{}c@{}}Boookcorpus, CC-news, \\ Openwebtext, Stories\end{tabular}}                          & \begin{tabular}[c]{@{}c@{}}SQuAD, MNLI, \\ ELI5, XSum, \\ ConvAI2, CNN/DM\end{tabular}    \\ \hline
  2019          & T5~\cite{raffel2020exploring}                  & Encoder-Decoder                & Same as BERT                                                                                          & \multicolumn{1}{c|}{60M-11B}             & \multicolumn{1}{c|}{Self-Supervised}                                                                & \multicolumn{1}{c|}{\begin{tabular}[c]{@{}c@{}}Colossal Clean \\ Crawled Corpus\end{tabular}}                                       & \begin{tabular}[c]{@{}c@{}}GLUE, CNNDM, \\ SQuAD, SGLUE, \\ EnDe, EnFr, EnRO\end{tabular} \\ \hline
  2018          & GPT-1~\cite{radford2018improving}               & Decoder-Only                   & Same as BERT                                                                                          & \multicolumn{1}{c|}{117M}                & \multicolumn{1}{c|}{Self-Supervised}                                                                   & \multicolumn{1}{c|}{\begin{tabular}[c]{@{}c@{}}BooksCorpus,\\ English Wikipedia\end{tabular}}                                       & SQuAD, SNLI                                                                               \\ \hline
  2019          & GPT-2~\cite{radford2019language}               & Decoder-Only                   & Same as BERT                                                                                          & \multicolumn{1}{c|}{1.5B}                & \multicolumn{1}{c|}{Self-Supervised}                                                                   & \multicolumn{1}{c|}{WebText}                                                                                                        & \begin{tabular}[c]{@{}c@{}}SQuAD, CoQA,\\ WMT, \\ CNN/Daily Mail\end{tabular}             \\ \hline
  2020          & GPT-3~\cite{brown2020language}               & Decoder-Only                   & Same as BERT                                                                                          & \multicolumn{1}{c|}{175B}                & \multicolumn{1}{c|}{Unsupervised}                                                                   & \multicolumn{1}{c|}{\begin{tabular}[c]{@{}c@{}}Common Crawl, WebText2\\ Books1/2, Wikipedia\end{tabular}}                           & \begin{tabular}[c]{@{}c@{}}LAMBADA, CBT,\\ SuperGLUE\end{tabular}                         \\ \hline
  2021          & GLM~\cite{du2021glm}         & Decoder-Only                   & Same as BERT                                                                                          & \multicolumn{1}{c|}{110M-130B}                & \multicolumn{1}{c|}{Unsupervised}             & \multicolumn{1}{c|}{\begin{tabular}[c]{@{}c@{}}BooksCorpus, English\\ Wikipedia\end{tabular}}                           & SuperGLUE                       \\ \hline
  2022          & InsturctGPT~\cite{ouyang2022training}         & Decoder-Only                   & Same as BERT                                                                                          & \multicolumn{1}{c|}{175B}                & \multicolumn{1}{c|}{\begin{tabular}[c]{@{}c@{}}Unsupervised \\ RLHF\end{tabular}}             & \multicolumn{1}{c|}{\begin{tabular}[c]{@{}c@{}}Common Crawl, WebText2\\ Books1/2, Wikipedia\end{tabular}}                           & \begin{tabular}[c]{@{}c@{}}LAMBADA,\\ CBT, SuperGLUE\end{tabular}                         \\ \hline
  2022          & PaLM~\cite{chowdhery2023palm}         & Decoder-Only                   & Same as BERT                                                                                          & \multicolumn{1}{c|}{54B}                & \multicolumn{1}{c|}{\begin{tabular}[c]{@{}c@{}}Unsupervised\end{tabular}}             & \multicolumn{1}{c|}{\begin{tabular}[c]{@{}c@{}}Mixture of 780B Text\\ Source code\end{tabular}}                           & \begin{tabular}[c]{@{}c@{}}English NLP, BIG-bench\\ Reasoning, Code, etc.\end{tabular}                         \\ \hline
  2020          & wav2vec2~\cite{baevski2020wav2vec}            & Encoder-Decoder                & \begin{tabular}[c]{@{}c@{}}Auto Speech \\ Recognition\end{tabular}                                    & \multicolumn{1}{c|}{227-896M}            & \multicolumn{1}{c|}{Self-Supervised}                                                                & \multicolumn{1}{c|}{\begin{tabular}[c]{@{}c@{}}LibriSpeech,\\ Unlabeled Audio Data\end{tabular}}                                    & \begin{tabular}[c]{@{}c@{}}LibriSpeech,\\ TIMIT,\\ Common Voice\end{tabular}              \\ \hline
  2021          & HuBERT~\cite{hsu2021hubert}              & Encoder-Decoder                & \begin{tabular}[c]{@{}c@{}}Auto Speech \\ Recognition\end{tabular}                                    & \multicolumn{1}{c|}{281M-2.8B}           & \multicolumn{1}{c|}{Self-Supervised}                                                                & \multicolumn{1}{c|}{\begin{tabular}[c]{@{}c@{}}Libri-Light,\\ LibriSpeech\end{tabular}}                                             & \begin{tabular}[c]{@{}c@{}}LibriSpeech,\\ TIMIT\end{tabular}                              \\ \hline
  2023          & Whisper~\cite{radford2023robust}             & Encoder-Decoder                & \begin{tabular}[c]{@{}c@{}}Auto Speech \\ Recognition\end{tabular}                                    & \multicolumn{1}{c|}{39-1150M}            & \multicolumn{1}{c|}{\begin{tabular}[c]{@{}c@{}}Self-Supervised\\ Multi-task Learning\end{tabular}} & \multicolumn{1}{c|}{Unkown}                                                                                                         & \begin{tabular}[c]{@{}c@{}}LibriSpeech,\\ Multi-lingual dataset\end{tabular}              \\ \hline
  2023          & LLaMA~\cite{touvron2023llama}               & Decoder-Only                   & Text Generation                                                                                       & \multicolumn{1}{c|}{7-70B}               & \multicolumn{1}{c|}{\begin{tabular}[c]{@{}c@{}}Self-Supervised \\ RLHF\end{tabular}}               & \begin{tabular}[c]{@{}c@{}}Common Crawl,\\ C4, Github,\\ Wikipedia, Books,\\ ArXiv, StackExchange\end{tabular} & \begin{tabular}[c]{@{}c@{}}TruthfulQA, \\ ToxiGen, etc.\end{tabular}                                    \\ \hline
2023          & GPT-4~\cite{openai2023gpt4}               & \multirow{4}{*}{Close-Sourced} & \multirow{4}{*}{Text Generation}                                                                      & \multicolumn{3}{c|}{\multirow{4}{*}{Close-Sourced}}                                                                                                                                                                                                             & \begin{tabular}[c]{@{}c@{}}MMLU, HellaSwag,\\ ARC, WinoGrande,\\ HumanEval, DROP,\\ GSM-8K\end{tabular} \\ \cline{1-2} \cline{8-8} 
2023          & Claude2             &                                &                                                                                                       & \multicolumn{3}{c|}{}                                                                                                                                                                                                                                           & \multirow{2}{*}{Close-Sourced}                                                                          \\ \cline{1-2}
2023          & PaLM2               &                                &                                                                                                       & \multicolumn{3}{c|}{}                                                                                                                                                                                                                                           &                                                                                                         \\ \hline
\end{tabular}%
  }
  \caption{Milestone language foundation models and their typical tasks.}
  \label{tab:bkg-lm}
  \end{table}


\textbf{Encoder-only FMs.}
BERT~\cite{devlin2018bert} is an encoder-only Transformer model that utilizes a bidirectional masked language modeling approach during pre-training. In this approach, random words within a sentence are masked, and the model learns to predict these masked words by considering contextual clues. Fine-tuning BERT for various downstream tasks has led to SOTA performance, particularly in discriminative tasks like sentiment analysis and text classification.
DistilBERT~\cite{sanh2019distilbert} is a distilled version of BERT, being 40\% smaller and 60\% faster, yet retaining 97\% of BERT's language comprehension capabilities. 
RoBERTa~\cite{liu2019roberta} enhances BERT's efficacy through robust optimization techniques, including extended training duration, increased batch sizes, and a larger data corpus.
Sentence-BERT~\cite{reimers2019sentence} modifies BERT to generate semantically meaningful sentence embeddings. Employing siamese and triplet network structures, these embeddings can be directly compared using cosine similarity. This model has evolved into the widely utilized sentence-transformer tool\footnote{\url{https://www.sbert.net/}}, specializing in sentence embedding.

\textbf{Encoder-decoder FMs.}
T5~\cite{raffel2020exploring} employs an encoder-decoder architecture and is self-supervised, undergoing pre-training on the C4 dataset. This model introduces a unified framework that converts diverse text-based language problems into a text-to-text format, rendering it applicable to tasks such as summarization and question answering.
BART~\cite{lewis2019bart} serves as a denoising autoencoder in the pretraining phase, introducing corruption to text through an arbitrary noising function. The primary objective is to learn the reconstruction of the original text.

\textbf{Decoder-only FMs.}
The GPT family~\cite{radford2018improving, radford2019language, brown2020language} utilizes a decoder-only architecture for unsupervised training. GPT-1~\cite{radford2018improving}, the inaugural model in the series, features a transformer architecture with 117 million parameters and demonstrates the efficacy of pre-training on diverse internet text. GPT-2~\cite{radford2019language}, an enlarged iteration of GPT-1, undergoes unsupervised training on WebText, a dataset encompassing millions of webpages. 
GPT-3~\cite{brown2020language} represents a substantial increase in model size with 175 billion parameters, highlighting the advantages of scaling up. It demonstrated exceptional zero-shot performance. Instruct tuning~\cite{ouyang2022training} further enhances the model's capability to accurately follow instructions using human feedback, contributing to the creation of several open-source foundation models, including LLaMA~\cite{touvron2023llama}.
GLM~\cite{du2021glm} improves blank filling pretraining by adding 2D positional encodings and allowing an arbitrary order to predict spans, which results in performance gains over BERT and T5
on NLU tasks.
PaLM~\cite{chowdhery2023palm} was trained on 6144 TPU v4 chips using Pathways to further understand of the impact of model scale on few-shot learning.
Additionally, there are numerous close-sourced generative large FMs, including GPT-4\footnote{\url{https://openai.com/gpt-4}}, Claude2\footnote{\url{https://www.anthropic.com/index/claude-2}}, and PaLM 2\footnote{\url{https://ai.google/discover/palm2}}, among others.

\textbf{Speech FMs.}
Speech large FMs~\cite{radford2023robust, baevski2020wav2vec, hsu2021hubert} have been engineered to derive meaningful representations from raw audio signals. 
Wav2vec 2.0~\cite{baevski2020wav2vec} showcases, for the first time, that acquiring robust speech representations from unlabeled data could significantly improve performance in subsequent speech-related tasks. 
These models typically employ convolutional neural network to extract serial features and a transformer to capture contextual information. This approach is effective for various downstream tasks, including speech recognition and spoken language understanding. 
For instance, HuBERT~\cite{hsu2021hubert} leverages a transformer-based architecture for self-supervised learning of speech representations, trained on the 960-hour LibriSpeech audio dataset~\cite{panayotov2015librispeech}. 
Whisper~\cite{radford2023robust} represents a state-of-the-art open-sourced automatic speech recognition system, trained on a vast corpus of 680,000 hours of multilingual and multitask supervised data sourced from the web.

\subsubsection{Cost Analysis}
\label{subsec: llm-cost-analysis}

\begin{figure*}
	\centering
	\begin{minipage}[b]{0.6\textwidth}
	\begin{minipage}[b]{0.98\textwidth}
		\centering
		\includegraphics[width=0.9\textwidth]{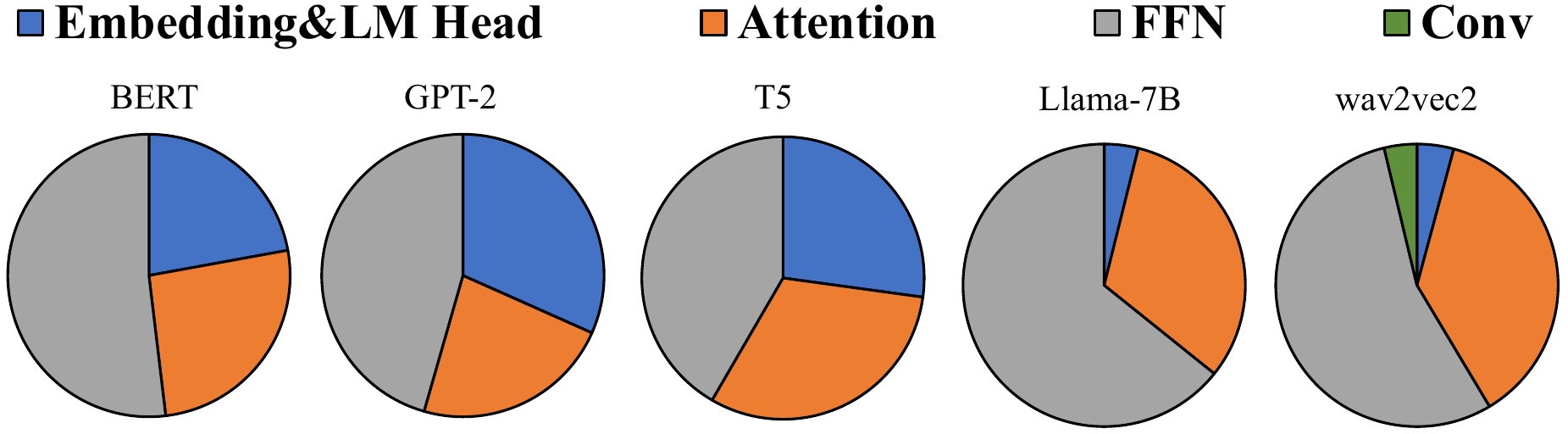}
		\subcaption{Weight proportion.}
		\label{fig:bkg-llm-storage-pct}
	\end{minipage}

	\begin{minipage}[b]{0.98\textwidth}
		\centering
		\includegraphics[width=0.9\textwidth]{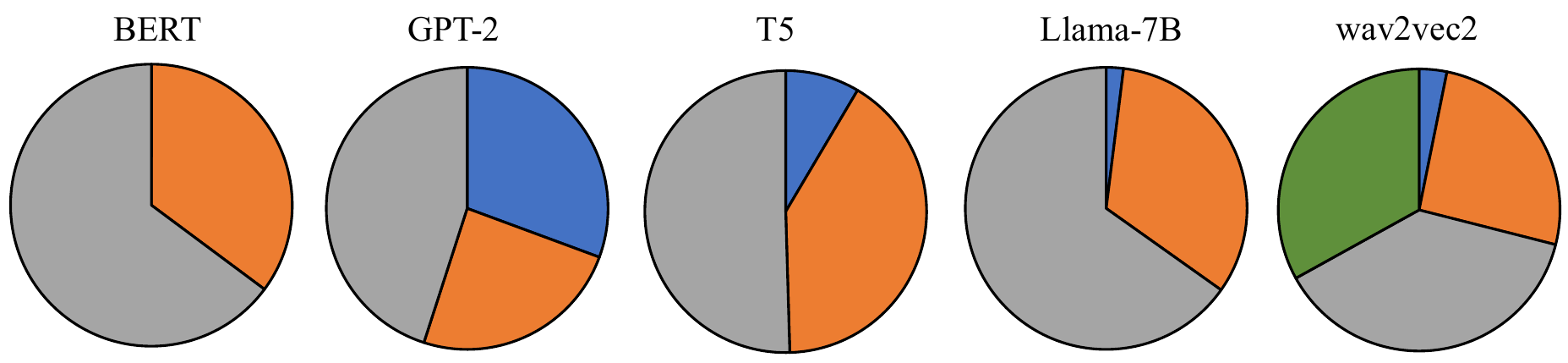}
		\subcaption{FLOPs proportion.}
		\label{fig:bkg-llm-flops-pct}
	\end{minipage}
	\caption{
		Weight storage and FLOPs cost of different language FMs. Input sequence length is 128.
		}
	\label{fig:bkg-llm-cost}
\end{minipage}
~
	\begin{minipage}[b]{0.33\textwidth}
		\centering
		\includegraphics[width=0.9\textwidth]{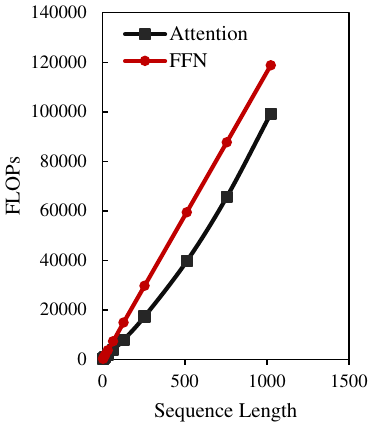}
		\caption{Inference FLOPs with different token length on GPT-2.
		}
		\vspace{0pt}
		\label{fig:bkg-llm-flops-gpt2}
	\end{minipage}
\end{figure*}

As depicted in Figure~\ref{fig:bkg-llm-cost}, we analyze the computational and storage costs associated with the primary components of large FMs\footnote{The analysis is conducted using the flops-profiler tool~\cite{flops-profiler}.}. The embedding component constitutes a significant portion of storage costs, approximately 25\% of the total. However, during inference, the embedding functions as a lookup table, incurring minimal computational costs. The FFN layer emerges as the most computation-intensive component, primarily due to the presence of two fully connected layers in each FFN block. lm\_head, which signifies the output layer of the model, varies depending on the task. For discriminative tasks like BERT, it takes the form of a classification layer with a softmax activation function, whereas for generative tasks like GPT/T5, it manifests as a linear layer. The size of this component is directly proportional to the vocabulary size.

\textbf{Cost Analysis Under Different Token Lengths.}
The attention mechanism in large FMs faces significant computational bottlenecks primarily due to its quadratic complexity. This complexity stems from calculating attention scores for every pair of positions within the input sequence, posing challenges in managing long sequences and impacting both training and inference efficiency. Additionally, beyond the attention mechanism, the computation complexity of the FFN scales linearly with input length but quadratically with the model's dimension.
As depicted in Figure~\ref{fig:bkg-llm-flops-gpt2}, an increase in the length of the input sequence causes a substantial rise in computational demand, attributable to the quadratic nature of the attention mechanism. 
In quantitative terms, the computation complexity of attention is $O(T^{2}D)$, while that of FFN is $O(TD^{2})$, where $T$ represents the sequence length and $D$ the hidden state dimension of the model~\cite{lin2022survey}. 
The decoder's attention mechanism, similar to that in the encoder, also experiences quadratic scaling with token length. 
This aspect becomes particularly significant in autoregressive decoding tasks, where each token's generation depends on the preceding ones, intensifying computational requirements. 
The implementation of a KV cache in the decoder can substantially mitigate computational costs by reusing key and value vectors across various positions. However, this comes at the expense of additional memory requirements~\cite{kwon2023efficient}.
Assuming that $B$ represents the batch size, $S$ the sequence length, $D$ the size of the hidden dimension, and $L$ the number of layers in the Transformer decoder, the memory required for storing the KV cache in single-precision format can be calculated as $(B \times S \times D \times L \times 2 \times 4)$ Bytes. 
This formula considers the dimensions of the batch, sequence, and hidden layer, as well as the number of layers in the decoder, with the factor of 2 representing the key and value in the cache and the factor of 4 accounting for the byte size of single-precision floating-point representation.

\textbf{Speech-specific Considerations.}
In speech processing applications, CNN  encoder block plays a significant role in computational complexity. 
The initial layers of the CNN block demand substantially more compute power, often exceeding that of each individual transformer layer by an order of magnitude. 
The increased requirement is attributed to the convolutional operations intrinsic to the CNN block, where a large number of calculations must be performed for each input token. For instance, despite the transformer having 19$\times$ more parameters, wav2vec 2.0 model~\cite{baevski2020wav2vec} incurs only 1.8$\times$ more computational load compared to the CNN block~\cite{gao2022federated}.

\subsection{Vision Foundation Models}
\begin{table}
    \centering
    \footnotesize
\resizebox{\columnwidth}{!}{%
\begin{tabular}{|c|c|c|c|c|c|c|c|}
\hline
\textbf{Year} &
  \textbf{Model Name} &
  \textbf{Model Arch.} &
  \textbf{Oriented Tasks} &
  \textbf{Parameters} &
  \textbf{Pre-training Method} &
  \textbf{Pre-training Datasets} &
  \textbf{Testing Datasets} \\ \hline
2020 &
  DETR~\cite{carion2020endtoend} &
  Encoder-Decoder &
  \begin{tabular}[c]{@{}c@{}}Object Detection\\ Instance Segmentation\\ Panoptic Segmentation\end{tabular} &
  40M &
  Supervised &
  COCO-2017 &
  COCO-2017 \\ \hline
2020 &
  ImageGPT~\cite{chen2020generative} &
  Decoder Only &
  \begin{tabular}[c]{@{}c@{}}Unconditional/Conditional \\ Image Generation\end{tabular} &
  3.7M--82M &
  Self-Supervised &
  ImageNet-1K &
  \begin{tabular}[c]{@{}c@{}}CIFAR-10\\ CIFAR-100\\ STL-10\\ ImageNet\end{tabular} \\ \hline
2021 &
  \begin{tabular}[c]{@{}c@{}}Vision \\ Transformer~\cite{dosovitskiy2020image}\end{tabular} &
  Encoder Only &
  Image Classification &
  \begin{tabular}[c]{@{}c@{}}86M--632M\\ ViT Base: 86M\\ ViT Large: 307M\\ ViT Huge: 632M\end{tabular} &
  \begin{tabular}[c]{@{}c@{}}Supervised\\ Self-Supervised\end{tabular} &
  ImageNet-21K &
  ImageNet-1K \\ \hline
2021 &
  DeiT~\cite{touvron2021training} &
  Encoder Only &
  Image Classification &
  5M--88M &
  Distilled &
  ImageNet-1K &
  \begin{tabular}[c]{@{}c@{}}ImageNet \\ CIFAR-10 \\ CIFAR-100 \\ Flowers \\ Cars \\ iNat-18 \\ iNat-19\end{tabular} \\ \hline
2021 &
  SegFormer~\cite{xie2021segformer} &
  Encoder-Decoder &
  Segmentation &
  3.7M--82M &
  supervised &
  Imagenet-1K &
  \begin{tabular}[c]{@{}c@{}}Cityscapes\\ ADE20K\\ COCO-Stuff\end{tabular} \\ \hline
2021 &
  YOLOS~\cite{YOLOS} &
  Encoder Only &
  Object Detection &
  5M &
  \begin{tabular}[c]{@{}c@{}}Supervised\\ Self-supervised\end{tabular} &
  ImageNet-1K &
  COCO \\ \hline
2022 &
  \begin{tabular}[c]{@{}c@{}}Swin \\ Transformer~\cite{liu2021swin}\end{tabular} &
  Encoder Only &
  \begin{tabular}[c]{@{}c@{}}Image Classification\\ Object Detection\\ Semantic Segmentation\\ Video Action Classification\end{tabular} &
  26M &
  Self-Supervised &
  ImageNet-22K &
  \begin{tabular}[c]{@{}c@{}}ImageNet-1K\\ ADE20K\\ COCO\\ Object 365 v2\end{tabular} \\ \hline
2022 &
  ViTMAE~\cite{he2022masked} &
  Encoder Only &
  Image Classification &
  86M--632M &
  Self-Supervised &
  ImageNet-1K &
  \begin{tabular}[c]{@{}c@{}}COCO\\ ADE20K\\ iNat\\ Places\end{tabular} \\ \hline
2022 &
  ViTDet~\cite{li2022exploring} &
  Encoder Only &
  Object Detection &
  86M--632M &
  \begin{tabular}[c]{@{}c@{}}Self-Supervised\\ MAE\end{tabular} &
  ImageNet-1K &
  COCO \\ \hline
2022 &
  BEiT~\cite{bao2022beit} &
  Encoder Only &
  \begin{tabular}[c]{@{}c@{}}Image Classification\\ Semantic Segmentation\end{tabular} &
  86M--632M &
  Self-Supervised &
  ImageNet-1K &
  \begin{tabular}[c]{@{}c@{}}ImageNet-1K\\ ADE20K\end{tabular} \\ \hline
\multicolumn{1}{|l|}{2023} &
  DINOv2~\cite{oquab2023dinov2} &
  Encoder Only &
  Image Classification &
  1.1B &
  Self-Supervised &
  LVD-142M &
  \begin{tabular}[c]{@{}c@{}}ImageNet-1K\\ Im-A\\ ADE-20k\\ Oxford-M\end{tabular} \\ \hline
\multicolumn{1}{|l|}{2023} &
  LVM~\cite{bai2023sequential} &
  Encoder-Decoder &
  \begin{tabular}[c]{@{}c@{}}Semantic Segmentation\\ Depth Estimation\\ Surface Normal Estimation\\ Edge Detection\end{tabular} &
  300M--3B &
  Self-Supervised &
  UVD-v1 &
  \begin{tabular}[c]{@{}c@{}}Kinetics-700\\ ImageNet\end{tabular} \\ \hline
\end{tabular}
}
    \caption{Milestone vision FMs and their typical tasks. }
    \label{tab:bkg-vit}
\end{table}
\label{subsec:bkgnd-vit}

\subsubsection{Model Architecture}

The advancements in transformers have propelled the emergence of foundation models in the field of computer vision. 

\textbf{Vision Transformer pipeline.} Vision Transformer (ViT) is the most classic transformer-empowered vision model.
It is inspired by the growing trend of self-supervised pre-trained NLP models like BERT, which is encoder-only.
Given an input image, ViT firstly splits image into fixed-size patches (i.e., tokens) by a convolutional embedding layer.
For instance, a standard size RGB image input (i.e., 3$\times$224$\times$224) will be splitted to 14$\times$14 patches with 16$\times$16 pixels.
This embedding overhead is almost negligible compared to the following compute-intensive transformer encoder (e.g., less than 5\%).
Besides, an extra learnable classification token ([CLS]) is added to the token sequence in order to perform classification.
After that, positional embeddings are added into each token, and tokens are fed to a standard Transformer encoder, which has been depicted in Figure~\ref{fig:bkg-tree} and $\S$\ref{subsec:bkgnd-LLM}. 
Depending on the specific downstream tasks, the hidden states generated by the Transformer encoder are finally fed into different heads, such as classification, detection, segmentation, etc.




\subsubsection{Representative Models and Downstream Tasks}

\textbf{Encoder-only.} 
Most visual foundation models are encoder-only architecture.
ViT~\cite{dosovitskiy2020image} is the first work that successfully trains a Transformer encoder on ImageNet, with alignment to BERT in terms of parameter amount.
It performs both supervised and self-supervised pre-training on a large-scale ImageNet-21k dataset.
Although it shows competitive accuracy and scalabilty, its demand for training data compared to traditional CNNs is still an obstacle.
To this end, DeiT~\cite{touvron2021training} is proposed with much impact.
DeiT performs a distillation-empowered pre-training, which enhances the data-efficiency of ViT training.

Another storyline is to push the boundary of self-supervised ViT pre-training.
BEiT~\cite{bao2022beit} turns the pre-training objective to recover the original visual tokens based on the corrupted image patches.
The authors claim that BEiT is ``a path to the BERT moment of CV''.
MAE~\cite{he2022masked} introduces a lightweight decoder to the encoder training for reconstructing the masked patches.
MAE can effectively mask out most of the patches (by even 80\%).
Thereby, the training of the encoders can be very cost-effective, which paves the way for large pre-trained vision models.

YOLOS~\cite{YOLOS} is an object detection model built atop ViT.
It demonstrates the transferability of the vanilla ViT pre-trained on mid-sized ImageNet-1k to the more challenging COCO object detection benchmark.
ViTDet~\cite{li2022exploring} enables a plain, non-hierarchical ViT to serve as a backbone network for object detection.
ViTDet allows the original ViT architecture to be fine-tuned for object detection without needing to redesign a hierarchical backbone for MAE-style pre-training.
Swin Transformer~\cite{liu2021swin} is a representative work that optimizes the attention mechanism.
This model is a hierarchical ViT whose
representation is computed with shifted windows.
DINOv2~\cite{oquab2023dinov2} conducts training on a ViT model with 1 billion parameters and then distills it into a set of smaller models that outperform the leading all-purpose features, such as OpenCLIP, on the majority of benchmarks at both image and pixel levels.

\textbf{Encoder-decoder.} 
DETR~\cite{carion2020endtoend} is an early effort to build an end-to-end detection pipeline with transformers.
The architecture of DETR is cascaded: it consists of a CNN backbone and an encoder-decoder transformer.
DETR supports object detection, instance segmentation and panoptic segmentation through supervised training.
The parameter amount of DETR aligns with Faster-RCNN~\cite{ren2016faster}, which has about 40M parameters. 
SegFormer~\cite{xie2021segformer} is a semantic segmentation model which unifies Transformers with lightweight multilayer perception (MLP) decoders.
LVM~\cite{bai2023sequential} has achieved effective learning of visual information using a purely visual approach through image sequence modeling, without the need for any linguistic data.


\subsubsection{Cost Analysis}
Due to the alignment of ViT's architecture with BERT, its resource consumption is also similar.
However, unlike language models such as BERT, visual models typically have fixed-length inputs.
For standard image inputs, such as 14$\times$14 patches or 16$\times$16 patches, the computational bottleneck lies in the fully connected layers in the FFN and attention.
Please refer to $\S$\ref{subsec:bkgnd-LLM} for more details.



\subsection{Multimodal Foundation Models}\label{subsec:bkgnd-mfm}
Multimodality is currently a hot research direction in FM research. 
A large FM often exhibits strong capabilities in cross-modal understanding, translation, and generation.

In general, there are two lines of research on multimodal FMs:
one is to encode data in different modalities into the same latent space, mostly adopting transformer encoders;
the other one is to generate data in different modalities, often using transformer decoders.
Specifically, the multimodal generation mainly centers around text-based image generation, a challenging and realistic ML task that sees great advancements in recent years.
The two lines of research have convergence, e.g., multimodal-to-multimodal (or even any-to-any) generation.

\subsubsection{Key Architectures}


To ingest and align multimodal input data, existing model architectures typically consist of multiple encoders, with each modality having its own set of transformer encoders.
Notably, these encoders are generally trained from scratch, utilizing paired data with the aligned modalities and current modality.
Upon receiving input from diverse modalities, they initially encode this data into normalized, fixed-dimensional embeddings.
By mapping these embeddings to a high-dimensional space and designing a loss function, researchers aim to minimize the distance between different modalities in the joint semantic space. This approach aligns different modalities and enhances the consistency of their representations.

With multimodal data aligned, existing research either
(i) reuse the LLM that is trained on pure text corpora to generate text;
(ii) or diffusion models to generate high-quality image pixels.
In the first case, the LLM module is designed to comprehend, reason about, and produce output based on input data aligned with the text modality. This module typically adopts a decoder-only architecture. Due to extensive pretraining on numerous large-scale corpus datasets, LLMs are endowed with rich semantic knowledge. This enables them to effectively comprehend data embedded within the text modality and generate text-based output in an autoregressive fashion when performing specific tasks.
In the second case, the diffusion module aims to generate high-quality images by eliminating redundant noise present in the input image.
In the training stage, the models introduce noise into an image, transforming it into a random state. In contrast, during the inference stage, this process is reversed, gradually removing the noise. This denoising process is essential for improving image clarity, leading to images with high resolution and detailed sharpness.
Stable diffusion models have significantly progressed this technology, showcasing distinctive capabilities in generating high-quality images customized to specific textual and pictorial descriptions. In addition to the multimodal embedding input, the diffusion module primarily consists of two components:  an image encoder/decoder and a denoising network.




\textit{Image Encoder/Decoder}. Diffusion model is the state-of-the-art approach for text-to-image generation. The encoder takes an input image and compresses it into a lower-dimensional latent representation. This compression is vital for reducing the computational load and enhancing the model's efficiency. The decoder operates in reverse, taking the latent representation and reconstructing it back into a high-resolution image. This process is critical for the model's capability to generate detailed visual content.
Variational autoencoder (VAE)~\cite{kingma2013auto} is a generative model that is employed to learn the latent space of the image. VAE consists of an encoder and a decoder: the encoder is responsible for mapping the image to the latent space, while the decoder is responsible for mapping the latent space to the image space. Both encoder and decoder networks are often built with convolutional neural layers. VAE is trained by minimizing the reconstruction loss and the KL divergence loss. The reconstruction loss is responsible for ensuring that the image generated by the decoder is similar to the original image, while the KL divergence loss is responsible for ensuring that the latent space is similar to the standard normal distribution. VAE is employed in the diffusion model to learn the latent space of the image. 
Another variant of VAE model that is often employed in diffusion tasks is VQ-VAE~\cite{van2017neural}, to learn the latent space of the image through a vector quantization layer. The vector quantization layer applies the vector quantization technique, quantizing each pixel of the image to the nearest codebook vector. In this way, the VQ-VAE model can encode and decode the image in a more efficient way.

\textit{Denoising Network}. Denoising network progressively removes noise from the encoded images by predicting the noise distribution and evicting it through sampling algorithms, such as DDPM~\cite{ho2020denoising} and DDIM~\cite{song2020denoising}. Initially, during the training phase, the model adds noise to the images, gradually leading them to a state of pure randomness. The denoising network then learns to reverse this noise addition, step by step, during the inference phase. This gradual denoising is crucial for enhancing the clarity and quality of the final image output. U-Net~\cite{ronneberger2015u} is often employed as the noise prediction network in diffusion models, which is a convolutional neural network model, consisting of a contracting path and an expansive path. The contracting path is responsible for capturing context information in the image by mapping the image to high-dimensional space, while the expansive path facilitates precise localization by upsampling. To retain the information on the contracting path, the expansive path is connected to the contracting path via skip connections. The U-Net model is a popular choice for image segmentation tasks and is also employed in the diffusion model to predict noise within an image.

\textit{Fusion decoder (FD)}. Moreover, FD module aims to enhance the understanding of images based on both the image itself and associated image prompts, and it produces outputs according to task requirements.
This module typically involves designing a fusion decoder and is pre-trained on both image and text datasets, enabling it to jointly process image and text representations.
This module typically encompasses the design of a fusion decoder and undergoes pretraining on both image and text datasets. This module enables it to collectively process image and text representations.


\begin{table}
    \centering
    \footnotesize
\resizebox{\columnwidth}{!}{%
\begin{tabular}{|c|c|c|c|c|c|c|c|}
\hline
\textbf{Year} &
  \textbf{Model Name} &
  \textbf{Model Arch.} &
  \textbf{Oriented Tasks} &
  \textbf{Parameters} &
  \textbf{Pre-training Method} &
  \textbf{Pre-training Datasets} &
  \textbf{Testing Datasets} \\ \hline

  
  2021  & 
  DALL-E~\cite{ramesh2021zero}    & 
  Encoder-Decoder & 
  Cross-Modal Generation   & 
  12B          & 
  Supervised     & 
  \begin{tabular}[c]{@{}c@{}}MS-COCO\\ YFCC100M\end{tabular}                 & 
  \begin{tabular}[c]{@{}c@{}}MS-COCO\\ CUB\end{tabular} \\ \hline
  
  2021 &
  CLIP ~\cite{radford2021learning} &
  Encoder-Only&
  Cross-Modal Generation &
  \begin{tabular}[c]{@{}c@{}}400M--1.6B\\CLIP-ViT-B: 400M\\ CLIP-RN50: 400M\\ CLIP-RN101: 500M \\ CLIP-RN50*4: 1.6B \end{tabular}  &
  \begin{tabular}[c]{@{}c@{}}Supervised\\ Self-Supervised\end{tabular} &
  WebImageText &
  NYU-D \\ \hline

  2021  & 
  Stable Diffusion~\cite{rombach2022high}   & 
  Encoder-Decoder & 
  Image Synthesis            & 
  \begin{tabular}[c]{@{}c@{}}LDM (LAION): 1.45B\\ SD2.1: 1.3B\end{tabular}              & 
  Supervised     & 
  \begin{tabular}[c]{@{}c@{}}CelebA-HQ\\ FFHQ\\ LSUN\\ ImageNet\\ LAION\end{tabular}              & 
  \begin{tabular}[c]{@{}c@{}}CelebA-HQ\\ FFHQ\\ LSUN\\ ImageNet\\ LAION\end{tabular}              \\ \hline

  2022 &
  Flamingo ~\cite{alayrac2022flamingo} &
  Encoder-Decoder &
  \begin{tabular}[c]{@{}c@{}}Visual Question Answering\\ Image Caption \end{tabular} &
  \begin{tabular}[c]{@{}c@{}}3B \\ 9B \\ 80B \end{tabular} &
  Supervised &
  \begin{tabular}[c]{@{}c@{}}M3W \\ ALIGN \\ LTIP \\  VTP \end{tabular} &
  \begin{tabular}[c]{@{}c@{}}COCO, VATEX \\ VizWiz, MSRVTTQA \\ VisDial, YouCook2 \\  TextVQA,  HatefulMeme \end{tabular}  \\ \hline

  2023 &
  SAM ~\cite{kirillov2023segment} &
  Encoder-Decoder &
 \begin{tabular}[c]{@{}c@{}}Edge Detection \\ Object Proposal Generation \\ Instance Segmentation \\  Text-to-mask Prediction \end{tabular} &
  1B &
  Supervised &
  \begin{tabular}[c]{@{}c@{}}COCO \\ SA-1B \\ LVIS V1 \\ ADE20K\\ Open Images V5  \end{tabular} &
  SA-1B  \\ \hline

  2023 &
  ImageBind ~\cite{girdhar2023imagebind}&
  Encoder-Only&
  Modal Alignment &
  1.2B &
  Supervised &
  \begin{tabular}[c]{@{}c@{}}Image-text pairs from \\ large-scale web data \end{tabular} &
  \begin{tabular}[c]{@{}c@{}}ImageNet-IN1K \\ Places-365-P365 \\ Kinetics400-K400  \\ MSR-VTT,  NYU-D \\ SUN-D, AS-A, VGGS\\ ESC, LLVIP, Ego4D \end{tabular}  \\ \hline

  2023 &
  CoDi ~\cite{tang2023any}&
  Encoder-Decoder &
  Cross-Modal Generation &
  3.5B &
  Supervised &
  \begin{tabular}[c]{@{}c@{}}Laion400M \\ Freesound 500K \\ WebVid \\ HD-Villa-100M \\ ACAV100M \end{tabular} &
  \begin{tabular}[c]{@{}c@{}}AudioCaps \\ MSR-VTT \end{tabular} \\ \hline

  2023  & 
  Consistance Models~\cite{song2023consistency} & 
  Encoder-Decoder & 
  Image Synthesis            & 
  \begin{tabular}[c]{@{}c@{}}ImageNet: 281M\\ LSRN: 500M\end{tabular}         & 
  Supervised     & 
  \begin{tabular}[c]{@{}c@{}}ImageNet\\ LSUN\end{tabular}                    & 
  \begin{tabular}[c]{@{}c@{}}ImageNet\\ LSUN\end{tabular}                \\ \hline   

  2023 &
  NExT-GPT ~\cite{wu2023next}&
  Encoder-Decoder &
  Cross-Modal Generation &
  9.1B-10B &
  Supervised &
  X-caption &
  \begin{tabular}[c]{@{}c@{}} MSR-VTT, AudioCaps\\ MosIT, COCO-caption \\ COCO, DAVIS, VCTK \end{tabular} \\ \hline

  2023 &
  MiniGPT-4 ~\cite{zhu2023minigpt} &
  Encoder-Decoder &
  Cross-Modal Generation &
  13B &
  Supervised &
  \begin{tabular}[c]{@{}c@{}}Conceptual Caption \\ SBU, LAION \end{tabular} &
  \begin{tabular}[c]{@{}c@{}} Localized Narratives \\ AOK-VQA, GQA \end{tabular}\\ \hline

2023          & GPT-4V~\cite{gpt4v}            &   Close-Sourced  & Cross-Modal  Generation                                                         & \multicolumn{3}{c|}{Close-Sourced}  
&\begin{tabular}[c]{@{}c@{}}COCO \\ ADE20K \\ Flickr30K \\ RefCOCO\\ DAVIS2017  \end{tabular}   \\ \hline                                                                                                                      
  2023 &
  LLaVA ~\cite{liu2023visual} &
  Encoder-Decoder &
  \begin{tabular}[c]{@{}c@{}}Visual Chat \\ Optical Character Recognition \\ Science Question Answering \end{tabular} &
  \begin{tabular}[c]{@{}c@{}}7B \\ 13B \end{tabular} &
  Supervised &
  \begin{tabular}[c]{@{}c@{}}CC \\ LAION \\ COCO \\  CC3M \end{tabular} &
  \begin{tabular}[c]{@{}c@{}}LLaVA-Bench (COCO) \\ LLaVA-Bench (In-the-Wild) \\ ScienceQA \end{tabular} \\ \hline

2023          & Gemini~\cite{team2023gemini}             &   Close-Sourced           &  Cross-Modal   Generation                                                                                                  & \begin{tabular}[c]{@{}c@{}}Gemini-Nano1: 1.8B \\ Gemini-Nano2: 3.25B  \end{tabular}                     &\multicolumn{2}{c|}{Close-Sourced}    &   \begin{tabular}[c]{@{}c@{}}MMLU, GSM8K, ChartQA \\ BIG-Bench-Hard, HumanEval\\ Natural2Code, DROP, HellaSwag\\ WMT23, MBPP, NaturalQuestions\\TydiQA, BoolQ, MGSM, XLsum\\ AI2D, MMMU, TextVQA, DocVQA \\ MATH, InfographicVQA, Wikilingua \\ XM-3600, VATEX, ActivityNet-QA \\NextQA, Perception Test MCQA   \end{tabular}                                                                                              \\ \hline

\end{tabular}
}
    \caption{Milestone multimodal FMs and their typical tasks.}
    \label{tab:bkg-mm}
\end{table}



\subsubsection{Representative Models and Downstream Tasks}

\textbf{Multi-Encoder FMs:}
CLIP ~\cite{radford2021learning}, ALBEF~\cite{li2021align}, and ALIGN~\cite{jia2021scaling} are some earliest works to propose cross-modal alignment, aiming to learn richer image representations from text by establishing connections between images and text. 
While these models initially demonstrated the potential of multimodality, their performance was limited by the capabilities of the image and text encoders, as well as the quantity and quality of image-text pair data. 
Subsequent works like ImageBind~\cite{girdhar2023imagebind} and LanguageBind~\cite{u2023languagebind} further extended modal alignment.
These models employ diverse intermediate modalities as the aligned modal, effectively mapping representations from various sources into the feature space of the intermediate modal, thereby facilitating cross-modal transformations within a joint vector space.  
However, significant challenges arise in aligning multimodal representations with those of intermediate modalities, primarily attributed to limitations in encoder capabilities. These limitations, in turn, impact the overall performance of the model.

\textbf{Encoder-Decoder FMs} utilize the embedding module for modality conversion, allowing the transformed modality to be compatible with the generator.

\textit{(1) Encoder-Large FMs}: 
PandaGPT~\cite{su2023pandagpt} utilizes multiple single-modal encoders to align inputs to an intermediate modality. 
PandaGPT feeds the intermediate modality into large FMs for generation, followed by further transformation by the decoder of the target modality.
Furthermore, BLIP-2~\cite{li2023blip} and MiniGPT-4~\cite{zhu2023minigpt} focus on cross-modal generation of text and images, by designing image modal encoders and using Q-Former to fuse these image modalities with text modalities before feeding them into multimodal large FMs for cross-modal generation. 
Meanwhile, mPLUG~\cite{ye2023mplug} and LLaVA~\cite{liu2023visual} focus on improving the capacity of modality transformation to enhance the availability and reliability of the generated results.
\update{
MobileVLM V2~\cite{chu2024mobilevlm}, a highly efficient vision-language model designed for resource-constrained devices, utilizes a CLIP-based encoder and MobileLLaMA-based decoder to achieve superior performance while maintaining fast inference speeds.
}
Additionally, Flamingo~\cite{alayrac2022flamingo} and LLaMA-Adapter~\cite{zhang2023llama} explore how to tune multimodal large FMs with lower costs, leading to the generation of higher-quality multimodal outputs.
PaLM-E~\cite{driess2023palm} and HuggingGPT~\cite{shen2023hugginggpt} focus on Embodied Intelligence by using large FMs as a central component to incorporate Embodied data into multimodal inputs. 
These models further design agents to decompose tasks and leverage generative capabilities to accomplish complex tasks.
    
    \textit{(2) Encoder-Diffusion FMs}:
    Stable diffusion~\cite{rombach2022high} is capable of generating high-quality images, by gradually evicting noise in images through a learned process, leading to clear and detailed visual outputs. This model is applied in various downstream tasks such as generating detailed images from text descriptions (text-to-image generation), restoring or completing parts of images (image inpainting), modifying specific aspects of existing images (image editing), and enhancing the resolution of images (image super-resolution). Stable diffusion's adaptability in these areas makes it a valuable tool in the field of image processing and generation.
    Consistency models~\cite{song2023consistency} are developed to enhance the efficiency of diffusion models in generating high-quality images, audio, and video. These models facilitate rapid single-step generation, overcoming the slow sampling speed associated with traditional diffusion models. They exhibit the capability to perform zero-shot data editing tasks such as image inpainting, colorization, and super-resolution, without necessitating specific training for these tasks. 
DALL-E~\cite{ramesh2021zero} is primarily employed for image generation, demonstrating the capability to create diverse and complex images based on textual descriptions. This model integrates elements of natural language understanding and computer vision, allowing it to produce images that faithfully represent a broad spectrum of textual prompts, ranging from simple descriptions to intricate scenarios.

In addition to stable diffusion, there is a notable emphasis on “any-to-any” generative models, designed to transform diverse types of inputs into a wide array of outputs. 
CoDi~\cite{tang2023any} is designed to produce various output modalities, such as language, image, video, or audio, from different input combinations. Its uniqueness lies in the capability to concurrently generate multiple modalities, without being constrained to specific input types. CoDi aligns modalities in input and output spaces, thereby facilitating the generation of combinations not present in training data. 
NExT-GPT~\cite{wu2023next} exhibits the capability to perceive inputs and generate outputs across various modalities, including text, images, videos, and audio. NExT-GPT integrates large FMs with multimodal adaptors and diffusion models. The system undergoes fine-tuning with minimal parameter changes, facilitating cost-effective training and straightforward modality expansion. Employing modality-switching instruction tuning and leveraging a specially curated dataset, NExT-GPT enhances cross-modal content generation and understanding, with the objective of modeling universal modalities.
\update{
Similarly, M4~\cite{yuan2024mobile} introduces a scalable mobile AI foundation model that unifies diverse AI tasks by employing multimodal embeddings and a transformer-based backbone for understanding and reasoning across inputs such as text, images, audio, and motion data, enhancing efficiency and scalability for mobile AI.
}

    \textit{(3) Encoder-FD FMs}:
    UNITER~\cite{chen2020uniter} is one of the earliest works to propose a universal fusion of image and text in multimodal settings. 
    It aims to combine image and text features through Transformers to obtain joint features. 
    Building upon this, subsequent works such as FLAVA~\cite{singh2022flava}, CoCa~\cite{yu2022coca}, and GLIP~\cite{li2022grounded} delve deeper into how to use decoders to better fuse and align image and text representations, thereby enhancing multimodal reasoning. 
    \update{
    Additionally, SAM~\cite{kirillov2023segment} and SAM 2~\cite{ravi2024sam2} take a step further by leveraging a decoder to fuse prompt embeddings corresponding to images, enabling zero-shot automatic segmentation of images/videos based solely on text prompts.
    }

\begin{figure*}
	\centering
	\includegraphics[width=0.7\textwidth]{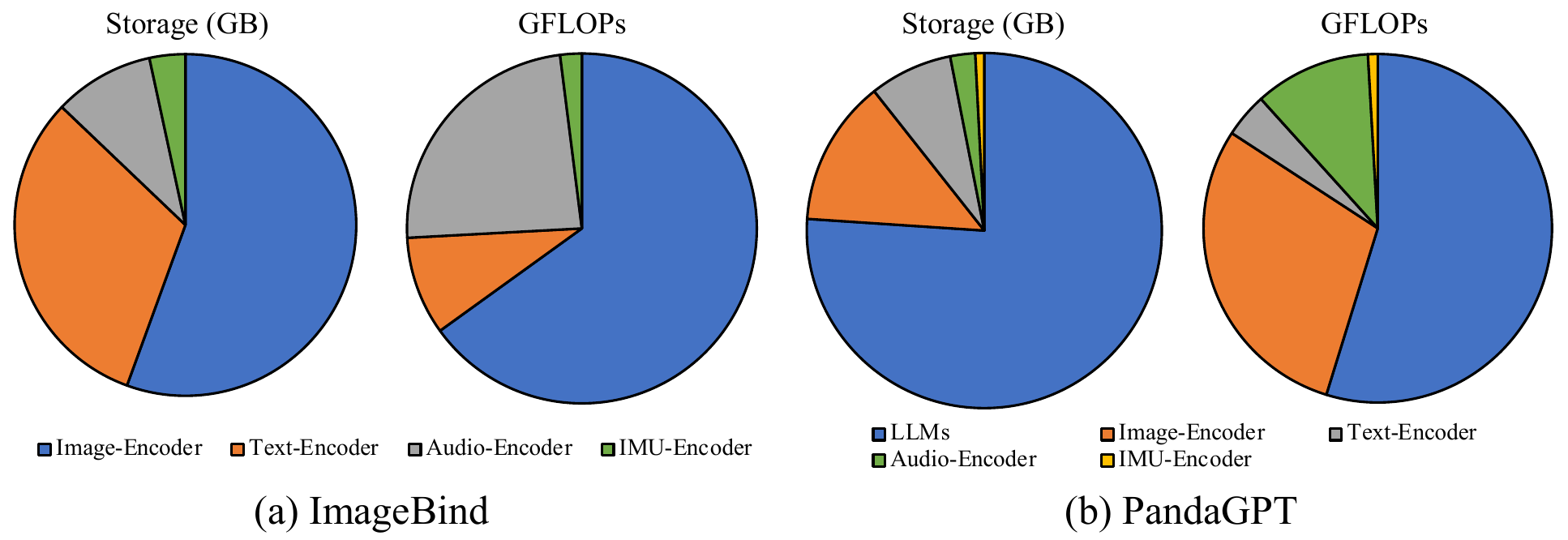}
	\caption{
		The cost of different modules in multimodal FMs. 
		}
	\label{fig:bkg-mm}
\end{figure*}

\subsubsection{Cost Analysis} 


\textbf{Multi-Encoder module:} 
Multi-encoder module is a specialized encoder module designed for modality alignment.
This module attuned to various modal inputs, effectively aligns these inputs into a unified semantic space using distinct encoder architectures.
Specifically, the primary encoder modules consist of the Image-encoder, Text-encoder, Audio-encoder, and IMU-encoder.
As shown in Figure~\ref{fig:bkg-mm}(a),  the encoder module has 0.27B parameters (on average),  occupies 1.1G memory (on average), with total GFLOPs of 65.9 for processing a sample (on average). 
Notably, the Image-encoder emerges as the most resource-intensive component, with 0.63B parameters, occupying 2.4G memory, and executing 167.59 GFLOPs for a single sample.

\begin{figure*}
    \centering
    \includegraphics[width=0.7\textwidth]{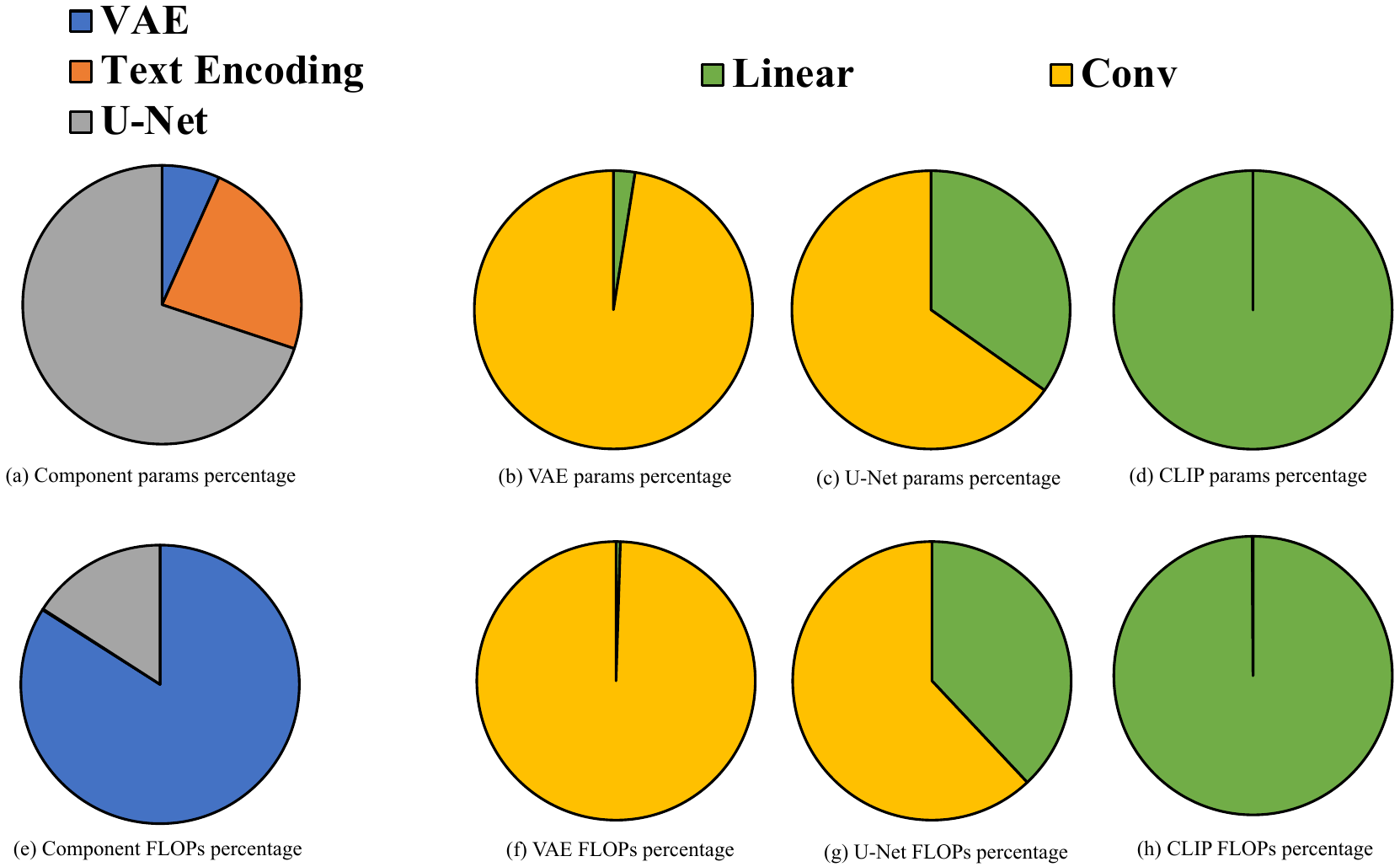}
    \caption{Parameters and FLOPs of different modules in Stable Diffusion 2.1.}
    \label{fig:bkg-diffusion-cost}
\end{figure*}

\textbf{Decoder module:} 
In multimodal models, in addition to the multi-encoder module, there is also a decoder module composed of large FMs, diffusion module, and FD module.

(1) Large FMs module:
This module receives inputs aligned from diverse modalities, enabling autoregressive generation. 
The cost of the module mainly depends on the size of the large FM parameters. 
As shown in Figure~\ref{fig:bkg-mm}(b), taking an example of the integrated Vicuna-7B, this model consists of 7B parameters, occupies 14G memory, and with total GFLOPs of 312 for processing a sample, significantly exceeding the resource requirements of the encoder module.

(2) Diffusion module:
This module receives optional conditioning inputs and generates high-quality images. Given the variable sizes of these modules in diffusion models, we focus on Stable Diffusion 2.1 as a representative example for this discussion. Stable Diffusion 2.1 incorporates a U-Net for denoising, a VAE for image encoding and decoding, and a CLIP model as the text encoder. Figure~\ref{fig:bkg-diffusion-cost}(a) and (e) show the FLOPs and parameter number percentage in the stable diffusion 2.1 model. We use the textual prompt consisting of 10 tokens for illustration. Figure~\ref{fig:bkg-diffusion-cost}(b) and (f) show the FLOPs and parameter number percentage in the VAE model. Figure~\ref{fig:bkg-diffusion-cost}(c) and (g) show the FLOPs and parameter number percentage in the U-Net model. Figure~\ref{fig:bkg-diffusion-cost}(d) and (h) show the FLOPs and parameter number percentage in the CLIP model.
U-Net of Stable Diffusion 2.1 takes image latents of shape 4$\times$96$\times$96 as input and predicts noise in the latent space. The model is trained on the LAION-5B dataset~\cite{schuhmann2022laion}. The U-Net has 865M parameters, with total FLOPs of 759G for processing a sample.
Stable Diffusion 2.1 incorporates a VAE that encodes images to latent space and decodes them. VAE takes images with resolution 3$\times$768$\times$768 as input and encodes them to 4$\times$96$\times$96. VAE is co-trained with the U-Net on the same dataset. VAE has 83M parameters, with total FLOPs of 4T for processing a sample.
Stable Diffusion 2.1 uses CLIP~\cite{radford2021learning} model for text encoding. CLIP  is trained on various (image, text) pair datasets, such as the LAION dataset and DataComp dataset~\cite{gadre2023datacomp}. CLIP takes sentences as the input and encodes each token to a hidden vector with a size of 1024. The model has 289M parameters, with a total FLOPs of 289M for processing a single token. 

(3) FD module: Due to the similar structures between FD and ViT, their resource consumption and computational bottleneck are similar. 
For more details, please refer to $\S$\ref{subsec:bkgnd-vit}.

	\section{RESOURCE-EFFICIENT ARCHITECTURES}
\label{sec:architecture}

\tikzstyle{my-box}=[
    rectangle,
    draw=black,
    rounded corners,
    text opacity=1,
    minimum height=1.5em,
    minimum width=5em,
    inner sep=2pt,
    align=center,
    fill opacity=.5,
    line width=0.5pt,
]
\tikzstyle{leaf}=[my-box, minimum height=1.5em,
    fill=hidden-red!10, text=black, align=left,font=\normalsize,
    inner xsep=2pt,
    inner ysep=4pt,
    line width=0.8pt,
]

\begin{figure*}[htpb]
    \centering
    \resizebox{\textwidth}{!}{
        \begin{forest}
            forked edges,
            for tree={
                grow=east,
                reversed=true,
                anchor=base west,
                parent anchor=east,
                child anchor=west,
                base=center,
                font=\large,
                rectangle,
                draw=black,
                rounded corners,
                align=left,
                text centered,
                minimum width=4em,
                edge+={black, line width=1pt},
                s sep=3pt,
                inner xsep=2pt,
                inner ysep=3pt,
                line width=0.8pt,
                ver/.style={rotate=90, child anchor=north, parent anchor=south, anchor=center},
            },
            where level=1{text width=12em,font=\normalsize,}{},
            where level=2{text width=12em,font=\normalsize,}{},
            [
                \textbf{Resource-efficient Architecture}, ver
                        [
                          \textbf{Efficient Attention}\\
                          \textbf{~~~~~~~($\S$\ref{subsec:efficient-attention})}, fill=blue!10
                          [
                          \textbf{Sparse Attention ($\S$\ref{subsubsec:sparse-attention})}, fill=yellow!10
                          [
                          Longformer~\cite{beltagy2020longformer}{, }
                          ETC~\cite{ainslie2020etc}{, }
                          BIGBIRD~\cite{zaheer2020big}{, }
                          HEPOS~\cite{huang2021efficient}{, }\\
                          MATE~\cite{eisenschlos2021mate}{, }
                          LittleBird~\cite{lee2022littlebird}{, }
                          ALBERT~\cite{lan2019albert}{, } 
                          TDANet~\cite{li2022efficient},
                          leaf, 
                          text width= 33em
                          ]
                          ]
                          [
                          \textbf{Approximate Attention}\\
                          \textbf{~~~~~~~~~~($\S$\ref{subsubsec:approximate-attention})}, fill=yellow!10
                          [
                          Reformer~\cite{kitaev2019reformer}{, }
                          Transformers are rnns~\cite{katharopoulos2020transformers}{, }
                          Linformer~\cite{wang2020linformer}{, }\\
                          Performers~\cite{choromanski2020rethinking}{, }
                          Luna~\cite{ma2021luna}{, }
                          Mega~\cite{ma2022mega}{, }\\
                          Deformable Attention~\cite{xia2022vision}{, }
                          CrossViT~\cite{chen2021crossvit}{, }
                          PolySketchFormer~\cite{kacham2023polysketchformer},
                          leaf, text width=33em
                          ]
                          ]
                          [
                          \textbf{Attention-Free Approaches}\\
                          \textbf{~~~~~~~~~~~~($\S$\ref{subsubsec:attention-free})}, fill=yellow!10
                          [
                          AFT~\cite{zhai2021attention}{, }
                          Perceiver~\cite{jaegle2021perceiver}{, }
                          Hyena hierarchy~\cite{poli2023hyena}{, }
                          RMT~\cite{bulatov2022recurrent, bulatov2023scaling}{, } \\
                          State space model (SSM)~\cite{gu2022efficiently, dao2022hungry, orvieto2023resurrecting, gu2023mamba}{, }
                          RWKV~\cite{peng2023rwkv}{, }
                          RetNet~\cite{sun2023retentive},
                          leaf, text width=33em
                          ]
                          ]
                        ]
                        [
                          \textbf{Dynamic Neural Network}\\
                          \textbf{~~~~~~~~~~~($\S$\ref{subsec:dynamic-neural-network})}, fill=blue!10 
                          [
                          \textbf{Mixture-of-Expert ($\S$\ref{subsubsec:mixture-of-expert})}, fill=yellow!10
                          [
                          Switch transformer~\cite{fedus2022switch}{, }
                          V-MoE~\cite{riquelme2021scaling}{, }
                          GLaM~\cite{du2022glam}{, }
                          LIMoE~\cite{mustafa2022multimodal}{, }\\
                          Mistral~\cite{jiang2023mistral}{, }
                          FFF~\cite{belcak2023fast}
                          MoEfication~\cite{zhang2022moefication}{, }
                          Sparse Upcycling~\cite{komatsuzaki2023sparse},
                          leaf, text width=33em
                          ]
                          ]
                          [
                          \textbf{Early-exiting ($\S$\ref{subsubsec:early-exit-optimization})}, fill=yellow!10
                          [
                          Simplifying transformer blocks~\cite{he2023simplifying}{, }
                          FREE~\cite{bae2023fast}{, }
                          PABEE~\cite{zhou2020bert}{, }\\
                          DeeBERT~\cite{xin2020deebert}{, }
                          LGViT~\cite{xu2023lgvit}{, }
                          Bakhtiarnia et al.\cite{bakhtiarnia2021multi}{, }
                          SkipDecode~\cite{del2023skipdecode},
                          leaf, text width=33em
                          ]
                          ]
                        ] 
                        [
                          \textbf{Diffusion-specific}\\\textbf{Optimization ($\S$\ref{subsec:diffusion-optim})}, fill=blue!10 
                          [
                          \textbf{Efficient Sampling ($\S$\ref{subsubsec:efficient-sampling})}, fill=yellow!10
                          [
                          iDDPM~\cite{nichol2021improved}{, }
                          DDIM~\cite{song2020denoising}{, }
                          PNDM~\cite{liu2022pseudo}{, }
                          DPM-Solver~\cite{lu2022dpm}{, }
                          ReDi~\cite{zhang2023redi}{, }\\
                          Nirvana~\cite{agarwal2024approximate},
                          leaf, text width=33em
                          ]
                          ]
                          [
                          \textbf{Diffusion in Latent Space}\\
                          \textbf{~~~~~~~~~~~($\S$\ref{subsubsec:diffusion-in-latent-space})}, fill=yellow!10
                          [
                          LDM~\cite{rombach2022high}{, }
                          LD-ZNet~\cite{pnvr2023ld}{, }
                          SALAD~\cite{koo2023salad}{, }
                          Takagi et al.~\cite{takagi2023high},
                          leaf, text width=33em
                          ]
                          ]
                          [
                          \textbf{Diffusion Architecture}\\
                          \textbf{~~~~Variants ($\S$\ref{subsubsec:diffusion-arch-var})}
                          , fill=yellow!10
                          [
                          SnapFusion~\cite{li2023snapfusion}{, }
                          ERNIE-ViLG~\cite{feng2023ernie}{, }
                          ScaleCrafter~\cite{he2023scalecrafter},
                          leaf, text width=33em
                          ]
                          ]
                        ]  
                        [
                          \textbf{ViT-specific Optimizations}\\
                          \textbf{~~~~~~~~~~~($\S$\ref{subsec:ViT-specific Optimization})}, fill=blue!10 
                          [
                          LeViT~\cite{graham2021levit}{, } PoolFormer~\cite{yu2022metaformer}{, } MobileViT~\cite{mehta2022mobilevit}{, } EfficientFormer~\cite{li2022efficientformer}{, } 
                          EfficientViT~\cite{cai2022efficientvit},
                          leaf, text width=40em 
                          ]
                        ] 
                    ]
        \end{forest}
    }
    
    \caption{An overview of resource-efficient architectures.}
    \label{fig:tree-resource-efficient architecture tree}
\end{figure*}
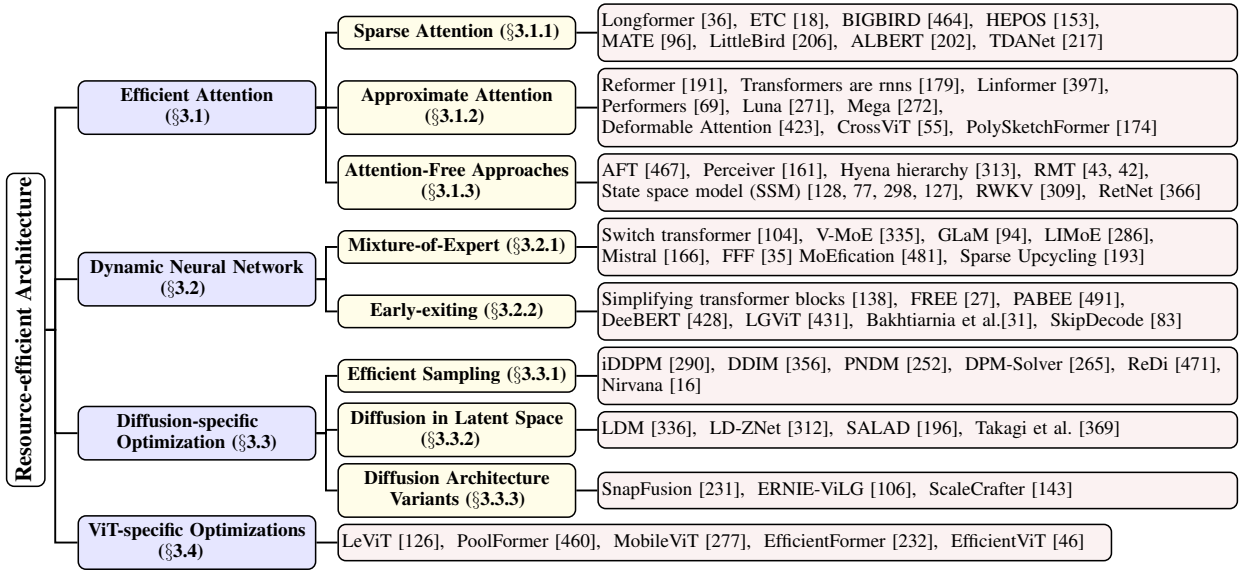


Model architecture is the core to resource-efficient large FMs, including attention mechanisms, decoders, and their alternatives. 
The primary objective is to reduce computational and memory expenses. 
Figure~\ref{fig:tree-resource-efficient architecture tree} visually illustrates this classification of resource-efficient architecture, considering the standard core blocks and the conventional taxonomy of large FMs. 
Resource-efficient architecture consists of efficient attention, dynamic neural network, diffusion-specific optimization, and ViT-specific optimization. 


\subsection{Efficient Attention}
\label{subsec:efficient-attention}

\begin{figure*}
	\centering
	\begin{minipage}[b]{0.195\textwidth}
		\centering
		\includegraphics[width=1\textwidth]{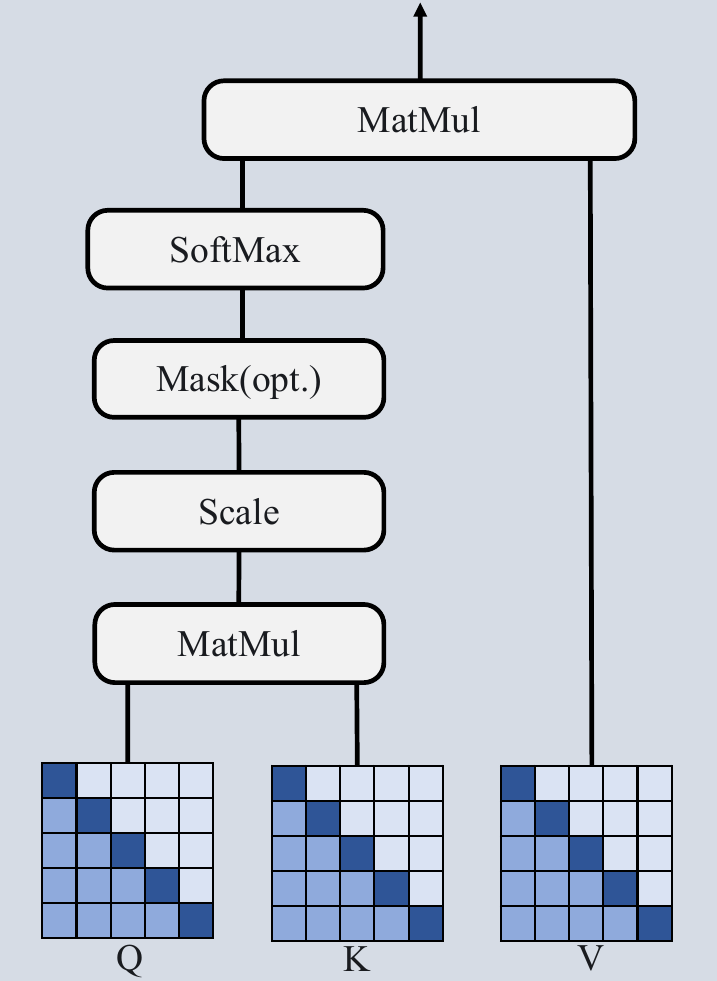}
		\subcaption{Self-Attention.}
		\label{fig:arch-attention-ori}
	\end{minipage}
	~
	\centering
	\begin{minipage}[b]{0.19\textwidth}
		\centering
		\includegraphics[width=1\textwidth]{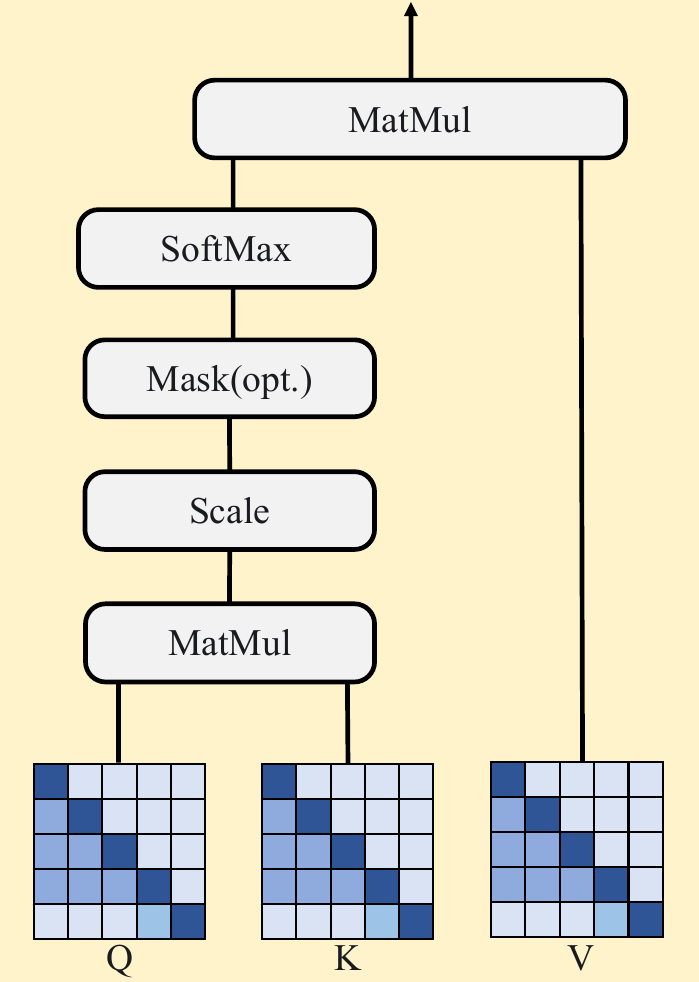}
		\subcaption{Sparse Attention.}
		\label{fig:arch-sparse-attention}
	\end{minipage}
	~
	\begin{minipage}[b]{0.275\textwidth}
		\centering
		\includegraphics[width=1\textwidth]{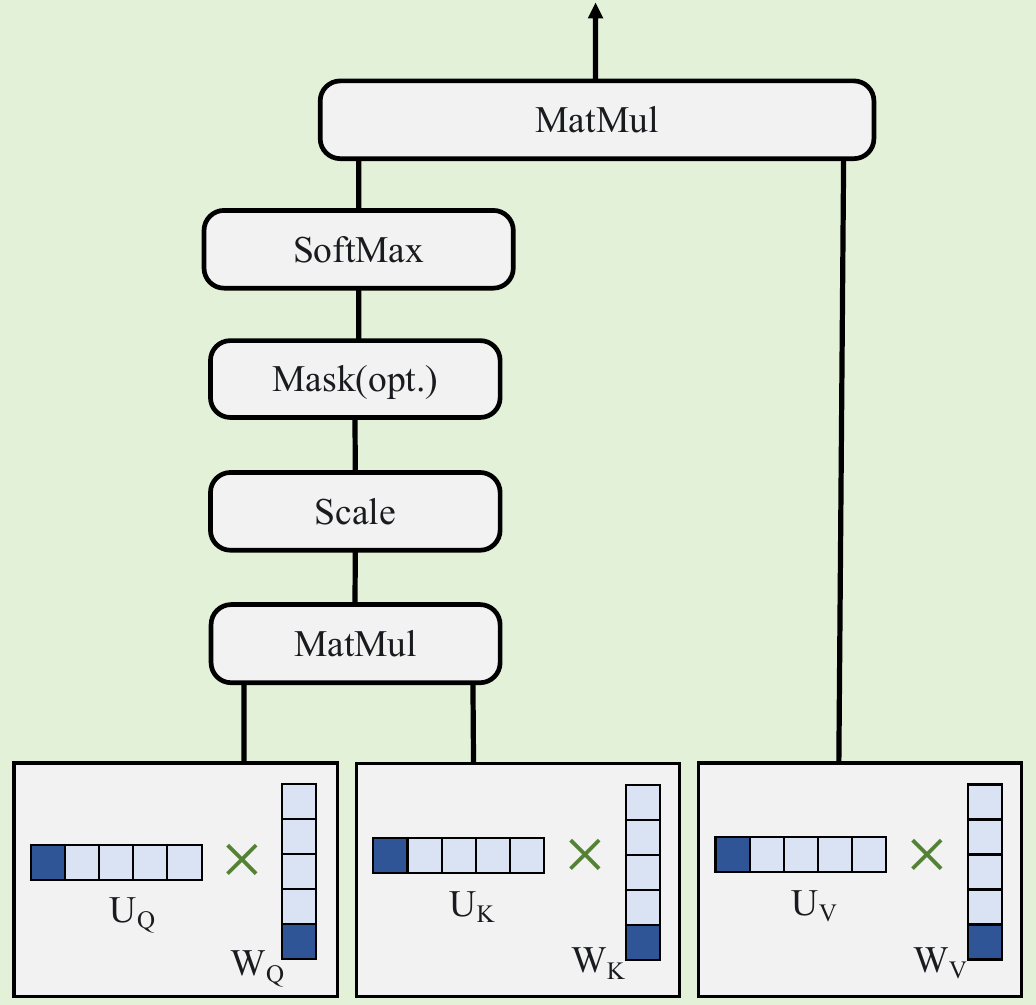}
		\subcaption{Approximate attention.}
		\label{fig:arch-approximation}
	\end{minipage}
	~
	\begin{minipage}[b]{0.283\textwidth}
		\centering
		\includegraphics[width=1\textwidth]{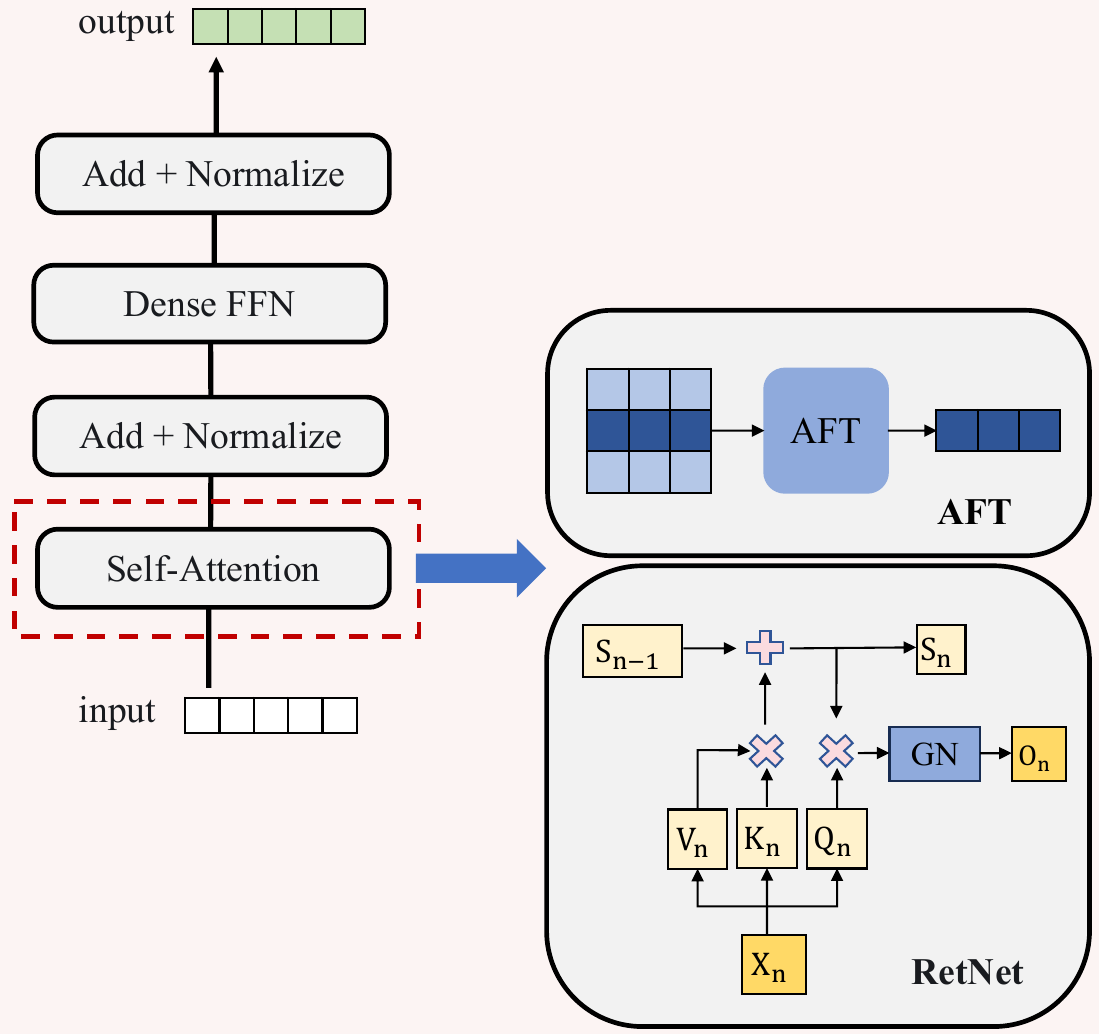}
		\subcaption{Attention-free.}
		\label{fig:arch-attention-free}
	\end{minipage}
	\caption{
		Illustrations of efficient attentions architectures. 
		}
	\label{fig:arch-attention}
\end{figure*}

The quadratic time complexity associated with attention architectures, particularly concerning sequence length, presents significant challenges during training and inference. Previous efforts~\cite{fournier2023practical, lin2022survey, kim2023full, ding2023efficiency, wan2023efficient} has explored methods to reduce this complexity to linear or identify viable alternatives. The diverse approaches for achieving efficient attention are visually summarized in Figure~\ref{fig:arch-attention}. 
\begin{table}[htbp]
    \centering
    \begin{tabular}{lll}
    \hline
    Model              & Time       & Space  \\ \hline
    Transformer~\cite{vaswani2017attention}       & $O(T^2d)$  & $O(T^2 + Td)$  \\ 
    AFT~\cite{zhai2021attention}           & $O(T^2d)$ & $O(Td)$  \\ 
    Reformer~\cite{kitaev2019reformer}           & $O(T \log Td)$ & $O(T \log T + Td)$ \\ 
    Hyena~\cite{poli2023hyena}      & $O(T \log T d)$   & $O(Td)$  \\
    SSM~\cite{gu2022efficiently}      & $O(T \log T d)$  & $O(Td)$  \\
    Linear Transformers~\cite{katharopoulos2020transformers} & $O(Td^2)$  & $O(Td + d^{2})$ \\ 
    RetNet~\cite{sun2023retentive}      & $O(Td)$   & $O(Td)$  \\ 
    RWKV~\cite{peng2023rwkv}      & $O(Td)$  & $O(d)$   \\ \hline
    \end{tabular}
    \caption{The time and space comparative analysis of complexity among transformers and its variants, where $T$ represents sequence length, and $d$ represents hidden dimension.}
    \label{tab:arch-attention-complexity}
    \end{table}

\subsubsection{Sparse Attention}
\label{subsubsec:sparse-attention}

Motivated by graph sparsification, sparse attention~\cite{beltagy2020longformer, ainslie2020etc, zaheer2020big, huang2021efficient, eisenschlos2021mate, lee2022littlebird, li2022efficient} aims to build a sparse attention matrix. This approach aims to retain the empirical advantages of a fully quadratic self-attention scheme while employing a reduced number of inner products.
For instance, Longformer \cite{fournier2023practical}, ETC~\cite{lin2022survey}, and BIGBIRD ~\cite{zaheer2020big} decompose conventional attention into local windowed attention and task-specific global attention, effectively reducing self-attention complexity to linear. 
HEPOS~\cite{huang2021efficient} introduces head-wise positional strides.
This innovation allows each attention head to concentrate on a specific subset of the input sequence, facilitating efficient encoder-decoder handling.
MATE~\cite{eisenschlos2021mate} transforms attention into a multi-view format, efficiently addressing either rows or columns in a table.
TDANet~\cite{li2022efficient} emulates the human brain's top-down attention mechanism to selectively focus on the most relevant information, thereby enhancing speech separation efficiency. 
ALBERT~\cite{lan2019albert} implements parameter sharing across layers, resulting in an 89\% reduction in parameter count while guaranteeing accuracy, compared to traditional BERT.

\subsubsection{Approximate Attention}
\label{subsubsec:approximate-attention}

Approximate attention, as explored in numerous works \cite{kitaev2019reformer, katharopoulos2020transformers, wang2020linformer, choromanski2020rethinking, ma2021luna, ma2022mega, kacham2023polysketchformer, chen2021crossvit, xia2022vision}, involves low-rank approximations of the self-attention matrix and innovative reformulations of the self-attention mechanism. 
These approaches avoid the direct computation of the $N\times N$ matrix, aiming to reduce computational complexity and enhance efficiency, particularly in scenarios with long sequence lengths.
Linformer \cite{wang2020linformer} demonstrates that the effective decomposition of the attention matrix into a low-rank matrix. This technique involves projecting the length dimensions of keys and values into a lower-dimensional space, resulting in a significant reduction in memory complexity. 
Reformer \cite{kitaev2019reformer} utilizes locality-sensitive hashing to replace the conventional dot-product attention.
Katharopoulos et al. \cite{katharopoulos2020transformers} introduced a kernel-based approach to self-attention, leveraging the associative property of matrix multiplication for computing self-attention weights.
Polysketchforme~\cite{kacham2023polysketchformer} employs polynomial functions and sketching techniques to approximate softmax attention outputs, providing a novel perspective on attention mechanism approximation.
Mega~\cite{ma2022mega}, featuring a single-head gated attention mechanism, incorporates exponential moving average. This addition effectively integrates inductive bias related to position-aware local dependencies into the inherently position-agnostic attention mechanism. 
Deformable Attention~\cite{xia2022vision} proposes a data-aware, deformable attention mechanism. , contributing to improved performance within the ViT architecture, This method proves especially beneficial when contrasted with traditional dense attention methods. 
CrossViT~\cite{chen2021crossvit} introduces linear cross-attention, empowering the ViT architecture to to efficiently handle variably-sized input tokens while mitigating computational costs.

\subsubsection{Attention-Free Approaches}
\label{subsubsec:attention-free}

Despite the dominance of attention-based transformer architectures in large FMs, several works~\cite{zhai2021attention, jaegle2021perceiver, poli2023hyena, bulatov2022recurrent, bulatov2023scaling, gu2022efficiently, dao2022hungry, orvieto2023resurrecting, peng2023rwkv, sun2023retentive} have put forth innovative architectures that hold the potential to replace the traditional transformer model. For instance,
Hyena~\cite{poli2023hyena} introduces an architecture that interleaves implicitly parametrized long convolutions with data-controlled gating. This design provides a subquadratic alternative to attention in large-scale language models, thereby enhancing efficiency in processing long sequences.
Another notable trend is the substitution of the attention mechanism with state space models (SSMs), as explored in \cite{gu2022efficiently, dao2022hungry, orvieto2023resurrecting}. 
Mamba~\cite{gu2023mamba} seamlessly integrates selective SSMs into a streamlined neural network architecture, eliminating attention and MLP blocks. This model achieves a notable 5$\times$ speed increase over traditional transformers and exhibits linear scaling with sequence length.
Gu et al. [119] also offered a comprehensive restructuring of the SSM literature into an informative and cohesive format.
Recurrent-Style Transformers (RMT)~\cite{bulatov2022recurrent, bulatov2023scaling} adopts an recurrent neural network-based architecture, replacing attention with a RNN to achieve linear complexity.
RWKV~\cite{peng2023rwkv} combines the efficient parallelizable training of Transformers with the effective inference capabilities of RNNs.
RetNet \cite{sun2023retentive} troduces an architecture that replaces multi-head attention with a multi-scale retention mechanism. This architecture captures and retains information from prior sequence steps, utilizing varying gamma values across heads to regulate retention strength. RetNet not only maintains a constant inference cost regardless of sequence length but also outperforms Transformer models with key-value caches in efficiency. 
Furthermore, during training, RetNet demonstrates a 25-50\% memory saving and a 7$\times$ acceleration compared to the standard Transformer.


\subsection{Dynamic Neural Network}
\label{subsec:dynamic-neural-network}


\subsubsection{Mixture-of-Experts}
\label{subsubsec:mixture-of-expert}

\begin{figure*}
	\centering
	\begin{minipage}[b]{\textwidth}
		\centering
		\includegraphics[width=0.95\textwidth]{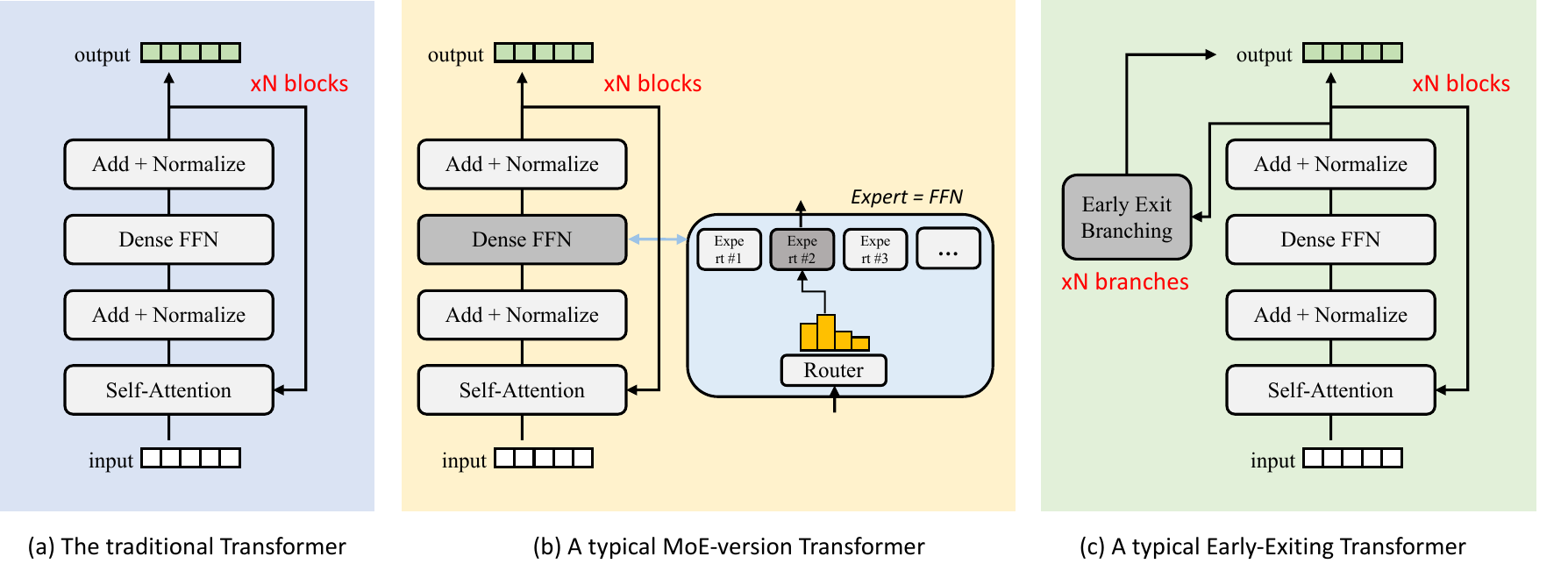}
		\caption{Traditional and typical dynamic transformers.}
		\label{fig:arch-moe}
	\end{minipage}
	\label{fig:arch-dynamic}
\end{figure*}

Mixture-of-Experts (MoE), illustrated in Figure~\ref{fig:arch-moe}(b), represents an efficient and sparse approach for training and deploying large FMs with extensive parameter sets. This model utilizes routed sparse parameters during inference. 
Switch Transformer~\cite{fedus2022switch} introduces a switch routing algorithm, leading to models with improved efficiency and reduced computational and communication costs. Switch Transformer demonstrates the scalability and effectiveness of MoE framework by managing up to one trillion parameters, even with as many as 2,048 experts.
GLaM~\cite{du2022glam}, a family of decoder-only language models, leverages a sparsely activated MoE design. This innovative approach substantially reduces training costs while simultaneously increasing model capacity compared to dense models.
V-MoE~\cite{riquelme2021scaling} presents a sparse adaptation of the ViT, scaling to 15 billion parameters, and achieves performance matching dense models while requiring less training time.
LIMoE~\cite{mustafa2022multimodal} represents the first multimodal model to incorporate sparse MoE, significantly outperforming CLIP in various tasks.
Mistral AI introduces Mistral\footnote{https://mistral.ai/}, an MoE model comprising 8 experts, each with 7 billion parameters. This model outperforms the performance of LLaMA2-70B model \cite{touvron2023llama}.
MoEfication \cite{zhang2022moefication} converts a model into its MoE variant with equivalent parameters. This technique conditionally utilizes portions of feedforward network parameters, maintaining performance levels comparable to the original dense model.
Sparse upcycling~\cite{komatsuzaki2023sparse} initializes sparsely activated MoE from dense checkpoints.
Sparse upcycling enables models, such as T5 Base, Large, XL, and Vision Transformer Base and Large, to significantly outperform their dense counterparts on benchmarks like SuperGLUE and ImageNet, while using only about 50\% of the original dense pretraining costs.
FFF~\cite{belcak2023fast} divides the feed-forward layer into separate leaves instead of copying the entire feed-forward layer as an expert. 
FFF is a log-time alternative to feedforward networks, being up to 220$\times$ faster than the original feed-forward layer and up to 64$\times$ faster than MoE with about 5\% accuracy loss.
$\S$\ref{subsec:distributed-training-systems} will detail systematic optimizations applied to MoE models.

\subsubsection{Early Exiting}
\label{subsubsec:early-exit-optimization}

As illustrated in Figure~\ref{fig:arch-moe}(c), early exiting optimization is a strategy that allows a model to terminate its computational process prematurely when it attains high confidence in the prediction or encounters resource constraints.
He et al. \cite{he2023simplifying} investigates modifications to the standard transformer block, aiming for simpler yet efficient architectures without sacrificing performance. This model involves removing elements such as residual connections, layer normalization, and specific parameters in projection and value, along with serializing MLP sub-blocks.
M4 \cite{yuan2023rethinking} introduces a multi-path task execution framework, enabling elastic fine-tuning and execution of foundational model blocks for different training and inference tasks.
FREE \cite{bae2023fast} proposes a shallow-deep module that synchronizes the decoding of the current token with previously processed early-exit tokens. FREE utilizes an adaptive threshold estimator for determining appropriate early-exit confidence levels.
SkipDecode \cite{del2023skipdecode} is designed for batch inferencing and KV caching, overcoming previous limitations by establishing a unique exit point for each token in a batch at every sequence position.
PABEE \cite{zhou2020bert} enhances the efficiency of pre-trained language models by integrating internal classifiers at each layer. The inference process halts when predictions stabilize for a set number of steps, facilitating quicker predictions with reduced layer usage.
DeeBERT \cite{xin2020deebert} augments BERT's inference efficiency by incorporating early exit points. DeeBERT allows instances to terminate at intermediate layers based on confidence levels, effectively reducing computational demands and accelerating inference.
Bakhtiarnia et al. \cite{bakhtiarnia2021multi} proposed 7 distinct architectural designs for early-exit branches suitable for dynamic inference in ViTs backbones.
LGViT \cite{xu2023lgvit} presents an early-exiting framework tailored for general ViTs, featuring diverse exiting heads, such as local perception and global aggregation heads, to balance efficiency and accuracy.
This approach achieves competitive performance with an approximate 1.8$\times$ increase in speed.



\subsection{Diffusion-specific Optimization}
\label{subsec:diffusion-optim}

Generating images through diffusion models typically involves iterative process with numerous denoising steps. Recent research has focused on accelerating the denoising process and reducing the resource requirements during image generation, which fall into three main categories: (1) efficient sampling, (2) diffusion in latent space, and (3) diffusion architecture variants.


\subsubsection{Efficient Sampling}
\label{subsubsec:efficient-sampling}

To enhance the denoising process of diffusion model while maintaining or improving sample quality, many efforts~\cite{nichol2021improved, lyu2022accelerating, song2020denoising, zhang2022gddim, liu2022pseudo, karras2022elucidating, lu2022dpm, salimans2022progressive, zhang2022fast, song2020score, watson2022learning,agarwal2024approximate,zhang2023redi} have been made to improve the sampling process. These works emphasize resource and time efficiency in their architectures.
Nichol et al.~\cite{nichol2021improved} made strides in enhancing the traditional DDPM by focusing on resource efficiency. Their improved model not only competes in log-likelihoods but also enhances sample quality. This efficiency is achieved by learning the variances of the reverse diffusion process and employing a hybrid training objective. This methodology leads to more resource-efficient sampling, requiring fewer forward passes, and shows improved scalability in terms of model capacity and computational power.
DDIM~\cite{song2020denoising} represents a significant improvement in latency efficiency for diffusion models. By introducing a non-Markovian, deterministic approach to sampling, DDIM accelerates the generation process, allowing for faster sampling without compromising sample quality. This ordinary differential equations (ODE) variant of DDPM~\cite{ho2020denoising} efficiently navigates through noise levels during generation, making it a more latency-efficient option compared to traditional DDPM.
PNDM~\cite{liu2022pseudo} enhances the efficiency of DDPM in generating high-quality samples. The approach treats the diffusion process as solving differential equations on manifolds, greatly accelerating the inference process. This method, with a focus on latency efficiency, surpasses existing acceleration techniques such as DDIMs, particularly in scenarios that require high-speed sampling, all while maintaining sample fidelity.
DPM-Solver~\cite{lu2022dpm} focuses on improving the sampling efficiency of diffusion models. This method utilizes a high-order solver that exploits the semi-linear structure of diffusion ODEs, 
facilitating fast and high-quality sample generation. Remarkably, DPM-Solver achieves this with as few as 10-20 denoising steps, highlighting the latency efficiency in sample generation.
\update{
ReDi~\cite{zhang2023redi} accelerates sampling by retrieving and reusing pre-computed diffusion trajectories.
Nirvana~\cite{agarwal2024approximate} introduces an approximate caching technique to reduce computational cost and latency in text-to-image diffusion models by reusing intermediate noise states.
This leads to significant GPU and time savings without compromising image quality.
}

\subsubsection{Diffusion in Latent Space}
\label{subsubsec:diffusion-in-latent-space}

In traditional diffusion models, operations are usually performed within the pixel space of images. However, this approach proves to be inefficient for high-resolution images because of the considerable computational demands and significant memory requirements.
In response to these challenges, researchers proposed a shift towards conducting diffusion processes in latent space through VAEs~\cite{rombach2022high, koo2023salad, wang2023binary, pnvr2023ld, schramowski2023safe, takagi2023high, barquero2023belfusion, ma2023unified}. This paradigm results in substantial memory-efficient advancements, allowing for the generation of high-resolution images with reduced computational resources.
LDM~\cite{rombach2022high}, also known as stable diffusion, serves as a notable example of memory-efficient image generation. By performing diffusion processes within a latent space derived from pixel data through a VAE, LDM effectively tackles scalability issues present in earlier diffusion models. This approach enables the efficient synthesis of high-resolution images and, with the incorporation of cross-attention layers interacting with text encoder inputs, facilitating the creation of detailed, text-guided visuals.
LD-ZNet~\cite{pnvr2023ld} leverages the memory-efficient properties of LDM for image segmentation tasks. By training segmentation models within the latent space of LDM, LD-ZNet significantly enhances segmentation accuracy, especially for AI-generated images. This approach capitalizes on the deep semantic understanding inherent in LDM’s internal features, providing a nuanced bridge between real and AI-generated imagery.
SALAD~\cite{koo2023salad} introduces a memory-efficient methodology for 3D shape generation and manipulation. Employing a cascaded diffusion model based on part-level implicit representations, SALAD operates in high-dimensional spaces with a two-phase diffusion process. This memory-efficient design facilitates the effective generation and part-level editing of 3D shapes without the need for additional training in conditional setups.
Takagi et al.~\cite{takagi2023high} presented a novel approach to reconstruct high-resolution images from Functional Magnetic Resonance Imaging (fMRI) brain activity using LDM. The method utilizes straightforward linear mappings to project fMRI data into the latent space of LDMs. This method not only accomplishes high-fidelity image reconstruction from brain activity but also provides a deeper understanding of LDM's internal mechanisms from a neuroscience perspective.
Each of these works highlights the advantages of leveraging latent spaces in diffusion models, particularly in terms of memory efficiency, marking a clear distinction from traditional pixel-space approaches.

\subsubsection{Diffusion Architecture Variants}
\label{subsubsec:diffusion-arch-var}

Another method for enhancing diffusion models involves the adoption of more efficient model architectures~\cite{li2023snapfusion, feng2023ernie, balaji2022ediffi, he2023scalecrafter, luo2023image}. This strategy focuses on refining the structural framework of diffusion models to optimize their performance. Through the implementation of advanced architectural designs, these models can achieve more effective processing capabilities while potentially reducing computational complexity and resource consumption.
SnapFusion~\cite{li2023snapfusion} introduces an optimized text-to-image diffusion model for mobile devices, featuring a resource-efficient network architecture. This model overcomes the computational and latency limitations of existing models through a redesigned network architecture and improved step distillation. By generating high-quality 512$\times$512 images in under 2 seconds, SnapFusion surpasses Stable Diffusion in FID and CLIP scores. Notably, this achievement is realized with fewer denoising steps.
ScaleCrafter~\cite{he2023scalecrafter} addresses the generation of ultra-high-resolution images using pre-trained diffusion models with an innovative and resource-efficient network design. ScaleCrafter incorporates techniques like “re-dilation”, “dispersed convolution”, and “noise-damped classifier-free guidance” to dynamically adjust convolutional perception fields during inference. This model enables the generation of ultra-high-resolution images without the need for additional training or optimization.
ERNIE-ViLG~\cite{feng2023ernie} introduces a novel text-to-image diffusion model that integrates fine-grained textual and visual knowledge into a highly efficient network architecture. With a mixture-of-denoising-experts mechanism and scaling up to 24B parameters, ERNIE-ViLG outperforms the existing models on MS-COCO with a remarkable zero-shot FID-30k score of 6.75. The model significantly enhances image fidelity and text relevancy, adeptly addressing attribute confusion in complex text prompts. 
Each of these contributions highlights the significance of resource-efficient network architectures in advancing diffusion models, expanding their capabilities, and broadening their applicability across diverse scenarios.

\subsection{ViT-specific Optimizations}
\label{subsec:ViT-specific Optimization}

\begin{figure*}
	\centering
	\includegraphics[width=0.38\textwidth]{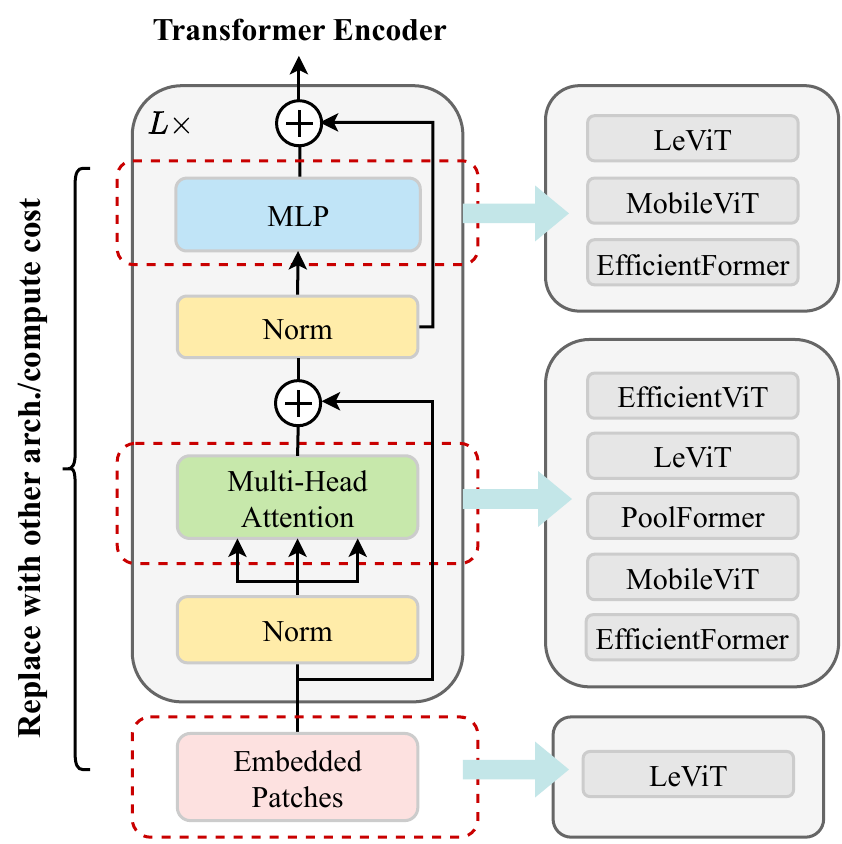}
	\caption{
		A summary of resource-efficient VIT variants.
		}
	\label{fig:arch-vit-specific}
\end{figure*}

As a Transformer variant, ViT benefits from general optimizations aforementioned;
yet, there exists also ViT-specific architecture optimizations.
LeViT~\cite{graham2021levit} is a hybrid neural network designed for efficient image classification during inference. LeViT utilizes an increased number of convolutional layers for embedding, enhancing its ability to process pixel information. The main backbone features a pyramid architecture, progressively reducing the dimensionality of features while concurrently increasing the number of attention heads. Notably, the model introduces a learned, per-head translation-invariant attention bias, serving as a replacement for the positional embedding in ViT. LeViT achieves an impressive 80\% ImageNet top-1 accuracy and demonstrates a remarkable 5$\times$ improvement in speed compared to EfficientNet when executed on CPU~\cite{tan2020efficientnet}.
PoolFormer~\cite{yu2022metaformer} provides valuable insights by emphasizing that the success of ViTs is rooted in its overall architectural design named MetaFormer.
PoolFormer employs simple pooling layers commonly found in CNNs and achieves an impressive 82.1\% top-1 accuracy on the ImageNet-1K dataset. 
MobileViT~\cite{mehta2022mobilevit} adheres to the idea of utilizing CNNs to construct a more lightweight transformer architecture.
Through the design of a convolution-like MobileViT block, the model achieves a lightweight and low-latency implementation, specifically tailored for practical hardware platforms. MobileViT optimization extends beyond FLOPs to include considerations for memory access, parallelism, and platform-specific features.
MobileViT attains a top-1 accuracy of 78.4\% on the ImageNet-1k dataset with approximately 6 million parameters. 
EfficientFormer~\cite{li2022efficientformer} designs a lightweight CNN-Transformer hybrid architecture, achieving more efficient on-device inference. The most rapid model, EfficientFormer-L1, attains a top-1 accuracy of 79.2\% on ImageNet-1K, with a mere 1.6 ms inference latency on the iPhone 12 (compiled with CoreML). This performance matches the speed of MobileNetV2×1.4 (1.6ms, 74.7\% top-1). 
EfficientViT~\cite{cai2023efficient} introduces a linear attention mechanism to alleviate the computational cost linked with the high overhead of softmax in non-linear attention. In the domain of super-resolution, EfficientViT achieves a speedup of up to 6.4 $\times$ compared to Restormer~\cite{zamir2022restormer}.

	\section{RESOURCE-EFFICIENT ALGORITHMS}
\label{sec:algorithms}
\tikzstyle{my-box}=[
    rectangle,
    draw=black,
    rounded corners,
    text opacity=1,
    minimum height=1.5em,
    minimum width=5em,
    inner sep=2pt,
    align=center,
    fill opacity=.5,
    line width=0.8pt,
]
\tikzstyle{leaf}=[my-box, minimum height=1.5em,
    fill=hidden-red!10, text=black, align=left,font=\normalsize,
    inner xsep=2pt,
    inner ysep=4pt,
    line width=0.8pt,
]

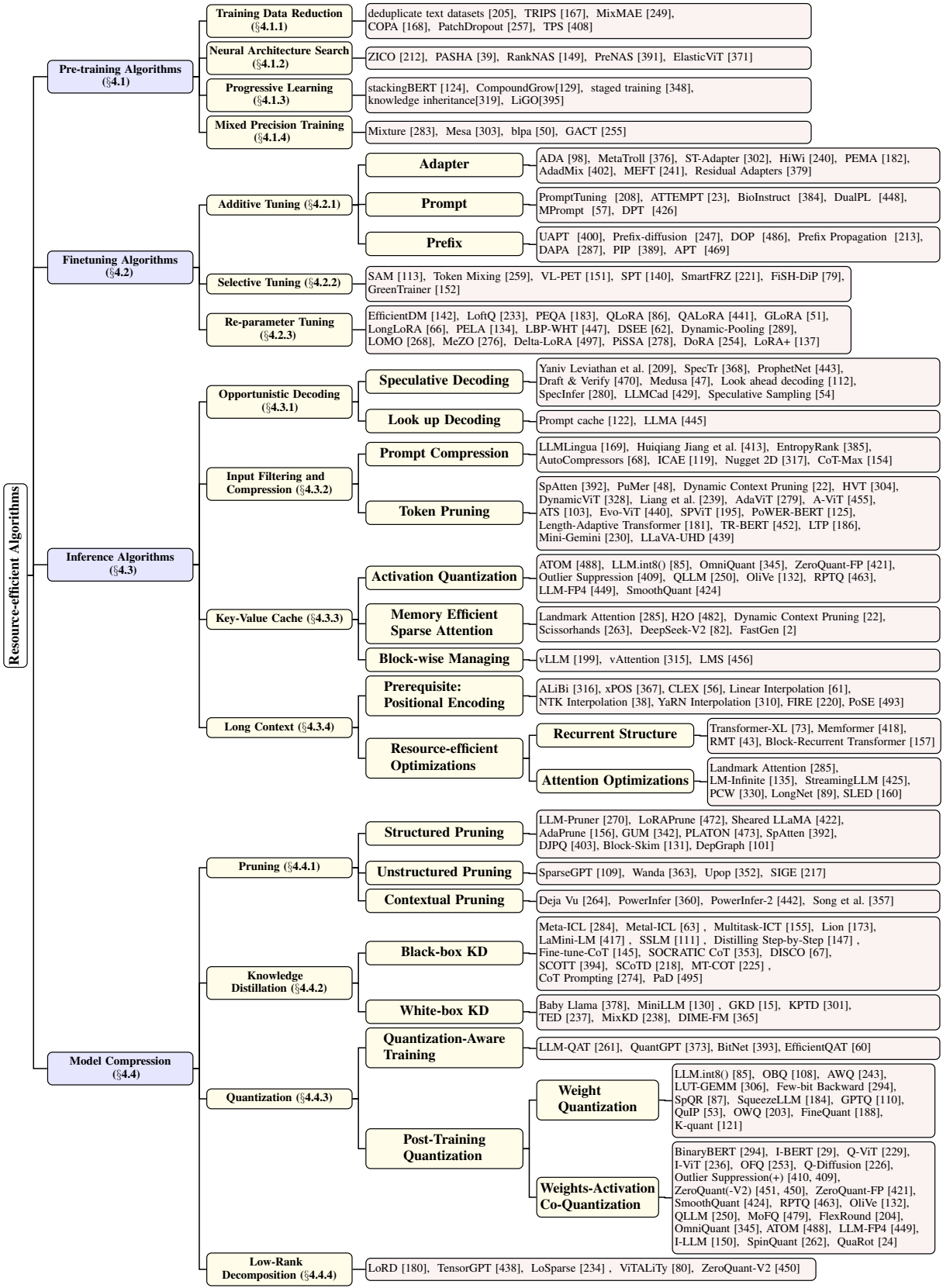
\begin{figure*}[htpb]
    \centering
    \resizebox{0.98\textwidth}{!}{
        \begin{forest}
            forked edges,
            for tree={
                grow=east,
                reversed=true,
                anchor=base west,
                parent anchor=east,
                child anchor=west,
                base=center,
                font=\large,
                rectangle,
                draw=black,
                rounded corners,
                align=left,
                text centered,
                minimum width=4em,
                edge+={black, line width=1pt},
                s sep=3pt,
                inner xsep=2pt,
                inner ysep=3pt,
                line width=0.8pt,
                ver/.style={rotate=90, child anchor=north, parent anchor=south, anchor=center},
            },
            where level=1{text width=12em,font=\normalsize,}{},
            where level=2{text width=12em,font=\normalsize,}{},
            [
                \textbf{Resource-efficient Algorithms}, ver
                [
                  \textbf{Pre-training Algorithms}\\
                  \textbf{~~~~~~~~~~~($\S$\ref{subsec:Pre-training-algorithms})}, fill=blue!10
                  [
                    \textbf{Training Data Reduction}\\
                    \textbf{~~~~~~~~~($\S$\ref{subsubsec:training-data-reduction})}, fill=yellow!10
                    [
                      deduplicate text datasets~\cite{lee2022deduplicating}{, }
                      TRIPS~\cite{jiang2022trips}{, }
                      MixMAE~\cite{liu2023mixmae}{, } \\
                      COPA~\cite{jiang2023copa}{, }
                      PatchDropout~\cite{liu2023patchdropout}{, }
                      TPS~\cite{wei2023joint},
                      leaf, text width=37.6em
                    ]
                  ]
                  [
                    \textbf{Neural Architecture Search}\\
                    \textbf{~~~~~~~~~~~($\S$\ref{subsubsec:neural-arch-search})}, fill=yellow!10
                    [
                      ZICO~\cite{li2022zico}{, }
                      PASHA~\cite{bohdal2022pasha}{, }
                      RankNAS~\cite{hu2021ranknas}{, }
                      PreNAS~\cite{wang2023prenas}{, }
                      ElasticViT~\cite{tang2023elasticvit},
                      leaf, text width=37.6em
                    ]
                  ]
                  [
                    \textbf{Progressive Learning}\\
                    \textbf{~~~~~~~($\S$\ref{subsubsec:progressive-learning})}, fill=yellow!10
                    [
                      stackingBERT~\cite{gong2019efficient}{, }
                      CompoundGrow\cite{gu2021progressivebert}{, }
                      staged training~\cite{shen2022staged}{, } \\
                      knowledge inheritance\cite{qin2022knowledge}{, }
                      LiGO\cite{wang2022learning},
                      leaf, text width=37.6em
                    ]
                  ]
                  [
                    \textbf{Mixed Precision Training}\\
                    \textbf{~~~~~~~~~~($\S$\ref{subsubsec:mixed-precision-training})}, fill=yellow!10
                    [
                      Mixture~\cite{micikevicius2018mixed}{, }
                      Mesa~\cite{pan2021mesa}{, }
                      blpa~\cite{chakrabarti2019backprop}{, }
                      GACT~\cite{liu2022gact},
                      leaf, text width=37.6em
                    ]
                  ]F
                ]
                [
                \textbf{Finetuning Algorithms}\\
                \textbf{~~~~~~~~~~~($\S$\ref{subsec:finetuning-alg})}, fill=blue!10  
                        [
                            \textbf{Additive Tuning ($\S$\ref{subsubsec:additive-tuning})}, fill=yellow!10
                            [
                            \textbf{Adapter}, fill=yellow!10, text width=13em
                            [
                            ADA~\cite{ermis2022memory}{, }
                            MetaTroll~\cite{tian2023metatroll}{, }
                            ST-Adapter~\cite{pan2022st}{, }
                            HiWi~\cite{liao2023parameter}{, }
                            PEMA~\cite{kim2023pema}{, }\\
                            AdadMix~\cite{wang2022adamix}{, }
                            MEFT~\cite{liao2023make}{, }
                            Residual Adapters~\cite{tomanek2021residual}, 
                            leaf, text width=34em
                            ]
                            ]
                            [
                            \textbf{Prompt}, fill=yellow!10, text width=13em
                            [
                            PromptTuning ~\cite{lester2021power}{, }
                            ATTEMPT~\cite{asai2022attempt}{, }
                            BioInstruct ~\cite{tran2023bioinstruct}{, }
                            DualPL ~\cite{yang2023dual}{, }\\
                            MPrompt ~\cite{chen2023mprompt}{, }
                            DPT ~\cite{xiao2023decomposed}, 
                            leaf, text width=34em
                            ]
                            ]
                            [
                            \textbf{Prefix}, fill=yellow!10, text width=13em
                            [
                            UAPT ~\cite{wang2023user}{, }
                            Prefix-diffusion ~\cite{liu2023prefix}{, }
                            DOP ~\cite{zhao2022domain}{, }
                            Prefix Propagation ~\cite{li2023prefix}{, } \\
                            DAPA ~\cite{nair2023domain}{, }
                            PIP ~\cite{wan2023pip}{, }
                            APT ~\cite{zhang2023towards},
                            leaf, text width=34em
                            ]
                            ]
                        ]
                        [
                          \textbf{Selective Tuning ($\S$\ref{subsubsec:selective-tuning})}, fill=yellow!10
                          [
                            SAM~\cite{fu2023effectiveness}{, }
                            Token Mixing~\cite{liu2023token}{, }
                            VL-PET~\cite{hu2023vl}{, }
                            SPT~\cite{he2023sensitivity}{, }
                            SmartFRZ~\cite{li2022smartfrz}{, }
                            FiSH-DiP~\cite{das2023unified}{, }\\
                            GreenTrainer~\cite{huang2023towards},
                          leaf, 
                          text width=41em
                          ]
                        ]
                        [
                          \textbf{Re-parameter Tuning}\\
                          \textbf{~~~~~~~~~($\S$\ref{sec:alg:finetune:reparam})}, fill=yellow!10
                          [
                            EfficientDM~\cite{he2023efficientdm}{, }
                            LoftQ~\cite{li2023loftq}{, }
                            PEQA~\cite{kim2023memory}{, }
                            QLoRA~\cite{dettmers2023qlora}{, }
                            QALoRA~\cite{xu2023qa}{, }
                            GLoRA~\cite{chavan2023one}{, }\\
                            LongLoRA~\cite{chen2023longlora}{, }
                            PELA~\cite{guo2023learning}{, }
                            LBP-WHT~\cite{yang2023efficient}{, }
                            DSEE~\cite{chen2021dsee}{, }
                            Dynamic-Pooling~\cite{nawrot2022efficient}{, }\\
                            LOMO~\cite{lv2023full}{, }
                            MeZO~\cite{malladi2023fine}{, }
                            Delta-LoRA~\cite{zi_delta-lora_2023}{, }
                            PiSSA~\cite{meng_pissa_2024}{, }
                            DoRA~\cite{liu_dora_2024}{, }
                            LoRA+~\cite{hayou_lora_nodate},
                          leaf, text width=41em
                          ]
                        ] 
                    ]
                [
					\textbf{Inference Algorithms}\\
     \textbf{~~~~~~~~~~~($\S$\ref{subsec:inference-alg})}, fill=blue!10
          [
            \textbf{Opportunistic Decoding}\\
            \textbf{~~~~~~~~~~~($\S$\ref{subsubsec:opportunistic-decoding})}, fill=yellow!10
            [
                \textbf{Speculative Decoding}, fill=yellow!10, text width=13em
                [
                Yaniv Leviathan et al.~\cite{Leviathan2022FastIF}{, }
                SpecTr~\cite{sun2023spectr}{, }
                ProphetNet~\cite{Yan2020ProphetNetPF}{, } \\
                Draft \& Verify~\cite{Zhang2023DraftV}{, }
                Medusa~\cite{medusa}{, }
                Look ahead decoding~\cite{fu2023lookahead}{, } \\
                SpecInfer~\cite{miao2023specinfer}{, }
                LLMCad~\cite{xu2023llmcad}{, }
                Speculative Sampling~\cite{Chen2023AcceleratingLL},
                leaf, text width=34em
                ]
            ]
            [
                \textbf{Look up Decoding}, fill=yellow!10, text width=13em
                [
                Prompt cache~\cite{gim2023prompt}{, }
                LLMA~\cite{yang2023inference},
                leaf, text width=34em
                ]
            ]
        ]
        [
            \textbf{Input Filtering and}\\
            \textbf{Compression ($\S$\ref{subsubsec:input-filtering-and-compression})}, fill=yellow!10
            [
                \textbf{Prompt Compression}, fill=yellow!10, text width=13em
                [
                LLMLingua~\cite{jiang2023llmlingua}{, }
                Huiqiang Jiang et al.~\cite{wingate2022prompt}{, }
                EntropyRank~\cite{tsvetkov2023entropyrank}{, } \\
                AutoCompressors~\cite{chevalier2023adapting}{, }
                ICAE~\cite{ge2023incontext}{, }
                Nugget 2D~\cite{qin2023nugget}{, }
                CoT-Max~\cite{huang2023boosting},
                leaf, text width=34em
                ]
            ]
            [
                \textbf{Token Pruning}, fill=yellow!10, text width=13em
                [
                SpAtten~\cite{wang2021spatten}{, }
                PuMer~\cite{cao2023pumer}{, }
                Dynamic Context Pruning~\cite{anagnostidis2023dynamic}{, }
                HVT~\cite{pan2021scalable}{, }\\
                DynamicViT~\cite{rao2021dynamicvit}{, }
                Liang et al.~\cite{liang2022patches}{, }
                AdaViT~\cite{meng2021adavit}{, }
                A-ViT~\cite{yin2022adavit}{, }\\
                ATS~\cite{fayyaz2022adaptive}{, }
                Evo-ViT~\cite{xu2021evovit}{, }
                SPViT~\cite{kong2022spvit}{, }
                PoWER-BERT~\cite{goyal2020powerbert}{, }\\
                Length-Adaptive Transformer~\cite{kim2021lengthadaptive}{, }
                TR-BERT~\cite{ye2021trbert}{, }
                LTP~\cite{kim2022learned}{, }\\
                Mini-Gemini~\cite{li2024minigemini}{, }
                LLaVA-UHD~\cite{xu2024llava},
                leaf, text width=34em
                ]
            ]
        ]
        [
            \textbf{Key-Value Cache ($\S$\ref{subsec:kv-cache})}, fill=yellow!10
            [
                \textbf{Activation Quantization}, fill=yellow!10, text width=13em
                [
                ATOM~\cite{zhao2023atom}{, }
                LLM.int8()~\cite{dettmers2022llm}{, }
                OmniQuant~\cite{shao2023omniquant}{, }
                ZeroQuant-FP~\cite{wu2023zeroquant}{, }\\
                Outlier Suppression~\cite{wei2023outlier}{, }
                QLLM~\cite{liu2023qllm}{, }
                OliVe~\cite{guo2023olive}{, }
                RPTQ~\cite{yuan2023rptq}{, }\\
                LLM-FP4~\cite{liu2023llmfp4}{, }
                SmoothQuant~\cite{xiao2023smoothquant},
                leaf, text width=34em
                ]
            ]
            [
                \textbf{Memory Efficient}\\ 
                \textbf{Sparse Attention}, fill=yellow!10, text width=13em
                [
                Landmark Attention~\cite{mohtashami2023landmark}{,}
                H2O~\cite{zhang2023h2o}{, }
                Dynamic Context Pruning~\cite{anagnostidis2023dynamic}{, }\\
                Scissorhands~\cite{liu2023scissorhands}{, }
                DeepSeek-V2~\cite{deepseekai2024deepseekv2strongeconomicalefficient}{, }
                FastGen~\cite{deepspeed-fastgen},
                leaf, text width=34em
                ]
            ]
            [
                \textbf{Block-wise Managing}, fill=yellow!10, text width=13em
                [
                vLLM~\cite{kwon2023efficient}{, }
                vAttention~\cite{prabhu2024vattentiondynamicmemorymanagement}{, }
                LMS~\cite{yin2024llmservicemobiledevices},
                leaf, text width=34em
                ]
            ]
        ]
        [
            \textbf{Long Context ($\S$\ref{subsubsec:long-context})}, fill=yellow!10
            [
                \textbf{Prerequisite:}\\ \textbf{Positional Encoding}, fill=yellow!10, text width=13em
                [
                ALiBi~\cite{press2021train}{,}
                xPOS~\cite{sun2022length}{,}
                CLEX~\cite{chen2023clex}{,}
                Linear Interpolation~\cite{chen2023extending}{,}
                \\NTK Interpolation~\cite{SketchyRedditPost2023}{,}
                YaRN Interpolation~\cite{peng2023yarn}{,}
                FIRE~\cite{DBLP:journals/corr/abs-2310-04418}{,}
                PoSE~\cite{zhu2023pose}
                , leaf, text width=34em
                ]
            ]
            [
                \textbf{Resource-efficient}\\ 
                \textbf{Optimizations}, fill=yellow!10, text width=13em
                [
                    \textbf{Recurrent Structure}, fill=yellow!10, text width=13em
                    [
                    Transformer-XL~\cite{dai2019transformer}{,}
                    Memformer~\cite{wu2020memformer}{,}\\
                    RMT~\cite{bulatov2022recurrent}{,}
                    Block-Recurrent Transformer~\cite{hutchins2022block}
                    , leaf, text width=19.5em 
                    ]
                ]
                [
                    \textbf{Attention Optimizations}, fill=yellow!10, text width=13em
                    [
                    Landmark Attention~\cite{mohtashami2023landmark}{, }\\
                    LM-Infinite~\cite{han2023lminfinite}{, }
                    StreamingLLM~\cite{xiao2023efficient}{,}\\
                    PCW~\cite{ratner2023parallel}{,}
                    LongNet~\cite{ding2023longnet}{,}
                    SLED~\cite{ivgi2023efficient}
                    , leaf, text width=19.5em
                    ]
                ]
            ]
        ]
                ]
                [
                \textbf{Model Compression}\\
                \textbf{~~~~~~~~~~~($\S$\ref{subsec:model-compression})}, fill=blue!10 
                        [
                          \textbf{Pruning ($\S$\ref{subsubsec:pruning})}, fill=yellow!10
                          [
                          \textbf{Structured Pruning}, fill=yellow!10, text width=13em
                          [                          
                          LLM-Pruner~\cite{ma2023llm}{, }
                          LoRAPrune~\cite{zhang2023pruning}{,}
                          Sheared LLaMA~\cite{xia2023sheared}{,}\\
                          AdaPrune~\cite{hubara2021accelerated}{,}
                          GUM~\cite{santacroce2023matters}{,} 
                          PLATON~\cite{zhang2022platon}{,}
                          SpAtten~\cite{wang2021spatten}{,}\\
                          DJPQ~\cite{wang2020differentiable}{,}
                          Block-Skim~\cite{guan2022block}{,}
                          DepGraph~\cite{fang2023depgraph}, 
                          leaf, text width=34em
                          ]
                          ]
                          [
                          \textbf{Unstructured Pruning}, fill=yellow!10, text width=13em
                          [
                          SparseGPT~\cite{frantar2023sparsegpt}{, }
                          Wanda~\cite{sun2023simple}{, }
                          Upop~\cite{shi2023upop}{, }
                          SIGE~\cite{li2022efficient},
                          leaf, text width=34em
                          ]
                          ]
                          [
                          \textbf{Contextual Pruning}, fill=yellow!10, text width=13em
                          [
                          Deja Vu~\cite{liu2023deja}{, }
                          PowerInfer~\cite{song2023powerinfer}{, }
                          PowerInfer-2~\cite{xue2024powerinfer}{, }
                          Song et al.~\cite{song2024achieving},
                          leaf, text width=34em
                          ]
                          ]
                        ]
                        [
                          \textbf{~~~~Knowledge}\\\textbf{Distillation ($\S$\ref{subsubsec:know-distil})}, fill=yellow!10 
                          [
                          \textbf{Black-box KD}, fill=yellow!10, text width=13em
                          [
                            Meta-ICL~\cite{min2021metaicl}{, }
                            Metal-ICL~\cite{chen2021meta} {, }
                            Multitask-ICT~\cite{huang2022context}{, }
                            Lion~\cite{jiang2023lion}{, }\\
                            LaMini-LM~\cite{wu2023lamini} {, } 
                            SSLM~\cite{fu2023specializing} {, }
                            Distilling Step-by-Step~\cite{hsieh2023distilling} {, }\\
                            Fine-tune-CoT~\cite{ho2022large}{, }
                            SOCRATIC CoT~\cite{shridhar2023distilling}{, } 
                            DISCO~\cite{chen2023disco}{, } \\
                            SCOTT~\cite{wang2023scott}{, }
                            SCoTD~\cite{li2023symbolic}{, }
                            MT-COT~\cite{li2022explanations} {, } \\
                            CoT Prompting~\cite{magister2022teaching}{, }
                            PaD~\cite{zhu2023pad},
                          leaf, text width=34em
                          ]
                          ]
                          [
                          \textbf{White-box KD}, fill=yellow!10, text width=13em
                          [
                            Baby Llama~\cite{timiryasov2023baby}{, }
                            MiniLLM~\cite{gu2023knowledge} {, }
                            GKD~\cite{agarwal2023gkd}{, }
                            KPTD~\cite{padmanabhan2023propagating}{, }\\
                            TED~\cite{liang2023less}{, } 
                            MixKD~\cite{liang2020mixkd}{, }
                            DIME-FM~\cite{sun2023dime},
                          leaf, text width=34em
                          ]
                          ]
                        ] 
                        [
                          \textbf{Quantization ($\S$\ref{subsubsec:quantization})}, fill=yellow!10 
                          [
                          \textbf{Quantization-Aware}\\\textbf{Training}, fill=yellow!10, text width=13em
                         [
                          LLM-QAT~\cite{liu2023llm}{, }
                          QuantGPT~\cite{tao2022compression}{,}
                          BitNet~\cite{wang2023bitnet}{,}
                          EfficientQAT~\cite{chen2024efficientqat},
                          leaf, text width=34em
                          ]
                          ]
                          [
                          \textbf{Post-Training}\\\textbf{Quantization}, fill=yellow!10, text width=13em
                          [
                          \textbf{Weight }\\\textbf{Quantization}, fill=yellow!10, text width=10em
                          [
                          LLM.int8()~\cite{dettmers2022llm}{, }
                          OBQ~\cite{frantar2022optimal}{, } 
                          AWQ~\cite{lin2023awq}{, } \\
                          LUT-GEMM~\cite{park2023lut}{, } 
                          Few-bit Backward~\cite{novikov2023few}{, } \\
                          SpQR~\cite{dettmers2023spqr}{, }
                          SqueezeLLM~\cite{kim2023squeezellm}{, }
                          GPTQ~\cite{frantar2022gptq}{, }\\
                          QuIP~\cite{chee2023quip}{, }
                          OWQ~\cite{lee2023owq}{, } 
                          FineQuant~\cite{kim2023finequant}{,} \\
                          K-quant~\cite{llamacpp},
                          leaf, text width=22em
                          ]
                          ]
                          [
                        \textbf{Weights-Activation}\\\textbf{Co-Quantization}, fill=yellow!10, text width=10em
                          [
                          BinaryBERT~\cite{novikov2023few}{, } 
                          I-BERT~\cite{bai2020binarybert}{, } 
                          Q-ViT~\cite{li2022q}{, }\\
                          I-ViT~\cite{li2023vit}{, }  
                          OFQ~\cite{liu2023oscillation}{, } 
                          Q-Diffusion~\cite{li2023q}{, }  \\
                          Outlier Suppression(+)~\cite{wei2022outlier,wei2023outlier}{, } \\
                          ZeroQuant(-V2)~\cite{yao2022zeroquant,yao2023zeroquant}{, }
                          ZeroQuant-FP~\cite{wu2023zeroquant}{, }\\
                          SmoothQuant~\cite{xiao2023smoothquant}{, }
                          RPTQ~\cite{yuan2023rptq}{, } 
                          OliVe~\cite{guo2023olive}{, }\\
                          QLLM~\cite{liu2023qllm}{, }
                          MoFQ~\cite{zhang2023integer}{, } 
                          FlexRound~\cite{lee2023flexround}{,}\\
                          OmniQuant~\cite{shao2023omniquant}{,}
                          ATOM~\cite{zhao2023atom}{, }
                          LLM-FP4~\cite{liu2023llmfp4}{,}\\
                          I-LLM~\cite{hu2024llm}{, }
                          SpinQuant~\cite{liu2024spinquant}{, }
                          QuaRot~\cite{ashkboos2024quarot}, 
                          leaf, text width=22em
                          ]
                          ]
                          ]
                        ] 
                        [
                          \textbf{~~~~~~~Low-Rank }\\\textbf{Decomposition ($\S$\ref{subsubsec:low-rank-decomposition})}, fill=yellow!10 
                          [
                            LoRD~\cite{kaushal2023lord}{, }
                            TensorGPT~\cite{xu2023tensorgpt}{, }
                            LoSparse~\cite{li2023losparse} {, }
                            ViTALiTy~\cite{dass2023vitality}{, }
                            ZeroQuant-V2~\cite{yao2023zeroquant},
                          leaf, text width=38em 
                          ]
                        ] 
                    ]
            ]
        \end{forest}
    }
    \caption{An overview of resource-efficient algorithms.}
    \label{fig:alg-overview}
\end{figure*}


This section focuses on resoruce-efficient large FMs techniques at the algorithm level. 
Compared to traditional DNNs, large FMs exhibit new characteristics such as its huge parameter set and autoregressive inference.
This disparity has led to the emergence of numerous resource-efficient algorithms, which are categorized based on the lifecycle of FMs: pre-training, fine-tuning, serving algorithms, and model compression as illustrated in Figure 11.

\subsection{Pre-training Algorithms}
\label{subsec:Pre-training-algorithms}
Pre-training for large FMs relies on a substantial amount of computation resources.
For instance, GPT-3-175B~\cite{brown2020language} consumes $3.14\times 10^{23}$ flops and LLaMa-70B~\cite{touvron2023llama} takes $1.7\times 10^{6}$ GPU hours.
Consequently, optimizing the utilization of computational resources is crucial for the efficient pre-training of FMs.
Resource-efficient algorithms can be categorized into training data deduction, neural architecture search, progressive learning, and mixed precision training.

\subsubsection{Training Data Reduction}
\label{subsubsec:training-data-reduction}
Pre-training for large FMs needs a dataset at the trillion scale, exemplified by 0.3 trillion tokens for GPT-3-175B ~\cite{brown2020language} and 2 trillion tokens for LLaMa-2-70B~\cite{touvron2023llama}.
More data indicates more resource expenditure.
Thereby, prior literature resorts to reducing the resource cost on vast training data through two aspects: deduplicate text datasets and image patch removal.

\textbf{Deduplicate text datasets}~\cite{lee2022deduplicating} shows training data has redundancy caused by near-duplicate examples and long repetitive substrings.
The reduction of repetitions can lead to fewer training steps without compromising performance.

\textbf{Image patch removal} is achieved by either reducing the number of patch inputs to the model or reorganizing image tokens based on modified model architectures.
For instance, TRIPS~\cite{jiang2022trips} employs a patch selection layer to reduce image patches.
This layer computes attentive image tokens through text guidance, resulting in a 40\% reduction in computation resources, compared to previous pre-training vision-language models. 
Masked autoencoders (MAE)~\cite{he2022masked} as a pre-training method of ViT has been widely used.
MAE mask image patches in pre-training phrase, but the large masking ratio brings significant computation resource wastage.
MixMAE~\cite{liu2023mixmae} introduces a method for mixing multiple images at the patch level, thereby avoiding the need for introducing [MASK] symbols. The use of a visible image patch in lieu of [MASK] symbols contributes to a reduction in the size of the training dataset.
COPA~\cite{jiang2023copa} introduces an auxiliary pre-training task called patch-text alignment. This patch-level alignment strategy aims to decrease redundancy in image patches, and the reduction in patch sequences contributes to saving computation resources.
PatchDropout~\cite{liu2023patchdropout} introduces the concept of patch dropout to enhance both computation and memory efficiency. This method involves the random sampling of a subset of original image patches to effectively shorten the length of token sequences. TPS~\cite{wei2023joint} employs a more aggressive compression technique. TPS categorizes tokens into pruned and reserved sets based on their importance and subsequently assigns each pruned token to its associated reserved token instead of discarding the pruned set.

\subsubsection{Neural Architecture Search}
\label{subsubsec:neural-arch-search}
Neural Architecture Search (NAS) is an automatic model design algorithm for optimal model efficiency and performance without human intervention. 
Zero-shot NAS~\cite{abdelfattah2020zero} introduces a concept where models predict the performance of neural network architectures without the need for actual training. By employing prior knowledge and surrogate models for architecture evaluation, this technique significantly reduces computational costs and time, facilitating efficient exploration of complex architecture spaces with minimal resource expenditure.
ZICO~\cite{li2022zico} introduces a zero-shot NAS proxy that outperforms traditional NAS proxies, including one-shot and multi-shot NAS methods, in terms of performance. Notably, it reduces computing resource consumption by eliminating the need to train the model during the search phase.
PASHA~\cite{bohdal2022pasha} allocates limited resources for initialization and increases the resource allocation as needed.
With dynamic resource allocation, PASHA speeds up NAS and efficiently manages computation resources.
RankNAS~\cite{hu2021ranknas} formulates the NAS problem as a ranking problem as a ranking problem and further simplifies this ranking problem to a binary classification problem. Additionally, the method evaluates promising architectures using features.
PreNAS~\cite{wang2023prenas} employs a search-free NAS approach. 
This approach identifies the preferred architectures by zero-shot NAS proxy. 
Then it only trains once on the preferred architectures.
ElasticViT~\cite{tang2023elasticvit} has a NAS approach to automatically design lightweight ViT models with fewer than $1 \times 10^{9}$ FLOPs.
The method introduces two subnet sampling techniques to address gradient conflicts, leading to high accuracy and low inference latency.

\subsubsection{Progressive Learning}
\label{subsubsec:progressive-learning}
Progressive learning is a training strategy that begins by training a small model and then gradually increases the model size, continuing the training process. This approach optimizes computational resource usage by reusing the computations from the previous stage.
Inspired by the insight that knowledge can be shared across models of different depths, stackingBERT~\cite{gong2019efficient} introduces a progressive stacking algorithm. This algorithm trains a large model by sequentially stacking attention layers from smaller models. StackingBERT demonstrates that this progressive stacking approach can achieve similar performance to training a large model from scratch but with reduced computational resource consumption.
CompoundGrow~\cite{gu2021progressivebert} identifies the similarity between progressive training algorithms and NAS. CompoundGrow introduces a progressive training algorithm for BERT that facilitates the growth of multiple dimensions of the model, including depth, width, and input length. The method mitigates computational resource consumption by employing compound growth operators.
Staged training~\cite{shen2022staged} adopts a strategy where a small model is pre-trained initially, and subsequently, the depth and width of the model are increased, continuing the training process. The entirety of the training state, encompassing model parameters, optimizer state, learning rate schedule, etc., is transferred to the next stage using a growth operator. This approach reuses computations from the previous stage, effectively reducing both training time and computational resource requirements.
Knowledge inheritance~\cite{qin2022knowledge} suggests employing existing pre-trained language models as teacher models to provide guidance during the training of larger models. The supplementary auxiliary supervision offered by the teacher model can effectively enhance the training speed of the larger model.
LiGO~\cite{wang2022learning} introduces small model parameters to initialize the large model through a trainable parameter linear map. LiGO achieves this by factorizing the growing transformation into a composition of linear operators at width and depth dimensions.

\subsubsection{Mixed Precision Training}
\label{subsubsec:mixed-precision-training}
Mixed precision training often utilizes half-precision floating-point data representation instead of single precision. This approach significantly reduces memory requirements, approximately halving the storage space needed for weights, activations, and gradients in half-precision.
Mesa~\cite{pan2021mesa} proposes the combination of activation compressed training~\cite{chakrabarti2019backprop} with mixed precision training to further reduce the memory used by activations. The method quantifies activation based on the distribution of multi-head self-attention layers to minimize the approximation error.
GACT~\cite{liu2022gact} introduces a dynamically adjusted compression ratio based on the importance of each gradient. This innovation allows for the reduction of compute resource and memory consumption in various model architectures, including convolutional neural networks, graph neural networks, and transformer-based models.

\subsection{Finetuning Algorithms}
\label{subsec:finetuning-alg}



















Efficient fine-tuning algorithms are designed to reduce the workload to adapt a pre-trained FM to downstream tasks. These techniques can be categorized into three groups: additive tuning, selective tuning, and re-parameter tuning.

\subsubsection{Additive Tuning}
\label{subsubsec:additive-tuning}
Large FMs can achieve high performance with low costs by incorporating additional parameters and fine-tuning them for new tasks.
In particular, this additive tuning process in large FMs can be categorized into three main classes: adapter tuning, prompt tuning, and prefix tuning.

\textbf{Adapter tuning.}
Adapter tuning aims to reduce training costs by introducing adapter modules to specific layers (or all layers) of pre-trained large FMs.
During tuning, the backbone of the pre-trained model remains frozen, and Adapter modules are utilized to acquire task-specific knowledge.
Some works~\cite{ermis2022memory, tian2023metatroll, pan2022st} focus on designing adapters for multi-task or multi-modal extensions. 
ADA~\cite{ermis2022memory} and MetaTroll~\cite{tian2023metatroll} concentrate on incrementally extending pre-trained Transformers' capabilities across multiple tasks. This approach helps alleviate catastrophic forgetting during learning while simultaneously reducing computational expenses.
ST-Adapter~\cite{pan2022st} introduces built-in spatiotemporal reasoning abilities, allowing pre-trained models to significantly reduce the number of parameters that need to be updated in cross-modal tasks.
Other works~\cite{liao2023parameter, wang2022adamix, tomanek2021residual, liao2023make} aim to further decrease fine-tuning costs.
HiWi~\cite{liao2023parameter} demonstrates that the PEFT method, which involves adding new parameters, often introduces additional inference latency. PEFT improves inference speed by applying adapters to pre-trained parameters rather than hidden representations.
AdaMix~\cite{wang2022adamix} designs a combined mechanism that merges the weights of different Adapters into a single adapter at each Transformer layer. This innovation significantly reduces the additional storage cost introduced by multiple adapters.
MEFT~\cite{liao2023make} designs a method for inserting adapters into LLM by modifying LLM to its reversible variant, reducing activation memory and thus improving the memory efficiency of fine-tuning.
Residual Adapters~\cite{tomanek2021residual} addresses the issue of performance degradation in Automatic Speech Recognition caused by non-standard speech. The approach involves designing personalized residual adapters, which contribute to a reduction in the number of updated parameters.
Moreover, other works such as PEMA~\cite{kim2023pema} focus on ensuring data confidentiality during the fine-tuning of models. These approaches involve designing adapters that provide contextual prompts, allowing pre-trained models to generate corresponding context representations. Additionally, tuning is performed using a Gradual Unrolling approach.

\begin{figure*}
	\centering
	\includegraphics[width=0.74\textwidth]{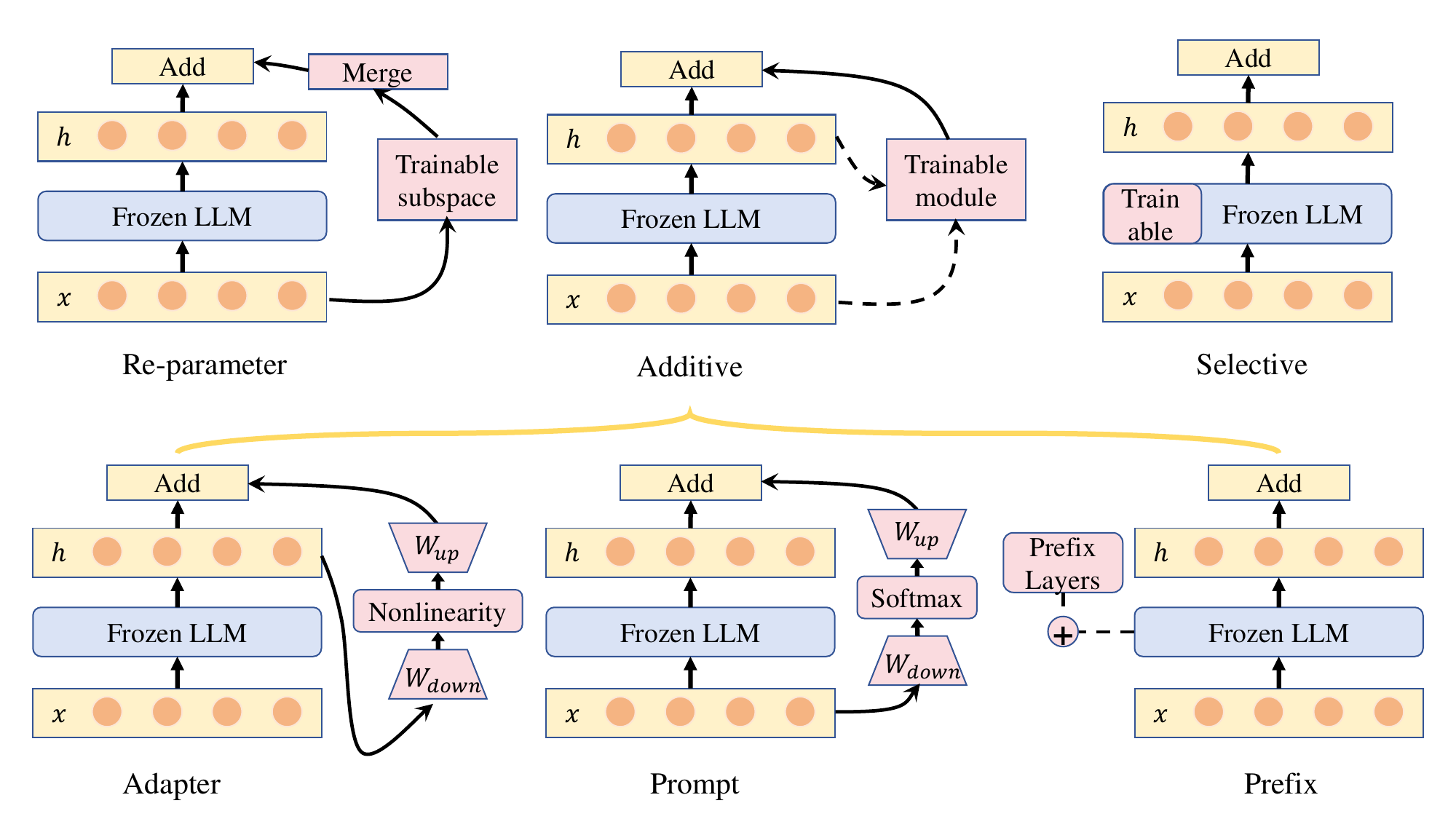}
	\caption{
		A summary of various fine-tuning algorithms. 
		}
	\label{fig:resource-efficient-finetuing-arch}
\end{figure*}

\textbf{Prompt tuning.}
Prompt tuning involves designing a task-specific prompt for each task, with the aim of replacing the traditional fine-tuning of pre-trained large FM parameters.
By tuning the input prompts instead, this method significantly reduces the resources and time required for fine-tuning.
Some works~\cite{asai2022attempt, lester2021power, tran2023bioinstruct} focus on improving the efficient scalability of prompts in multi-task settings. 
For example, PromptTuning~\cite{lester2021power}, ATTEMPT~\cite{asai2022attempt}, and BioInstruct~\cite{tran2023bioinstruct} investigate how the utilization of mixed soft prompts can efficiently transfer knowledge across different tasks. These approaches help mitigate parameter update costs by reusing the frozen pre-trained large model.
Furthermore, some works~\cite{yang2023dual, chen2023mprompt} focus on minimizing prompt fine-tuning costs for specific tasks. For instance, DualPL~\cite{yang2023dual} designs two prompts and separately captures the relevant knowledge of both tasks. This approach addresses the high cost associated with collecting state labels for slots and values in dialogue state tracking systems. 
In machine reading comprehension tasks, MPrompt~\cite{chen2023mprompt}introduces task-specific multi-level prompt tuning to enhance the understanding of input semantics at different granularities while reducing the number of parameter updates.
Additionally, DPT~\cite{xiao2023decomposed} addresses the issue of low efficiency in prompt tuning arising from the use of the same initialization. This approach explores the decomposition of initialization prompts, further reducing trainable parameters while ensuring prompt effectiveness.

\textbf{Prefix tuning.}
Prefix tuning introduces a trainable, task-specific prefix part to each layer of large FMs.
This technique aims to reduce the tuning cost by limiting the updates to the parameters in this prefix.
Some works~\cite{wang2023user, liu2023prefix, zhao2022domain, nair2023domain, wan2023pip} focus on enhancing the performance of prefix tuning in specific domains.
For example, UAPT~\cite{wang2023user} and Prefix-diffusion~\cite{liu2023prefix} address the issue of limited diversity in generating captions for images. These approaches extract image features from large FMs and design prefixes to enhance performance while reducing additional overhead.  
DOP~\cite{zhao2022domain} and DAPA~\cite{nair2023domain} concentrate on domain-generalization problems in abstract summarization.
These approaches design prefixes for each source domain to improve the model's generalization capabilities.
PIP~\cite{wan2023pip} focuses on syntactic control in paraphrase generation and reduces training costs by designing parsing-indicating prefixes. Additionally, other works such as Prefix Propagation~\cite{li2023prefix} and APT~\cite{zhang2023towards} further optimize prefix tuning to enhance their efficiency and effectiveness.

\subsubsection{Selective Tuning}
\label{subsubsec:selective-tuning}
Selective tuning aims to maintain high performance on new tasks with low training costs by freezing the majority of parameters in large FMs and selectively updating only a small portion of the parameters.
Some works focus on optimizing the performance of selective tuning. For example, SAM~\cite{fu2023effectiveness} explores how the choice of tunable parameters affects tuning. By proposing a second-order approximation method, it tunes fewer parameters to achieve better model performance. 
SmartFRZ~\cite{li2022smartfrz} focuses on improving the efficiency of layer freezing by introducing an adaptive layer freezing technique based on different network structures. This innovation enhances system accuracy and training speed. 
FiSH-DiP~\cite{das2023unified} explores the effectiveness of tuning with limited data by introducing a sample-aware dynamic sparse tuning strategy. This approach selectively tunes partial parameters using sample feedback to enhance the model's generalization in resource-constrained situations.
Additionally, some works continue to concentrate on specific domains.
For instance, Token Mixing~\cite{liu2023token} and VL-PET~\cite{hu2023vl} focus on visual-language tasks, enhancing fine-tuning efficiency by adjusting and selecting a subset of trainable parameters.
SPT~\cite{he2023sensitivity} emphasizes the efficient tuning of visual parameters by designing sensitivity-aware parameter budgets, enabling selective tuning for specific tasks.
\update{
From the perspective of sustainable AI computing, GreenTrainer~\cite{huang2023towards} minimizes the FLOPs of LLM fine-tuning by adaptively selecting the most appropriate set of tensors, as guided by their importance and backpropagation cost.
It demonstrates remarkable results compared to full fine-tuning, achieving up to a 64\% reduction in fine-tuning FLOPs.
}

\subsubsection{Re-parameter Tuning}
\label{sec:alg:finetune:reparam}
Re-parameter tuning adapts large FMs by targeting a significantly smaller subspace than the original, expansive training space. 
This approach involves fine-tuning low-rank matrix parameters, a technique that effectively reduces the overall training cost.
The majority of existing research centers on reparameterization tuning through the implementation of the low-rank adapter design.
For example, EfficientDM~\cite{he2023efficientdm}, QLoRA~\cite{dettmers2023qlora}, PEQA~\cite{kim2023memory}, QALoRA~\cite{xu2023qa} and LoftQ~\cite{li2023loftq} incorporate quantization techniques, building upon the foundation of LoRA, to enhance memory efficiency. This collaborative effort aims to improve training efficiency while maintaining model performance.
GLoRA~\cite{chavan2023one} enhances LoRA's generality, improving model transferability, few-shot capabilities, and domain generalization with a reduced number of parameters and computational overhead.
PELA~\cite{guo2023learning} derives inspiration from LoRA and devises a low-rank approximation compression method. This innovation enhances fine-tuning efficiency by exclusively updating the low-rank model while keeping all backbone parameters frozen.
LongLoRA~\cite{chen2023longlora} extends the capabilities of LoRA by incorporating context expansion through shift short attention. This implementation achieves computational cost savings while preserving model performance.
Moreover, LOMO~\cite{lv2023full} and MeZO~\cite{malladi2023fine} optimize the gradient computation process to reduce the memory requirements of gradient tensors during the fine-tuning process.
For ViT’s linear layers, LBP-WHT~\cite{yang2023efficient} diminishes the computational costs of matrix multiplication by employing low-rank backward propagation based on the Walsh-Hadamard transform. This approach contributes to the simultaneous enhancement of model performance.
Additionally, DSEE~\cite{chen2021dsee} investigates the application of sparse-aware low-rank updates on pre-trained model weights. This methodology substantially reduces inference FLOPs while ensuring an enhancement in model performance.
Dynamic-Pooling~\cite{nawrot2022efficient} mechanisms are designed to predict inference boundaries through autoregressive prediction. This design guides the model to re-parameterize more rapidly during tuning.

\update{
\begin{figure*}
	\centering
	\includegraphics[width=0.98\textwidth]{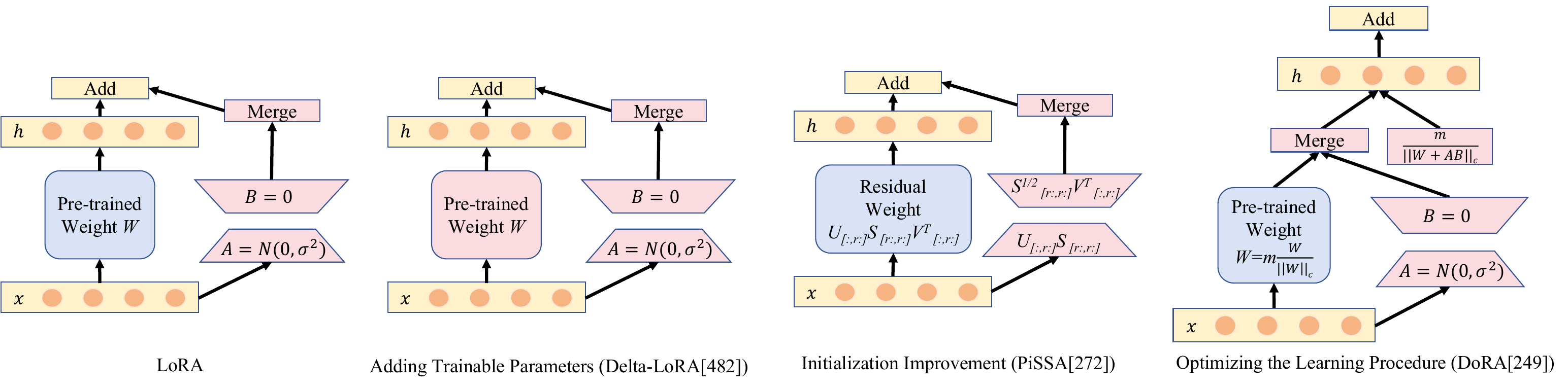}
	\caption{
		\update{LoRA and its optimization methods.} 
		}
	\label{fig:resource-efficient-finetuing-lora}
\end{figure*}
LoRA freezes the pre-trained weight matrix while fine-tuning the low-rank matrices A and B, which can have ranks as low as 2. This approach significantly reduces the number of trainable parameters, consequently decreasing the amount of gradient values required for storage during training. However, LoRA still exhibits performance gaps when compared to full fine-tuning. To address this, various methods have been developed to enhance LoRA's performance, as Figure~\ref{fig:resource-efficient-finetuing-lora}, mainly categorized into three strategies: Adding Trainable Parameters, Initialization Improvement, and Optimizing the Learning Procedure.
Delta-LoRA~\cite{zi_delta-lora_2023} aims to bridge the performance gap by updating the pre-trained weights through the product of low-rank matrices A and B, thus adding trainable parameters without incurring additional memory overhead. On the other hand, PiSSA~\cite{meng_pissa_2024} identifies an issue where LoRA initializes low-rank matrices with Gaussian random values and zeros, resulting in very small initial gradient values and slow convergence. PiSSA proposes using Singular Value Decomposition (SVD) on the pre-trained weight matrix, freezing the matrix with smaller singular values as a residual matrix, and initializing the trainable low-rank matrices A and B with the largest singular values. This optimized initialization method yields faster convergence and improved performance.
Lastly, DoRA~\cite{liu_dora_2024} and LoRA+~\cite{hayou_lora_nodate} focus on enhancing the learning process itself to further improve efficiency and effectiveness.DoRA observes that LoRA and full fine-tuning exhibit different update patterns. Therefore, it decomposes the pre-trained weights into their magnitude and directional components, and fine-tunes the directional matrix. This brings the update pattern closer to that of full fine-tuning, leading to better training outcomes. LoRA+ demonstrates that using the same learning rate for matrices A and B is not optimal for training. By setting the learning rate for B higher than that for A, it can accelerate convergence and improve fine-tuning performance.}

\subsection{Inference Algorithms}
\label{subsec:inference-alg}


\subsubsection{Opportunistic Decoding}
\label{subsubsec:opportunistic-decoding}

The nature of autoregressive severely hinders the inference efficiency of large FMs.
Thereby, there are extensive approaches aiming to substitute autoregressive decoding with opportunistic non-autoregressive decoding.

\textbf{Speculative decoding.}
Speculative decoding involves the autoregressive generation of sequences using a cost-efficient small model, followed by the parallel verification of each token using a larger model.
The typical speculative decoding process is as follows.
Given a small model $M_{q}$ and a large model $M_{p}$, decode $\gamma$ tokens by $M_{p}$ in an autoregressive manner.
Append $\gamma$ to the prefix $\sigma$, then execute a forward pass by $M_{p}$.
Compare the logits at the corresponding token positions between $M_{q}$ and $M_{p}$, and accept or reject these tokens based on a certain criterion.
Notably, this process ensures complete accuracy, and the final decoding results are identical to those achieved using only $M_{p}$.
Yaniv et al.~\cite{Leviathan2022FastIF} achieves 2-3$\times$ improvement using speculative decoding on the T5X model.
A concurrent work~\cite{Chen2023AcceleratingLL} also demonstrates a similar speedup on a 70B Chinchilla model.
SpecTr~\cite{sun2023spectr} proposes a novel speculative decoding strategy to enhance algorithm performance by extending the number of candidate tokens and improving the draft selection method.
The suggested method results in a 2.13$\times$ improvement in wall clock speed, further outperforming speculative decoding by an additional 1.37$\times$ on standard benchmarks.
ProphetNet~\cite{Yan2020ProphetNetPF} introduces a sequence modeling architecture that predicts the future tokens, reducing autoregression to some extent.
In the draft stage, Draft \& Verify~\cite{Zhang2023DraftV} selectively skips certain intermediate layers without depending on an additional small model, rendering it more plug-and-play. 
Testing on Llama-2 indicates that this method achieves a speedup of 1.73$\times$.
Medusa~\cite{medusa} is another decoding architecture designed for non-autoregressive decoding, and doesn't necessitate an additional auxiliary model. It predicts multiple tokens simultaneously by pre-training multiple heads to predict different time steps and then verifies these tokens concurrently.
Look-ahead decoding~\cite{fu2023lookahead} accelerates inference in large FMs without requiring a draft model or a data store. This method linearly reduces the number of decoding steps relative to log(FLOPs) used per decoding step. Additionally, there are inference systems constructed on speculative decoding, including SpecInfer~\cite{miao2023specinfer}, which utilizes multiple draft models in the cloud, and LLMCad~\cite{xu2023llmcad}, deployed at the edge.

\textbf{Look up decoding.}
Another approach involves searching for patterns in a text corpus during the decoding process as a substitute for autoregressive generation. For instance, speculating that the currently decoded content might correspond to a sequence in the text corpus and then verifying in parallel. By precomputing and storing the attention states of frequently occurring text segments on the inference server, Prompt Cache~\cite{gim2023prompt} can efficiently reuse them when these segments appear in user prompts. The enhancements vary from an 8$\times$ improvement for GPU-based inference to a 60$\times$ improvement for CPU-based inference, all while preserving output accuracy and without requiring modifications to model parameters.
During the inference phase, LLMA~\cite{yang2023inference} leverages external text for verifying potential inference results and integrates a triggering mechanism to determine when to decode and when to reference. The increased computational parallelism allows LLMA to achieve a speedup of over 2$\times$ for large FMs, producing identical generation results to greedy decoding in numerous practical scenarios characterized by substantial overlap between in-context reference and outputs, such as search engines and multi-turn conversations.
In addition, there is research dedicated to providing parallelizable decoding prompts for large FMs, thereby mitigating the impact of autoregressive decoding. For instance, inspired by the cognitive processes of human thinking and writing, Skeleton-of-Thoughts~\cite{ning2023skeletonofthought} initially directs large FMs to produce the framework of an answer. Subsequently, this method employs parallel API calls or batched decoding to concurrently fill in the details for each point in the generated skeleton.

\subsubsection{Input Filtering and Compression}
\label{subsubsec:input-filtering-and-compression}

\textbf{Prompt compression.}
Computations can be effectively reduced by compressing the prompt to the model.
LLMLingua~\cite{jiang2023llmlingua} introduces a prompt compression approach from a coarse-to-fine perspective. This approach incorporates a budget controller to maintain semantic integrity at high compression ratios, a token-level iterative compression algorithm to improve the modeling of interdependence among compressed contents, and a method based on instruction tuning for aligning distributions between language models. LLMLingua can achieve up to 20$\times$ compression with minimal performance loss.
Jiang et al.~\cite{wingate2022prompt}investigated the feasibility, applicability, and potential of compressing natural language for large FMs while preserving semantics.
EntropyRank~\cite{tsvetkov2023entropyrank} presents an unsupervised approach for extracting keywords and keyphrases from textual data. This method leverages a pre-trained language large FMs and incorporates Shannon's information maximization.
LLMZip~\cite{valmeekam2023llmzip} employs LLaMA-7B for compressing natural language.
Experimental results demonstrate that LLMZip outperforms cutting-edge text compression methods, including BSC, ZPAQ, and paq8h. AutoCompressors~\cite{chevalier2023adapting} utilizes large FMs to compress natural language into compact summary vectors. These vectors can then serve as soft prompts for large FMs usage.
ICAE~\cite{ge2023incontext} utilizes the capabilities of large FMs to condense an extensive context into concise memory slots. These memory slots are directly adaptable by the large FMs for diverse purposes.
Nugget 2D~\cite{qin2023nugget} introduces a prompt compression method specifically designed to handle long contexts.
CoT-Max~\cite{huang2023boosting} is a context pruner, aiming to enhancethe Chain-of-Thought ability of large FMs.

\textbf{Token Pruning.}
Research has also explored the pruning of input sequences for transformers, often involving the incremental removal of less important tokens during inference.
PoWER-BERT~\cite{goyal2020powerbert} proposes the direct learning of token pruning configurations. Length-Adaptive Transformer~\cite{kim2021lengthadaptive} extends this idea by introducing LengthDrop, a technique that entails training the model with various token pruning configurations, followed by an evolutionary search. TR-BERT~\cite{ye2021trbert} formulates token pruning as a multi-step token selection problem and addresses it through reinforcement learning.
For ViT, there are also some dynamic token pruning methods. 
DynamicViT~\cite{rao2021dynamicvit} hierarchically prunes redundant tokens based on their importance scores.
AdaViT~\cite{meng2021adavit} and A-Vit~\cite{yin2022adavit}  employ adaptive token reduction mechanisms and select different tokens for different images. AdaViT dynamically determines the usage of patches, self-attention heads, and transformer blocks based on the input.
A-ViT discards tokens in vision transformers during inference, adapting the token retention based on the complexity of the input images.
SPViT~\cite{kong2022spvit} devises an adaptive instance-wise token selector and introduces a soft pruning technique.
PuMer~\cite{cao2023pumer} combines similar textual and visual tokens during inference for large-scale vision language models.
\update{
Mini-Gemini~\cite{li2024minigemini} utilizes twin encoders for both high-resolution images and low-resolution image embeddings without increasing the visual token count.
LLaVA-UHD~\cite{xu2024llava} is also able to process high-resolution images with any aspect ratio:
It utilizes modularized visual encoding to divide native-resolution images into smaller variable-sized slices.
These image tokens will be further condensed through a compression module, along with a spatial schema to inform LLMs about the slice positions.
}

\subsubsection{Key-Value Cache}
\label{subsec:kv-cache}

Optimizing memory for the KV cache is a crucial aspect of the autoregressive decoder-based model inference process.

\textbf{Quantizing KV cache as activation.} 
One strategy involves treating the KV cache as an activation and applying quantization techniques for low-precision compression. The primary challenge arises from quantization errors introduced by numerous extreme outliers in the KV cache. In Figure~\ref{fig:alg-overview}, we present the key methods for KV cache quantization. A detailed introduction to these methods will be provided in $\S$\ref{subsec:model-compression} as part of the Weights-Activation Co-Quantization

\textbf{Memory efficient sparse attention.}
An alternative approach involves leveraging sparse attention. However, it's noteworthy that most sparse attention designs, which primarily target the reduction of computational complexity~\cite{beltagy2020longformer, zaheer2020big}, do not necessarily lead to a reduction in KV cache memory consumption. This is because achieving a reduced memory footprint for the KV cache necessitates a more stringent sparsity pattern. Specifically, tokens that are sparsified should not be dynamically accessed in subsequent steps.
To address this, H2O~\cite{zhang2023h2o} introduces a KV cache eviction strategy designed for optimal memory efficiency. This strategy employs attention scores to identify and select the least important KV cache tokens in the current state for eviction. When compared to robust baselines, H2O demonstrates the capability to reduce latency by up to 1.9$\times$ and increase throughput by 29$\times$.
Dynamic Context Pruning~\cite{anagnostidis2023dynamic} learns a memory-efficient KV cache eviction strategy during the pre-training phase. This approach has demonstrated the ability to achieve up to a 2$\times$ increase in inference throughput and even greater memory savings.
Scissorhands~\cite{liu2023scissorhands} presents an efficient algorithm for large FM inference that utilizes a compact KV cache.
This innovative approach results in a notable reduction in KV cache inference memory usage, achieving up to a 5$\times$ reduction while maintaining model quality.
By employing a landmark token to demarcate a token block, Landmark Attention~\cite{mohtashami2023landmark} optimizes KV cache storage. This approach enables the storage of most KV cache in a slower but larger capacity memory, resulting in reduced memory requirements without compromising performance.
\update{
DeepSeek-V2~\cite{deepseekai2024deepseekv2strongeconomicalefficient} proposes a Multi-head Latent Attention (MLA) method that refines Grouped-Query Attention (GQA) to a latent representation with stronger ability.
It re-up-projects the GQA's K and V to a high-dimensional form by a matrix that can be absorbed to $W_{k}$ and $W_{V}$ at inference time.
FastGen~\cite{ge2024modeltellsdiscardadaptive} identifies the attention sparsity pattern of various heads, and determines the most suitable one for each head with lightweight attention profiling.
}

\textbf{Block-wise KV cache management.}
vLLM~\cite{kwon2023efficient} adopts the concept from paged memory in operating systems, and manages the KV cache susceptible to memory fragmentation as blocks. 
The computation results of this block-based attention mechanism are identical to those of the standard attention mechanism.
We will elaborate its details in $\S$\ref{subsec:cloud-serving-sys}.
\update{Although vLLM significantly reduces KV cache memory consumption through runtime reallocation, the practitioners have to rewrite the attention kernel or reuse vLLM's kernel, both of which lead to downgraded performance.
To tackle this, vAttention~\cite{prabhu2024vattentiondynamicmemorymanagement} directly relies on the OS/CUDA to do the reallocation on the physical memory. It further improves the end-to-end serving throughput by up to 1.29$\times$ compared to vLLM.
On mobile devices, LMS~\cite{yin2024llmservicemobiledevices} proposes a fine-grained KV cache management method that opportunistically compresses the KV cache chunks and orchestrates swapping and recomputing to manage them.
It ensures fast context switching for the on-device LLMaaS vision proposed by LMS.
}

\update{
\noindent \textbf{Miscellaneous.} 
Mooncake~\cite{qin2024mooncakekvcachecentricdisaggregatedarchitecture} proposes a prefill-decode disaggregated system.
The core is to optimize the heavy KV cache of the requests by caching and reusing similar content.
CacheGen~\cite{liu2024cachegenkvcachecompression} treats the KV cache as a data stream and encodes the stream via marking incremental values and arithmetic coding.
InfiniGen~\cite{lee2024infinigenefficientgenerativeinference} compresses the KV cache via Singular Value Decomposition (SVD) and prefetch the critical KV.
CachedAttention~\cite{gao2024costefficientlargelanguagemodel} uses HBM swapping to cache the KV cache for multi-turn conversations.
}

\subsubsection{Long Context}
\label{subsubsec:long-context}

To effectively process long sequences, transformers need to adapt their positional encoding to enhance their capability to capture long-range information. Notable works addressing this challenge include \cite{press2021train, sun2022length, chen2023clex, chen2023extending, SketchyRedditPost2023, peng2023yarn, DBLP:journals/corr/abs-2310-04418, zhu2023pose}. These works serve as a prerequisite to enable transformers to handle long-context and high-resolution input.
Due to the quadratic computational cost associated with attention mechanisms, various resource-efficient optimizations have been proposed to handle long inputs. These optimizations include Recurrent Structure and Attention Optimization.

\textbf{Recurrent structure.}
\label{subsec:recurrent-structure}
Transformer-XL~\cite{dai2019transformerxl} introduces a segment-level recurrence mechanism that maintains temporal coherence without disruption. This innovation results in evaluation speeds up to 1,800+ times faster than vanilla Transformers.
By integrating recurrence and a fixed-size memory, aligning with the Transformer-XL concept, RMT~\cite{bulatov2022recurrent} incorporates a memory mechanism into the Transformer model without making modifications. This is achieved by introducing special memory tokens to the input or output sequence.
Block-Recurrent Transformers~\cite{hutchins2022block} employ a recurrent cell that processes blocks of tokens, rather than individual tokens, during training. This approach harnesses parallel computation within a block, effectively utilizing accelerator hardware.
Memformer~\cite{wu2020memformer} accomplishes a theoretically infinite temporal range of memorization while preserving linear computational complexity and constant memory space complexity.

\textbf{Attention optimizations.}
LM-Infinite~\cite{han2023lminfinite} introduces a $\Lambda$-shaped attention mechanism to handle long contexts efficiently. Characterized by computational efficiency with O(n) time and space complexity, LM-Infinite consistently demonstrates fluency and quality in text generation for sequences as long as 128k tokens on ArXiv and OpenWebText2 datasets.
StreamingLLM~\cite{xiao2023efficient} facilitates large FMs trained with a finite-length attention window to generalize to infinite stream decoding without the need for any fine-tuning.
PCW~\cite{ratner2023parallel} segments a long context into chunks or ``windows'', constrains the attention mechanism to operate solely within each window, and reuses positional embeddings across the windows.
LongNet~\cite{ding2023longnet} introduces dilated attention, expanding the attentive field exponentially as the distance increases. This innovation allows LongNet to scale Transformers efficiently, enabling them to handle sequences of up to 1B tokens.
SLED~\cite{ivgi2023efficient}, short for SLiding-Encoder and Decoder, repurposes and capitalizes on well-validated short-text pre-trained language models. Despite competing effectively with specialized models that are up to 50$\times$ larger, SLED does not require a dedicated and expensive pretraining step.

\subsection{Model Compression}
\label{subsec:model-compression}

\begin{figure*}[t]
	\centering
	\includegraphics[width=\textwidth]{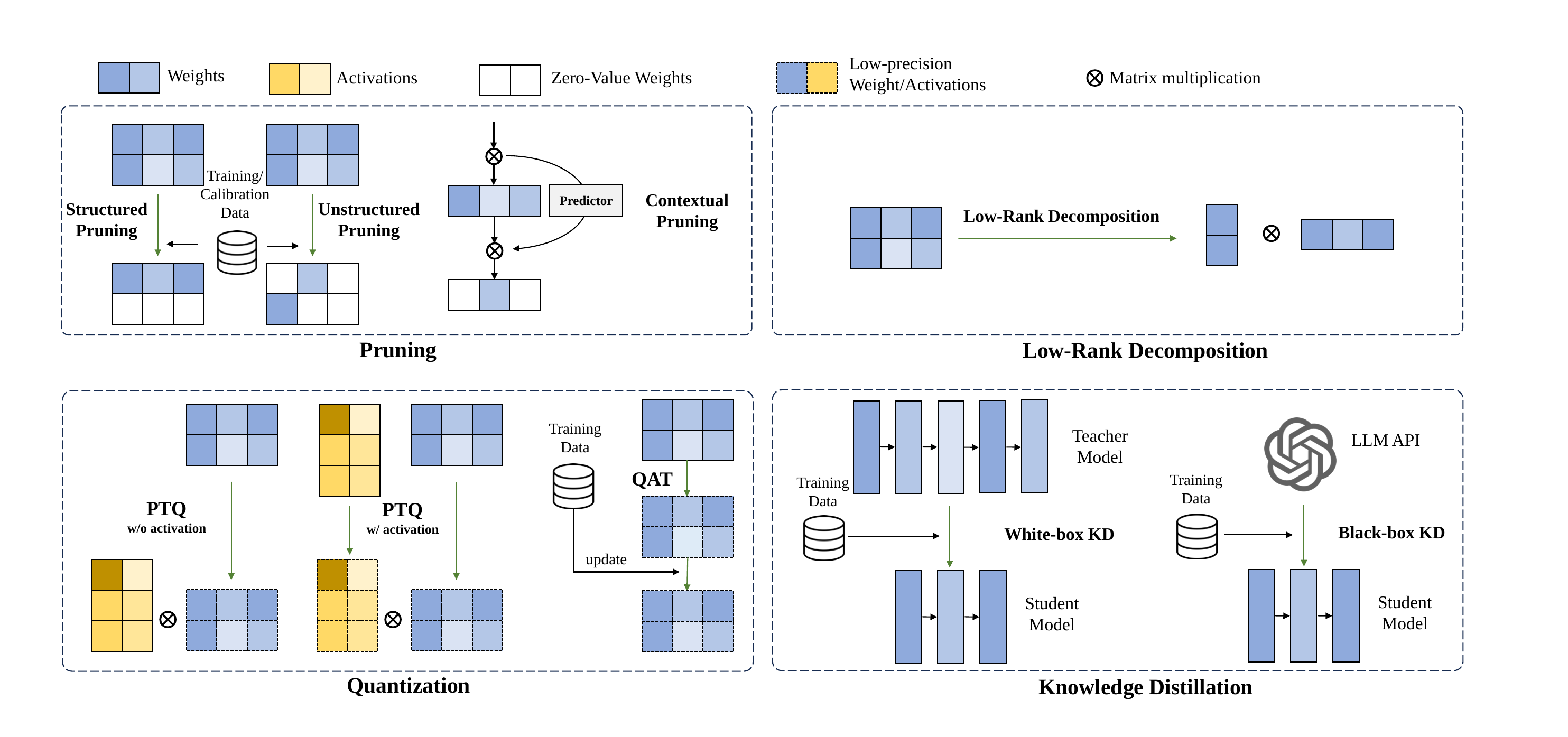}
 \vspace{-10pt}	
	\caption{
		Model compression techniques for LLMs. 
		}
	\label{fig:alg-compression-techniques}
\end{figure*}

Model compression refers to a set of techniques aimed at reducing the size of a model without significant performance degradation.
This survey investigates four main categories of model compression of large FMs: pruning, knowledge distillation, quantization, and Low-Rank decomposition, as shown in Figure~\ref{fig:alg-compression-techniques}. 


\subsubsection{Pruning}
\label{subsubsec:pruning}
Pruning technique removes redundant or non-essential connections, neurons, or layers from a neural network. The primary objective is to reduce the model size, subsequently decreasing computational and storage costs, while maintaining model accuracy. Structured pruning and unstructured pruning target weight reduction without modifying sparsity during inference. In contrast, Contextual Pruning dynamically selects activated neurons or layers during inference based on the sparsity of the model.

 \textbf{Structured Pruning.}
Structured pruning compresses large foundational models by eliminating entire structural components, such as groups of consecutive parameters or hierarchical structures. Examples of these structural components include channels or blocks of the model's weights.

Structured pruning is frequently combined with fine-tuning to mitigate accuracy losses. 
LLM-Pruner~\cite{ma2023llm} is a task-agnostic structured pruning algorithm that utilizes a small amount of data to assess the importance of coupled structure weights. The method selectively removes non-essential model structures based on gradient information. LLM-Pruner incorporates LoRA to recover the model's accuracy after pruning
LoRAPrune~\cite{zhang2023pruning} is another structured pruning approach based on LoRA, leveraging LoRA's weights and gradients for importance estimation. This method iteratively eliminates excess channels and attention heads, achieving superior results compared to LLM-Pruner.
Lagunas et al.~\cite{lagunas2021block} improved structured pruning techniques by incorporating blocks of variable sizes. This integration is applied within the movement pruning framework during fine-tuning, resulting in the removal of entire model components, such as attention heads. The outcome is a pruned model that achieves a 2.4$\times$ speedup and is 74\% smaller compared to the original BERT.

Structured pruning is employed in the training of large foundational models as well. Sheared LLaMA~\cite{xia2023sheared} adopts an end-to-end approach to remove channels, encompassing layers, attention heads, intermediate layers, and hidden layers. This process dynamically loads batches and alters the model structure for each training batch based on losses in various domains. Sheared LLaMA demonstrates the capability to prune the LLaMA2-7B model down to 1.3B parameters.
AdaPrune\cite{hubara2021accelerated} accelerates neural network training using transposable masks, resulting in a 2$\times$ speed-up in matrix multiplications during both inference and training. This is achieved with minimal accuracy degradation, and AdaPrune allows for a seamless transition between different structural constraints.
GUM~\cite{santacroce2023matters} considers neuron specificity and introduces pruning through network component-based global mobility and local uniqueness scores. This approach aims to simultaneously maximize sensitivity and uniqueness, effectively reducing redundant parameters in large FMs weights.
PLATON~\cite{zhang2022platon} tackles the uncertainty in importance scores during model pruning by employing the upper confidence bound of importance estimation. This approach ensures stability in training and leads to improved generalization.

Structured pruning is frequently combined with quantization techniques for model compression. DJPQ~\cite{wang2020differentiable} approaches neural network compression as a unified gradient-based optimization problem. DJPQ integrates variational information bottleneck-based structured pruning and mixed-bit precision quantization into a single differentiable loss function.
SpAtten~\cite{wang2021spatten} represents an algorithm-architecture co-design for efficient attention computation in NLP. This approach leverages token and head sparsity, along with quantization. Employing novel cascade token and head pruning and progressive quantization, SpAtten achieves a significant reduction in DRAM access. This leads to impressive speedup and energy savings across various accelerators and GPUs.

There is also substantial research specifically focusing on structured pruning. Block-Skim~\cite{guan2022block} enhances extractive question answering Transformers by selectively processing vital context using self-attention weights, optimizing through early pruning of unnecessary positions in lower layers. In a similar vein, DepGraph~\cite{fang2023depgraph} introduces a fully automatic method for general structural pruning in diverse architectures. This method addresses the challenge of structural coupling and consistently achieves performance improvement across various models using a norm-based criterion.

\textbf{Unstructured Pruning.}
Unlike structured pruning, unstructured pruning does not consider the inherent structure of the model.
Typically, it removes neurons with weights below a threshold, thereby reducing the computational load to compress the model. When deploying unstructured pruning, specialized techniques are required to implement model storage compression.
SparseGPT~\cite{frantar2023sparsegpt} stands as a one-shot pruning algorithm that eliminates the need for retraining. SparseGPT treats the pruning framework as a generalized sparse regression problem and employs an approximate sparse regression solver to address it. SparseGPT achieves 60\% unstructured pruning on large GPT models like 175B. 
Wanda~\cite{sun2023simple} leverages the observation of emergent large-magnitude features in large FMs. Wanda introduces sparsity by pruning weights with the smallest magnitudes multiplied by corresponding input activations, on a per-output basis.
UPop~\cite{shi2023upop} serves as a universal vision-language Transformer compression framework, which incorporates two key components: (1) unifiedly searching multimodal subnets in a continuous optimization space from the original model. This enables the automatic assignment of pruning ratios among compressible modalities and structures; 
(2) progressively searching and retraining the subnet, maintaining convergence between the search and the retrain to achieve higher compression ratios.
SSI is a technique optimizing deep generative models during image editing by selectively computing for edited regions. Leveraging cached feature maps of the original image, SSI minimizes redundant computations, resulting in significant acceleration for various generative models, such as conditional GANs and diffusion models. 
SIGE~\cite{shi2023upop} is proposed to convert computation reduction into latency reduction on standard hardware, achieving notable accelerations for models like DDPM, Stable Diffusion, and GauGAN with minimal edits.

\textbf{Contextual Pruning.}
Different from structured pruning and unstructured pruning, context pruning dynamically selects the sparse state of each layer, making it hardware-optimization friendly.
Deja Vu~\cite{liu2023deja} dynamically predicts the sparsity of the next layer using the activations of the previous layer. It determines which neurons of MLP blocks and the heads of attention blocks need to be retained. To mitigate the overhead of this predictor, Deja Vu asynchronously predicts the next layer.
PowerInfer~\cite{song2023powerinfer} utilizes the sparsity of activation to dynamically predict the hot-activated neurons of the next layer and computes them on the GPU, while other cold-activated neurons are computed on the CPU.
In comparison to llama.cpp~\cite{llamacpp}, PowerInfer achieves up to 11$\times$ acceleration, enabling the 40B model to output ten tokens per second on a personal computer.
\update{
PowerInfer-2~\cite{xue2024powerinfer} extends PowerInfer to smartphones by using a polymorphic neuron engine.
It dynamically adjusts computation based on neuron activation and a neuron-cluster-level pipelining technique, effectively optimizing sparse computations by grouping activated neurons into clusters, which significantly reduces overhead.
Song et al.~\cite{song2024achieving} present precise sparse activation specifically in small language models using gradient-based attribution scores with corrections for inter-layer dependencies.
}

\subsubsection{Knowledge Distillation}
\label{subsubsec:know-distil}

Knowledge Distillation (KD) transfers knowledge from a complex, heavy model (i.e., teacher model) to a simpler corresponding model (i.e., student model) for model compression.
In general, there are twy ways to apply KD to large FMs based on whether the internal structure of the teacher model is considered: white-box knowledge distillation and black-box knowledge distillation.

\textbf{Black-box Knowledge Distillation.}
Assuming that the internal structure of the teacher's large base model is not visible,
this approach fine-tunes the student model using prompt-response pairs generated by large FMs' API. The goal is to imbue the student model with the capabilities of the teacher model.
For large FMs, the insights gained due to the increased parameter count contribute to strong generalization abilities. Therefore, techniques such as In-Context Learning (ICL)~\cite{dong2022survey}, Chain-of-Thought (CoT)~\cite{wei2022chain}, and Instruction Following (IF) ~\cite{ouyang2022training} can be utilized to enable the student model to thoroughly learn the capabilities of the large FMs.

ICL distillation transfers few-shot learning and language model capabilities from the teacher model to the student model by integrating in-context learning objectives with traditional language modeling objectives.
In Meta-ICL~\cite{min2021metaicl} and Metal-ICL~\cite{chen2021meta}, language models undergo meta-training on diverse tasks using in-context learning objectives. This process enables them to fine-tune for unseen tasks through in-context learning.
Multitask-ICT~\cite{huang2022context}introduces the concept of in-context learning distillation, fine-tuning models with ICL objectives and examples from target tasks. This method makes predictions using in-context learning on these tasks. While Multitask-ICT outperforms Meta-ICT in terms of performance, it requires more multi-task learning along with explanations generated by large FMs. 

IF can enhance the zero-shot capability of models, where task descriptions serve as instructions for model fine-tuning. Lion~\cite{jiang2023lion} is an adversarial distillation architecture that prompts the large FMs to recognize challenging instructions and create new complex instructions for the student model, thereby enhancing the student model's capabilities.
LaMini-LM~\cite{wu2023lamini}, designed for resource-intensive language models, develops an extensive instruction set comprising existing and newly generated instructions for fine-tuning student models.

CoT introduces intermediate reasoning steps in prompts, guiding language models to solve complex reasoning tasks step by step. Black-box knowledge distillation can leverage this approach to transfer knowledge from large FMs to smaller student models.
Fu et al.~\cite{fu2023specializing} enhanced the mathematical reasoning capabilities of smaller models by instructing them through CoT distilled from LLM teachers.
Distilling Step-by-Step~\cite{hsieh2023distilling} extracts rationales from large FMs using CoT in a multi-task framework, providing additional guidance for training smaller models in a multi-task environment.
Fine-tune-CoT~\cite{ho2022large} uses zero-shot CoT prompting techniques, mploying random sampling to generate multiple reasoning solutions from large FMs to guide the training of student models.
SOCRATIC CoT~\cite{shridhar2023distilling} trains both a problem decomposer and a sub-problem solver as distilled models. The decomposer breaks down the original problem into a series of sub-problems solved by the solver.
DISCO~\cite{chen2023disco} generates perturbations with large FMs, and a specialized teacher model filters these perturbations to distill high-quality counterfactual data into student models.
SCOTT~\cite{wang2023scott} learns a small, self-consistent CoT model from a significantly larger teacher model. SCOTT uses teacher-generated rationales to train a student LM with a counterfactual reasoning objective, preventing the student from ignoring the rationales.
SCoTD~\cite{li2023symbolic} introduces a method called symbolic CoT distillation, involving extracting CoT rationales from LLMs using unlabeled data instances. A smaller model is then trained to predict both the sampled rationales and associated labels.
CoT Prompting~\cite{magister2022teaching} explores the transferability of this reasoning ability to smaller models through knowledge distillation, finding a trade-off between model and dataset size in reasoning ability.
PaD~\cite{zhu2023pad} distills  arge FMs to obtain specialized small models in reasoning tasks. PaD reinforces specialized models with program-aided reasoning and helps them overcome faulty reasoning steps with automated error checking.

\textbf{White-box Knowledge Distillation.}
In contrast to black-box knowledge distillation, white-box knowledge distillation not only has access to the output results of the teacher model but also to its structure and intermediate results. Therefore, white-box knowledge distillation can better leverage the structure of the teacher model, enabling smaller student models to replicate and learn the capabilities of larger teacher models.

Timiryasov et al.~\cite{timiryasov2023baby} trained an ensemble consisting of a GPT-2 and small LLaMA models on the developmentally plausible BabyLM dataset, which has 10 million words. 
Subsequently, they distilled it into a small LLaMA model with 58 million parameters, surpassing in performance both of its teachers as well as a similar model trained without distillation. 
MiniLLM~\cite{gu2023knowledge} distills smaller language models from generative larger language models. 
This approach replaces the forward Kullback-Leibler divergence (KLD) objective in the standard KD approaches with reverse KLD, which is more suitable for KD on generative language models, to prevent the student model from overestimating the low-probability regions of the teacher distribution. 
MiniLLM then derives an effective optimization approach to learn this objective. 
Instead of solely relying on a fixed set of output sequences, 
GKD~\cite{agarwal2023gkd} trains the student model using self-generated output sequences. This is achieved by leveraging feedback from the teacher on these self-generated sequences.
Unlike supervised KD approaches, GKD also provides the flexibility to use alternative loss functions between the student and teacher models. This flexibility is valuable, especially in cases where the student model lacks the expressivity to mimic the teacher's distribution.
KPTD~\cite{padmanabhan2023propagating} involves two stages: transfer set generation and distillation on the transfer set. In the first stage, KPTD generates a transfer set by prompting a language model to generate continuations from the entity definition. Subsequently, KPTD updates the model parameters to align the distribution of the student with the distribution of the teacher conditioned on the transfer set
TED~\cite{liang2023less} employs task-aware filters to align the hidden representations of the student and the teacher at each layer. These filters are designed to select task-relevant knowledge from the hidden representations. By narrowing down the focus to task-specific information, TED aims to reduce the knowledge gap between the two models, enhancing the student's performance on the target task.
MixKD~\cite{liang2020mixkd} is a data-agnostic distillation framework that incorporates mixup, a straightforward yet effective data augmentation technique. In MixKD, the student model is encouraged not only to learn from the original training examples but also to mimic the teacher's behavior on the linear interpolation of example pairs. This additional training strategy aims to enhance the generalization ability of the resulting model. 
DIME-FM~\cite{sun2023dime} is a method designed for transferring knowledge from large vision foundation models to smaller, customized foundation models. This is achieved using a relatively small amount of inexpensive, unpaired images and sentences, and notably, without relying on public or smaller-scale datasets. The approach provides a means of efficient knowledge transfer in scenarios where paired data may be limited or unavailable.
Li et al.~\cite{li2023distilling} proposed two principles from vision and language modality perspectives to enhance student's out-of-distribution generalization: by better imitating the teacher's visual representation space, and carefully promoting better coherence in vision-language alignment with the teacher; by enriching the teacher's language representations with informative and fine-grained semantic attributes to effectively distinguish between different labels.

\subsubsection{Quantization}
\label{subsubsec:quantization}

Quantization is a well-established model compression method to mitigate the storage and computational demands. This approach involves the conversion of weights and activations, initially expressed in conventional high-precision floating-point formats, into lower-bit high-precision or integer representations. 

Equation \eqref{eq:quantize} illustrates the typical process of converting a 32-bit floating-point tensor $X^{\text{FP32}}$ into an $N$-bit integer tensor $X^{\text{Int}N}$. 
Equation \eqref{eq:dequantize} demonstrates the dequantization process.
Symble $c^{\text{FP32}}$ represents the scaling factor.

\begin{equation}
    \begin{aligned}
    X^{\text{Int}N} 
    &= \text{quantize}\left(N, X^{\text{FP32}}\right) \\
    &= \text{Round} \left( \frac{2^N}{\text{absmax}(X^{\text{FP32}})} \times X^{\text{FP32}} \right) \\
    &= \text{Round}(c^{\text{FP32}} \times X^{\text{FP32}})
    \end{aligned}
    \label{eq:quantize}
\end{equation}

\begin{equation}
    \begin{aligned}
    X^{\text{FP32}}
    &= \text{dequantize}\left(c^{\text{FP32}}, X^{\text{Int}N}\right)  
    &= \frac{X^{\text{Int}N}}{c^{\text{FP32}}} 
    \end{aligned}
    \label{eq:dequantize}
\end{equation}

In practical applications, the dequantization process is not universally prevalent. 
DNN engines frequently exploit the integer computation capabilities of accelerators to directly execute matrix multiplication on quantized weights and activations, thereby accelerating computations. 
For example, ggml~\cite{ggml} and Intel Extension for Transformers~\cite{shen2023efficient, intel-extension-for-transformers} leverage the SIMD instruction sets of CPUs to perform integer matrix multiplication. Similarly, bitsandbytes~\cite{dettmers2022llm, bitsandbytes} and QNN~\cite{qualcomm_neural_processing_sdk} have implemented corresponding integer matrix multiplication for GPU and NPU.

According to the timing of the quantization process, quantization can be categorized into post-training quantization (PTQ) and quantization-aware training (QAT).

\textbf{Quantization-Aware Training.}
QAT involves training a quantized model in such a way that it adapts its parameters to the lower precision introduced by quantization. The primary objective of this process is to mitigate the accuracy loss that occurs as a result of quantization.
LLM-QAT tackles the issue of obtaining training data for LLMs by leveraging pre-trained models to generate samples through data-free distillation. Concurrently, it quantizes weights, activations, and KV cache, thereby improving training throughput. LLM-QAT demonstrates the capability to yield a precisely quantized LLM with 4-bit precision.
QuantGPT~\cite{tao2022compression} achieves this by incorporating contrastive distillation from a full-precision teacher model and distilling logit information to a quantized student model during autoregressive pretraining.
BitNet~\cite{wang2023bitnet} pioneered QAT for 1-bit language models, training the language model with 1-bit weights and activations.
\update{EfficientQAT~\cite{chen2024efficientqat} is a novel quantization technique for compressing LLMs. It consists of two consecutive phases: Block-wise training of all parameters (Block-AP), which sequentially conducts quantization-aware training for all parameters in each transformer block with block-wise reconstruction to maintain efficiency without training the entire LLM. Initialized with a quantized model, the second phase, end-to-end training of quantization parameters (E2E-QP), then trains only quantization parameters (step sizes) end-to-end, enhancing efficiency with a fixed quantized backbone and reduced trainable parameter count.}

Due to the substantial parameter count in large models often reaching tens or hundreds of billions, the training cost of QAT remains considerable. 
On the one hand, QAT for large foundation models is often combined with knowledge distillation to reduce the training cost, as seen in approaches such as LLM-QAT and QuantGPT.
On the other hand, quantization is frequently employed in the fine-tuning process of large models, such as in PEQA~\cite{wu2023understanding}, LoftQ~\cite{li2023loftq} and QLoRA~\cite{dettmers2023qlora}, as detailed in $\S$\ref*{sec:alg:finetune:reparam}.

\textbf{Post-Training Quantization.}
PTQ converts a trained full-precision model to a low-precision model without retraining. The advantage of PTQ lies in compressing models without altering the model structure or necessitating retraining, thereby reducing the storage and computational costs of models.
Due to its low deployment cost, PTQ is also the most easily deployable and widely applicable technique in model compression. However, unlike QAT and distillation, PTQ lacks the feedback loop for adjusting precision through training. Research related to PTQ often focuses on efficiently preserving relevant information in weights/activations while compressing models.

PTQ can be categorized into two groups: weight-only quantization and weight-activation co-quantization.


\textit{Weight quantization quantizes the model weights only of large foundation models.}
There are two primary methods for mitigating quantization errors in the weight quantization of large foundation models. 

The first category involves identifying outliers and important weights in weights that significantly contribute to accuracy and treating these outliers specially. For instance, SpQR~\cite{dettmers2023spqr} identifies outlier weights and maintains them with high precision, while quantizing the rest of the weights.
LLM.int8()~\cite{dettmers2022llm} employs vectorized quantization and mixed-precision decomposition to handle outlier values for efficient inference. LLM.int8() utilizes 8-bit quantization for matrix multiplication, effectively reducing GPU memory usage during inference. 
AWQ~\cite{lin2023awq} reduces quantization error by protecting the top 1\% important weights in the model, utilizing per-channel scaling to determine the optimal scaling factor. 
OWQ~\cite{lee2023owq} analysis suggests that abnormal activations amplify quantization errors, and it employs a mixed-precision quantization scheme, applying higher precision quantization to weights with a significant impact from activated outlier values. 
SqueezeLLM~\cite{kim2023squeezellm} observes that certain weights determine the final model's quantization performance and proposes a non-uniform quantization approach to minimize quantization errors in these sensitive weights. SqueezeLLM achieves this by leveraging second-order Hessian information from the loss to identify quantization-sensitive weights and placing quantization points near these sensitive weights.

The second category of quantization reduction methods is based on the second-order information updated weights. 
GPTQ~\cite{frantar2022gptq} employs layer-wise quantization with OBQ~\cite{frantar2022optimal}, utilizing inverse Hessian information to update weights. 
GPTQ reduces the bit-width of each weight to 3 or 4 bits, allowing quantization of GPT models with 175 billion parameters with minimal accuracy loss. 
QuIP~\cite{chee2023quip} uses an adaptive rounding process, minimizing a second-order proxy objective for quantization.
QuIP efficiently preprocesses and post-processes weights and Hessians using a random orthogonal matrix to ensure the weights and Hessian are uncorrelated, achieving 2-bit quantization for large FMs.

Simultaneously, certain quantization methods can yield computational acceleration while compressing the model.
For example, llama.cpp~\cite{llamacpp} employs the K-quant method. 
The K-quant quantization utilizes blocks of size $16\times8$ for quantization, with each block containing a total of 16 rows. 
Similar to QLoRA~\cite{dettmers2023qlora}'s secondary quantization, for further resource reduction, there is an additional FP16 secondary quantization parameter used to quantize the 16 primary quantization parameters. 
llama.cpp demonstrates that performing model inference using K-quant quantization is approximately 3-4 $\times$ faster than inference with the original FP16 model.

\textit{Weights-activation co-quantization quantizes both the weights and activations of large foundation models.}
Similar to weight quantization, identifying and handling outliers in both weight and activation  can also contribute to the reduction of quantization errors. 
SmoothQuant~\cite{xiao2023smoothquant} takes advantage of the similarity in the channel-wise activations of different tokens and performs quantization on both weight and activation using per-channel scaling transforms. 
RPTQ~\cite{yuan2023rptq} recognizes the substantial range differences across different channels, reordering the channels for quantization and integrating them into Layer Normalization and linear layer weights.
OliVe~\cite{guo2023olive} adopts outlier-victim pair (OVP) quantization and locally processes outliers. 
Outlier Suppression+~\cite{wei2023outlier} builds upon Outlier Suppression~\cite{wei2022outlier}, discovering that harmful outliers exhibit an asymmetric distribution mainly concentrated in specific channels. 
Considering the asymmetry of outliers and quantization errors from the weights of the next layer, this approach performs channel-level translation and scaling operations.
QLLM~\cite{liu2023qllm} addresses the issue of activation outliers through an adaptive channel reassembly method and mitigates the information loss caused by quantization using calibration data.
LLM-FP4~\cite{liu2023llmfp4} quantizes weights into 4-bit float points, proposes per-channel activation quantization, and reparameters additional scaling factors as exponential biases of weights.
LLM-FP4 successfully quantizes both weights and activations in the LLaMA-13B to only 4-bit and achieves only 8.4\% accuracy loss than the full-precision model. 
ZeroQuant~\cite{yao2022zeroquant} combines layer-wise knowledge distillation and optimized quantization support to achieve 8-bit quantization.
ZeroQuant-V2~\cite{yao2023zeroquant} introduces Low Rank Compensation (LoRC) for further optimization.
ZeroQuant-FP~\cite{wu2023zeroquant} supports FP4/FP8 quantization and achieves superior performance, compared to integer quantization with the corresponding bit precision.
FlexRound~\cite{lee2023flexround} updates the quantization scale of weights and activations by minimizing the error between the quantized values and the full-precision values.
ATOM~\cite{zhao2023atom} significantly boosts serving throughput by using low-bit operators and considerably reduces memory consumption via low-bit quantization. 
ATOM on 4-bit weight-activation quantization improves end-to-end throughput by up to 7.73$\times$ compared to the FP16 and by 2.53$\times$ compared to INT8 quantization, while maintaining the same latency target.
\update{I-LLM~\cite{hu2024llm} is a novel integer-only fully-quantized PTQ framework for LLMs. It develops Fully-Smooth Block-Reconstruction (FSBR) to smooth inter-channel variations of activations and weights. It also introduces Dynamic Integer-only MatMul (DI-MatMul) to alleviate degradation caused by inter-token variations and enables dynamic quantization in full-integer matrix multiplication. Additionally, it designs DI-ClippedSoftmax, DI-Exp, and DI-Normalization to execute non-linear operators efficiently with bit shift while maintaining accuracy. }

\update{
\textbf{Rotation-based approaches.}
To quantize outliers efficiently, recent studies adopt a distinct approach and reveal an intriguing property: multiplying the weight matrix by a random rotation can efficiently reduce outliers and improve quantizability.
Intuitively, due to the statistical characteristic of random rotation, this transformation results in a distribution of weight or activation entries that is free from outliers. 
Since rotation matrices can be constructed in pairs from identity mapping and can be integrated into nearby weights without altering the overall network outputs—a property known as rotational invariance—the transformed weights or activations can be quantized with lower reconstruction error and without incurring additional inference overhead.
SpinQuant~\cite{liu2024spinquant} optimizes the rotation matrix to minimize the final loss of the quantized network, with fixed weight parameters, by employing the Cayley SGD~\cite{li2020efficient}, a proficient technique for optimizing orthonormal matrices. This optimization does not alter the full-precision network output but refines the intermediate activations and weights, making them more quantization-friendly.
QuaRot~\cite{ashkboos2024quarot} has two stages. The first stage manipulates model weights (in full precision) and inserts two additional Hadamard operations in the model's forward pass. The second stage quantizes weights using GPTQ by default and quantizes activations on the fly with a simple round-to-nearest scheme.
}

There is also extensive quantization research for backbone networks in FMs like ViT and BERT. 
For instance, BinaryBERT~\cite{novikov2023few} and I-BERT~\cite{bai2020binarybert} have achieved higher accuracy for BERT under low-precision quantization. 
Wang et al.~\cite{wang2022towards} exploited the operator fusion~\cite{niu2021dnnfusion}, PTQ techniques, and structured pruning~\cite{lagunas2021block} to reduce the memory cost.
They also reduce the number of computation operations of DeiT-Tiny~\cite{touvron2012training}.
Q-ViT~\cite{li2022q}, I-ViT~\cite{li2023vit}, and OFQ~\cite{liu2023oscillation} also achieve high accuracy for ViT under low-precision quantization. 
Q-Diffusion~\cite{li2023q} compresses the noise estimation network to expedite the generation process of diffusion models.


\subsubsection{Low-Rank Decomposition}
\label{subsubsec:low-rank-decomposition}

Low-rank decomposition (LoRD) approximates weight matrix in large FMs by decomposing a given weight matrix into two or more smaller matrices. 
For an $m \times n$ weight matrix $W$, the decomposition is given by $W = UV$, where U is an $m \times k$ matrix, and $V$ is a $k \times n$ matrix, with $k$ much smaller than $m$ and $n$. 
The product of $U$ and $V$ approximates the original weight matrix, significantly reducing the number of parameters and computational overhead. 

As mentioned in $\S$\ref{sec:alg:finetune:reparam}, low-rank decomposition has been widely applied in large FMs finetuning methods like LoRA.
LoRD has also shown substantial compression capabilities with minimal impact on performance, highlighting its potential for large FMs compression~\cite{kaushal2023lord}.
To reduce the dimensionality of high-dimensional token embeddings underpinning large FMs,  TensorGPT~\cite{xu2023tensorgpt} proposes an approach based on the Tensor-Train Decomposition (TTD), where each token embedding is treated as a Matrix Product State (MPS) that can be efficiently computed in a distributed manner.
Through TensorGPT, the embedding layer can be compressed by a factor of up to 38.40$\times$. 
LoSparse~\cite{li2023losparse} employs low-rank approximation to compress the coherent and expressive elements.
The method uses iterative training to assess the significance scores of column neurons for the pruning process, showcasing superior performance compared to traditional iterative pruning techniques.
Saha et al.~\cite{saha2023matrix} compresses matrices through randomized low-rank and low-precision factorization, achieving compression ratios as aggressive as one bit per matrix coordinate while surpassing or maintaining the performance of traditional compression techniques.
ViTALiTy~\cite{dass2023vitality} is an algorithm-hardware codesigned framework to enhance the inference efficiency of ViTs.
It achieves approximation of the dot-product softmax operation with first-order Taylor attention, utilizing row-mean centering as the low-rank component to linearize the cost of attention blocks.

	\section{RESOURCE-EFFICIENT SYSTEMS}
\label{sec:systems}

Training and serving systems are key to practical large FMs.
This section investigates the system research to enable resource-efficient large FMs, notable at four aspects: (1) distributed training; (2) federated learning; (3) serving in cloud, and (4) serving in edge.
Figure~\ref{fig:tree-resource-efficient-system-tree} outlines the taxonomy of resource-efficient systems, while Table~\ref{tab:tab-open-source-FM-serving-tools} summarizes widely-used open-source frameworks in this domain.

Notably, due to the success of LLM and its vast parameter size, most of the systems and techniques introduced in this section are designed for LLM specifically.

\tikzstyle{my-box}=[
    rectangle,
    draw=black,
    rounded corners,
    text opacity=1,
    minimum height=1.5em,
    minimum width=5em,
    inner sep=2pt,
    align=center,
    fill opacity=.5,
    line width=0.8pt,
]
\tikzstyle{leaf}=[my-box, minimum height=1.5em,
    fill=hidden-red!10, text=black, align=left,font=\normalsize,
    inner xsep=2pt,
    inner ysep=4pt,
    line width=0.8pt,
]

\begin{figure*}[htpb]
    \centering
    \resizebox{\textwidth}{!}{
        \begin{forest}
            forked edges,
            for tree={
                grow=east,
                reversed=true,
                anchor=base west,
                parent anchor=east,
                child anchor=west,
                base=center,
                font=\large,
                rectangle,
                draw=black,
                rounded corners,
                align=left,
                text centered,
                minimum width=4em,
                edge+={black, line width=1pt},
                s sep=3pt,
                inner xsep=2pt,
                inner ysep=3pt,
                line width=0.8pt,
                ver/.style={rotate=90, child anchor=north, parent anchor=south, anchor=center},
            },
            where level=1{text width=16em,font=\normalsize,}{},
            where level=2{text width=12em,font=\normalsize,}{},
            [
                \textbf{Resource-efficient Systems}, ver  
                        [
                            \textbf{Distributed Training ($\S$\ref{subsec:distributed-training-systems})}, fill=blue!10
                            [
                            \textbf{Resilience}, fill=yellow!10
                            [
                            Varuna~\cite{athlur2022varuna}{, }
                            Gemini~\cite{wang2023gemini}{, }
                            Bamboo~\cite{thorpe2023bamboo}{, }
                            Oobleck~\cite{jang2023oobleck},
                            leaf, text width=34.5em
                            ]
                            ]
                            [
                            \textbf{Parallelism}, fill=yellow!10
                            [
                            ZeRO~\cite{rajbhandari2020zero}{, }
                            BFPP~\cite{lamy2023breadth}{, }
                            PipeFisher~\cite{osawa2023pipefisher}{, }
                            Sequence parallelism~\cite{korthikanti2023reducing,li2023sequence}{, }\\
                            Galvatron~\cite{miao2022galvatron}{, }
                            FTPipe~\cite{eliad2021fine}{, } 
                            Mobius~\cite{feng2023mobius},
                            leaf, text width=34.5em
                            ]
                            ]
                            [
                            \textbf{Communication}, fill=yellow!10
                            [
                            Optimux-CC~\cite{song2023optimus}{, }
                            MiCS~\cite{zhang2022mics}{, }
                            CoCoNet~\cite{jangda2022breaking}{, }
                            Wang et al{.}~\cite{wang2022overlap}{, }\\
                            Zhuang et al{.}~\cite{zhuang2023optimizing},
                            leaf, text width=34.5em
                            ]
                            ]
                            [
                            \textbf{Storage}, fill=yellow!10
                            [
                            ZeRO-Offload~\cite{ren2021zero}{, }
                            FlashNeuron~\cite{bae2021flashneuron}{, }
                            Behemoth~\cite{kim2021behemoth},
                            leaf, text width=34.5em
                            ]
                            ]
                            [
                            \textbf{Heterogeneous GPUs}, fill=yellow!10
                            [
                            HetPipe~\cite{park2020hetpipe}{, }
                            Whale~\cite{jia2022whale},
                            leaf, text width=34.5em
                            ]
                            ]
                            [
                            \textbf{MoE}, fill=yellow!10
                            [
                            Megablocks~\cite{gale2023megablocks}{, }
                            Brainstom~\cite{cui2023optimizing}{, }
                            FlexMoE~\cite{nie2023flexmoe}{, }\\
                            Tutel~\cite{hwang2023tutel}{, }
                            SmartMoE~\cite{zhai2023smartmoe}{, }
                            Janus~\cite{liu2023janus},
                            leaf, text width=34.5em
                            ]
                            ]
                        ]
                        [
                          \textbf{Federated Learning ($\S$\ref{subsec:federated-fine-tuning-systems})}, fill=blue!10
                          [
                        \textbf{Framework \& Benchmark}\\
                        \textbf{~~~~~~~~~~~($\S$\ref{subsubsec:related-framework-benchmark})}, fill=yellow!10
                        [
                        Flower~\cite{flower2023whisper, beutel2020flower} {, }
                        FedML~\cite{he2020fedml, lin2022fednlp}{, }
                        FATE~\cite{fan2023fate}{, }
                        Federatedscope-llm~\cite{kuang2023federatedscope}{, }\\
                        Gao et al.~\cite{gao2022federated}{, }
                        Woisetschl{\"a}ger et al.~\cite{woisetschlager2023federated} {, }
                        Zhao~\cite{zhao2023privacy},
                        leaf, 
                        text width=34.5em
                        ]
                      ]
                      [
                        \textbf{PEFT-based Approaches}\\
                        \textbf{~~~~~~~~~~($\S$\ref{subsubsec:peft-based-methods})}, fill=yellow!10 
                        [
                        FedAdapter~\cite{cai2023efficient}{, }
                        FeS~\cite{cai2023federated}{, }
                        FP-FL~\cite{jiang2023low}{, }
                        FedPrompt~\cite{zhao2023fedprompt}{, }
                        Malaviya et al.~\cite{malaviya2023reducing},
                        leaf, text width=34.5em
                        ]
                      ] 
                      [
                        \textbf{Model Decomposition}\\
                        \textbf{~~~~~~~~($\S$\ref{subsubsec:model-decomposition})}, fill=yellow!10 
                        [
                        FedBFPT~\cite{wang2023fedbfpt}{, }
                        FedOBD~\cite{chen2022fedobd}{, }
                        FedBERT~\cite{tian2022fedbert}{, }
                        FedPerfix et al.~\cite{sun2023fedperfix},
                        leaf, text width=34.5em
                        ]
                      ] 
                      [
                        \textbf{Backprop-free Approaches}\\
                        \textbf{~~~~~~~~~~~~($\S$\ref{subsubsec:zeroth-order-optimization})}, fill=yellow!10
                        [
                        FwdLLM~\cite{xu2023federated}{, }
                        Fed-BBFT~\cite{lin2023efficient}{, }
                        FedKSeed~\cite{qin2023federated},
                        leaf, text width=34.5em
                        ]
                      ]
                      ]
                        [
                          \textbf{Serving on Cloud ($\S$\ref{subsec:cloud-serving-sys})}, fill=blue!10
                          [
                            \textbf{Inference Accelerating}\\
                            \textbf{~~~~~~~~~($\S$\ref{subsubsec:inference-accel})}, fill=yellow!10
                            [
                              Flash-Decoding~\cite{flashdecoding}{, }
                              FlashDecoding++~\cite{flashdecoding++}{, }
                              DeepSpeed-Inference~\cite{deepspeed-inference}{, }\\
                              Pope et al.~\cite{pope2023efficiently}{, }
                              Orca~\cite{yu2022orca}{, }
                              Punica~\cite{chen2023punica}{, }
                              FlexGen~\cite{sheng2023flexgen}{, }
                              FastServe~\cite{wu2023fast}{, }\\
                              SARATHI~\cite{agrawal2023sarathi}{, }
                              DeepSpeed-FastGen~\cite{deepspeed-fastgen}{, }
                              Splitwise~\cite{patel2023splitwise},
                              leaf, text width=34.5em
                            ]
                          ]
                          [
                            \textbf{Memory Saving ($\S$\ref{subsubsec:mem-saving})}, fill=yellow!10
                            [
                              DeepSpeed-Inference~\cite{deepspeed-inference}{, }
                              FlexGen~\cite{sheng2023flexgen}{, }
                              vLLM~\cite{kwon2023efficient}{, }
                              S-LoRA~\cite{sheng2023s}{, }\\
                              SGLang~\cite{zheng2023efficiently},
                              leaf, text width=34.5em
                            ]
                          ]
                          [
                            \textbf{Emerging Platforms}\\
                            \textbf{~~~~~~~($\S$\ref{subsubsec:emerging-platforms})}, fill=yellow!10
                            [
                              SpotServe~\cite{miao2023spotserve}{, }
                              HexGen~\cite{jiang2023hexgen},
                              leaf, text width=34.5em
                            ]
                          ]
                        ]
                        [
                          \textbf{Serving on Edge ($\S$\ref{subsec:edge-serving-sys})}, fill=blue!10
                          [
                            \textbf{Edge-Cloud Collaboration}, fill=yellow!10
                            [
                              EdgeMoE~\cite{yi2023edgemoe}
                              ,leaf, text width=34.5em
                            ]
                          ]
                          [
                            \textbf{Edge-Only}, fill=yellow!10
                            [
                            LLMCad~\cite{xu2023llmcad}{, }
                            STI~\cite{guo2023sti}{, }
                            PowerInfer~\cite{song2023powerinfer}{, }
                            PowerInfer-2~\cite{xue2024powerinfer}{, }\\
                            LLM in a flash~\cite{alizadeh2023llm}{, }
                            LMS~\cite{yin2024llmservicemobiledevices}{, }
                            ELMS~\cite{yin2024elmselasticizedlargelanguage},
                            leaf, text width=34.5em
                            ]
                          ]
                        ]
                    ]
        \end{forest}
    }
    
    \caption{An overview of resource-efficient systems.}
    \label{fig:tree-resource-efficient-system-tree}
\end{figure*}
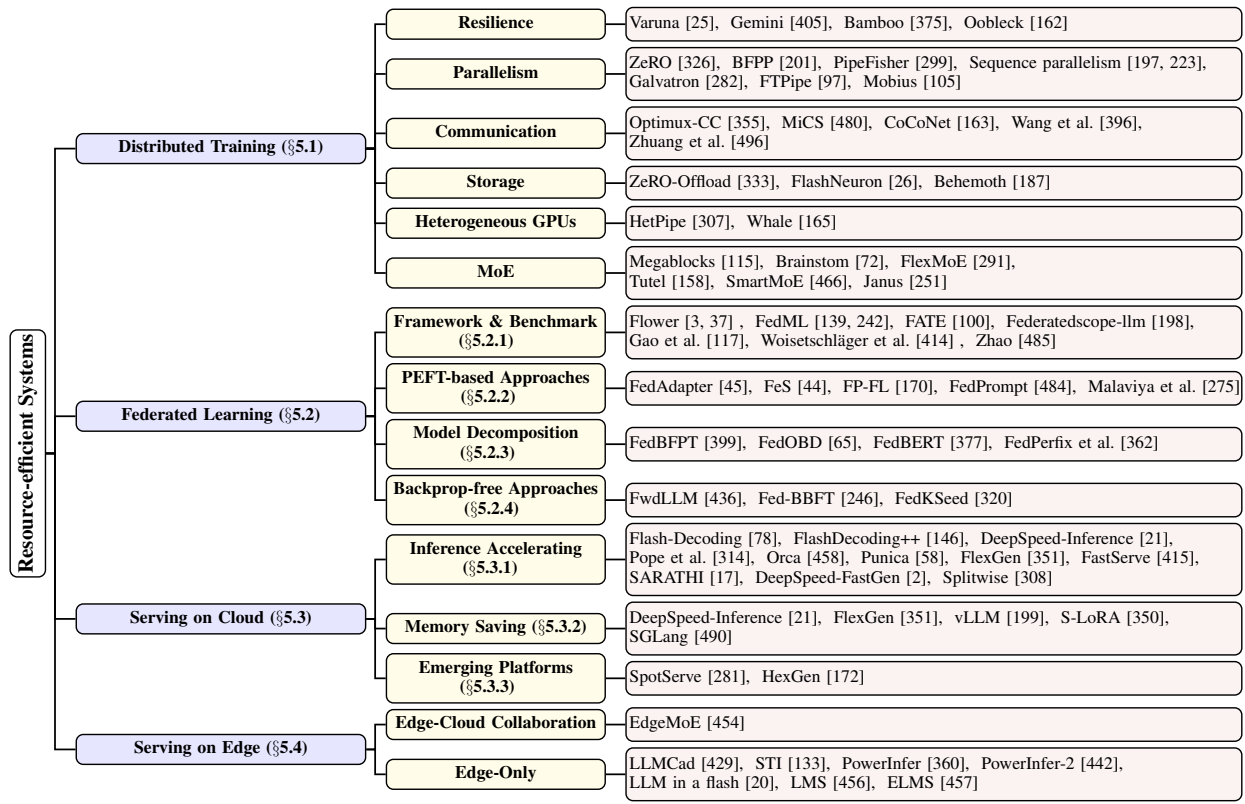

\begin{table*}[t]
\centering
\resizebox{\textwidth}{!}{%
\begin{tabular}{llrrrr}
\hline
\textbf{Frameworks} & \textbf{Descriptions}                                                                                                                                                                                                                                                                                                                                                              & \textbf{Cloud} & \textbf{Edge} & \textbf{Training} & \textbf{Inference} \\ \hline
DeepSpeed~\cite{deepspeed-fastgen}    & \begin{tabular}[c]{@{}l@{}} An open-sourced Python library proposed by Microsoft that supports MoE, long-sequence training, \\ RLHF, ZeRO optimizations, and model compression.  \\ It has already been used to train LLMs like MT-530B and BLOOM.\end{tabular} &   \checkmark    &      &    \checkmark      &     \checkmark      \\ \hline
Megatron-LM~\cite{narayanan2021efficient}    & \begin{tabular}[c]{@{}l@{}} The first cloud training system that introduces tensor parallelism to enable \\ distributed training of  language models, such as GPT, Bert, and T5. \\ It is  proposed by NVIDIA. \end{tabular} &   \checkmark    &      &     \checkmark     &           \\ \hline
Alpa~\cite{zheng2022alpa}   & \begin{tabular}[c]{@{}l@{}} A cloud FM training and serving engine that is proposed by UCB. \\ It supports automatic parallelization. \end{tabular} &   \checkmark    &      &     \checkmark     &     \checkmark      \\ \hline
FlexFlow~\cite{miao2023specinfer}  & \begin{tabular}[c]{@{}l@{}} A cloud FM training and serving compiler that is proposed by CMU and Stanford University. \\ It supports automatic parallelization in distributed. \end{tabular} &   \checkmark    &      &     \checkmark     &     \checkmark      \\ \hline
Colossal AI~\cite{colossalai}   & \begin{tabular}[c]{@{}l@{}} A cloud FM training and serving engine that is proposed by HPC-AI Tech, \\ which supports common parallelism strategies and heterogeneous memory management. \end{tabular} &   \checkmark    &      &     \checkmark     &     \checkmark      \\ \hline
FairScale~\cite{FairScale2021}  & \begin{tabular}[c]{@{}l@{}}  A PyTorch extension library that is proposed by Meta with new scaling techniques.\end{tabular} &   \checkmark    &      &     \checkmark     &     \checkmark      \\ \hline
PyTorch FSDP~\cite{zhao2023pytorch}   & \begin{tabular}[c]{@{}l@{}}  A cloud large-scale training system that is proposed by PyTorch. \\ It shards model parameters, optimizer states, and gradients.\end{tabular} &   \checkmark    &      &     \checkmark     &     \checkmark      \\ \hline
Huggingface PEFT~\cite{peft}    & \begin{tabular}[c]{@{}l@{}} An efficient fine-tuning systems that is proposed by Huggingface. \\ It supports a set of PEFT methods, such as LoRA, prefix tuning, and prompt tuning.\end{tabular} &   \checkmark     &      &    \checkmark (Fine-tune)      &           \\ \hline
OpenLLM~\cite{openllm}    & \begin{tabular}[c]{@{}l@{}} A open platform for LLMs in production proposed by BentoML. Support \\fine-tuning, serving, deploying, and monitoring any LLMs. \end{tabular} &   \checkmark     &      &    \checkmark (Fine-tune)      &     \checkmark      \\ \hline
    DeepSpeed MII~\cite{deepspeed-fastgen} & \begin{tabular}[c]{@{}l@{}} Open-sourced Python library proposed by DeepSpeed. Especially support \\DeepSpeed-FastGen. \end{tabular} & \checkmark & & & \checkmark  \\
    \hline
    vLLM~\cite{kwon2023efficient} & \begin{tabular}[c]{@{}l@{}} Popular serving engine proposed by UC Berkeley. The first one \\proposes PagedAttention. \end{tabular} & \checkmark & & & \checkmark \\
    \hline
    LightLLM~\cite{LightLLM} & \begin{tabular}[c]{@{}l@{}} An optimized serving framework that supports token-wise's KV cache memory management. \end{tabular} & \checkmark & & & \checkmark \\
    \hline
    LMDeploy~\cite{LMDeploy} & \begin{tabular}[c]{@{}l@{}} An optimized serving toolkit that is proposed by InternLM. \end{tabular} & \checkmark & & & \checkmark \\
    \hline
    Ray LLM~\cite{Ray-LLM} & \begin{tabular}[c]{@{}l@{}} An serving solution that is proposed by Anyscale. Support serving multiple LLMs. \end{tabular} & \checkmark & & & \checkmark \\
    \hline
    TensorRT-LLM~\cite{tensorrt-llm} & \begin{tabular}[c]{@{}l@{}} A TensorRT toolbox that is for optimized large language model inference. It supports AWQ, \\GPTQ, SmoothQuant quantization, speculative decoding, \\pipeline parallelism, tensor parallelism, PagedAttention. \end{tabular} & \checkmark & & & \checkmark \\
    \hline
    HuggingFace TGI~\cite{huggingface-tgi} & \begin{tabular}[c]{@{}l@{}} A high-performance serving engines that is developed and deployed in production at HuggingFace.\\ It supports tensor parallelism, quantization with bitsandbytes and GPT-Q, and PagedAttention.\end{tabular} & \checkmark & & & \checkmark \\
    \hline
MLC-LLM~\cite{mlc-llm}             &    \begin{tabular}[c]{@{}l@{}} A machine learning compilation system for large language  models (MLC LLM) \\ that is a high-performance universal deployment solution which allows native deployment \\ of any large language models with native APIs with compiler acceleration.  \end{tabular}                                                          &  \checkmark    &   \checkmark  &          &   \checkmark      \\ \hline
llama.cpp~\cite{llamacpp}           & \begin{tabular}[c]{@{}l@{}} A popular on-device LLM serving engine that supports mixed \\ F16 / F32 precision and 2-bit, 3-bit, 4-bit, 5-bit, 6-bit and 8-bit integer quantization. \\ This framework is mainly for LLaMA-based LLMs.\end{tabular}                                                                                                                        &    &    \checkmark   &        &    \checkmark     \\ \hline
mnn-llm~\cite{mnn-llm} & \begin{tabular}[c]{@{}l@{}} An edge large language models serving engine that was proposed by Alibaba and inherited from mnn. \\ It optimizes LLM inference by  dividing the inference procedure into  the prefill phase \\ and the decoding phase. During the prefill phase, it optimizes Matmul computation, \\ while it reduces memory access in the decoding phase.\end{tabular} &       &   \checkmark  &         &     \checkmark    \\ \hline

mllm~\cite{mllm} & \begin{tabular}[c]{@{}l@{}} A versatile and efficient inference engine for multimodal large language models on edge devices.\end{tabular}                                                                                                                        &    &    \checkmark   &        &    \checkmark     \\ \hline
Flower~\cite{beutel2020flower}             & \begin{tabular}[c]{@{}l@{}}A unified analytics and evaluation federated learning system  that is proposed and \\ updated by Cambridge. \\ It enables the federated fine-tuning of OpenAI's Whisper on Rasperberry Pi 5.\end{tabular} &       &   \checkmark  &  \checkmark   &         \\ \hline
FedML~\cite{he2020fedml}             & \begin{tabular}[c]{@{}l@{}} A federated learning framework that was initiated by USC. It released FedLLM version,\\ compatible with existing popular machine learning wheels such as DeepSpeed.\end{tabular} &       &   \checkmark  &  \checkmark   &         \\ \hline
FATE\cite{fan2023fate}         & \begin{tabular}[c]{@{}l@{}} The first industry-level federated learning framework proposed by WeBank,\\ that supports MPC encryption mechanism. It released \\FATE-LLM to promote the efficient FedLLM training using parameter-efficient tuning.\end{tabular} &       &   \checkmark  &  \checkmark   &         \\ \hline
\end{tabular}%
}
\caption{A summary of popular open-source tools for training and deploying large FMs.}
\label{tab:tab-open-source-FM-serving-tools}
\end{table*}

\subsection{Distributed Training}
\label{subsec:distributed-training-systems}
Distributed training systems serve as the foundation for training large FMs, encompassing pretraining and fine-tuning phases.
Pretraining, involving intensive computation and communication, demands substantial resources compared to other large FMs processes.
Fine-tuning is widely used to transform a general-purpose model into a specialized model for particular use cases.
Considering the large scale and new execution pattern of large FMs, designing resource-efficient systems for FMs has drawn great attention from the community.
We categorize techniques for optimizing distributed training systems, covering aspects such as resilience, parallelism, communication, storage, and heterogeneous GPUs. 
Additionally, MoE has emerged as a trend in training extremely large models, for which several approaches are tailored. These specialized methods are detailed at the end of this subsection.


\textbf{Resilience.}
The increasing size and duration of training for large FMs have led to a rise in failures, emphasizing the importance of resilient training~\cite{weng2022mlaas}. Fault tolerance approaches for large FMs primarily manifest in four forms.
First, Varuna and Gemini~\cite{athlur2022varuna, wang2023gemini} facilitate resilient training by implementing checkpoints to restart training. Varuna~\cite{athlur2022varuna} is designed for training in commodity clusters with low-bandwidth networks, frequent pre-emptions, and user-friendly features. On the other hand, Gemini~\cite{wang2023gemini} expedites failure recovery through in-memory checkpoints.
Second, Bamboo~\cite{thorpe2023bamboo} utilizes redundant computations where one node performs computations for both itself and its neighbors.
Bamboo avoids the overhead of recovering but introduces the overhead during training.
Third, activation checkpointing~\cite{zhang2022mics,korthikanti2023reducing}, which avoids storing the activation and recomputes it when needed, falls between the checkpointing and redundant computation approaches.
The fourth approach involves recovering partial layers, as demonstrated by Oobleck~\cite{jang2023oobleck}. In the event of a failure, the affected pipeline can be restored using partial layers from other replicas, incurring less overhead than employing the entire checkpoint.

\textbf{Parallelism.}
Parallelism plays a crucial role in distributed training, especially for large FMs.
Three types of parallelism are commonly employed for training large FMs.
\textit{Data parallelism} (DP) involves distributing the data across workers to scale up distributed training. DeepSpeed ZeRO~\cite{rajbhandari2020zero} optimizes memory usage by splitting the model states.
\textit{Model parallelism} (MP) partitions the model in intra-layer paradigm (Tensor parallelism~\cite{narayanan2021efficient}) or inter-layer paradigm (Pipeline parallelism~\cite{lamy2023breadth,osawa2023pipefisher}).
Tensor parallelism (TP) improves the training speed while leading to more communication.
Pipeline parallelism (PP) improves GPU utilization by filling the bubbles.
Breadth-first pipeline parallelism~\cite{lamy2023breadth} designs a looping placement and breadth-first schedule to achieve both high GPU utilization and low cost.
PipeFisher~\cite{osawa2023pipefisher} assigns extra work to the bubbles for further benefits.
Mobius~\cite{feng2023mobius} is designed for fine-tuning with a novel PP scheme and heterogeneous memory.
FTPipe~\cite{eliad2021fine} partitions the model into finer-grained blocks rather than layers for flexible execution and low resource demand.
\textit{Sequence parallelism}(SP)~\cite{li2023sequence,korthikanti2023reducing} is designed for the trend of long sequence training where training one sentence exceeds the memory capacity of one worker.
SP divides the long sequence into multiple chunks and puts them on different workers.
In practice, these parallelisms are usually used in a hybrid way.
Galvatron~\cite{miao2022galvatron} can automatically determine the most efficient hybrid parallelism strategy.

\textbf{Communication.}
The large scale and complex parallelism lead to significant communication overhead. We summarize the optimization of communication into two categories: reducing the communication time directly and hiding the communication.
Some work explores parallelism-aware communication compression~\cite{song2023optimus} and heterogeneity-aware traffic reduction~\cite{zhang2022mics}.
Existing work usually overlaps the communication with computation, by unifying the abstraction of computation and communication~\cite{jangda2022breaking}, decomposing the original communication collective~\cite{wang2022overlap}, or designing a novel pipelinging schedule~\cite{zhuang2023optimizing}.

\textbf{Storage.}
Large FMs require a significant amount of storage resources, e.g., GPU memory for model states, host memory for model analysis, and disk for dataset and checkpoint.
Various approaches have been proposed to alleviate the storage constraints for efficiency.
Offloading is a common way to reduce the stress of GPU memory.
ZeRO-Offload~\cite{ren2021zero} offloads data and computations to CPU to train large models on a single GPU.
FlashNeuron~\cite{bae2021flashneuron}, on the other hand, offloads selective data to the SSD for higher throughput.
Additionally, Behemoth~\cite{kim2021behemoth} replaces low-capacity, high-performance HBM with high-capacity, low-performance NAND flash to enable data-parallel training for large FMs.

\textbf{Heterogeneous GPUs.}
Training on specialized high-performance GPU clusters is impossible for most people or enterprises.
Moreover, heterogeneous GPUs commonly exist even in specialized GPU clusters.
Therefore, some efforts try to train large FMs on heterogeneous GPUs.
Hetpipe~\cite{park2020hetpipe} accelerates training with low-performance GPUs and Wave Synchronous Parallel to synchronize parameters among heterogeneous GPUs.
Whale~\cite{jia2022whale} introduces a hardware-aware load-balancing algorithm to speed up training.

\textbf{MoE.}
MoE is an efficient approach to scaling up DNN models.
The goals of optimizing MoE training systems are mainly efficiency and scalability.
Existing work mainly optimizes the dynamism-related mechanisms, parallelism, and communication in MoE training.
MegaBlocks~\cite{gale2023megablocks} leverages sparse primitives to handle dynamic routing and load-imbalanced computation.
Brainstorm~\cite{cui2023optimizing} is a framework for dynamic DNNs by abstracting the dynamism and profile-based optimization.
FlexMoE~\cite{nie2023flexmoe} focuses on the dynamic expert management and device placement problem.
Additionally, Tutel~\cite{hwang2023tutel} designs dynamic adaptive parallelism and pipelining strategies.
SmartMoE~\cite{zhai2023smartmoe} optimizes the parallelism strategy for efficient MoE training with a combination of offline and online mechanisms.
Janus~\cite{liu2023janus} changes communication from an expert-centric paradigm to a data-centric paradigm for faster communication in MoE training.

\subsection{Federated Learning}
\label{subsec:federated-fine-tuning-systems}

Data serves as the foundational fuel for large FMs. 
Federated learning (FL) has emerged as the predominant approach for training foundation models using data from multiple sources, ensuring data privacy in the process \cite{hard2018federated, yang2019federated, kairouz2021advances}. 
In light of this, significant efforts have been invested in the development of efficient federated learning systems specifically tailored for foundation models \cite{kang2023grounding, li2024synergizing}. 
Some recent works claim that the future of LLM pre-training lies in federated learning \cite{sani2024future, jiang2023fdapt}.
As depicted in Figure~\ref{fig:tree-resource-efficient-system-tree}, we classify the existing federated LLM systems into four primary categories: Framework \& Benchmark, PEFT-based Approaches, Model Decomposition, and Zeroth-Order Optimization.

\subsubsection{Frameworks \& Benchmarks}
\label{subsubsec:related-framework-benchmark}

Significant progress has been made in the development of federated learning frameworks and benchmarks for foundation models. 
These open-source federated learning frameworks are typically integrated with foundation models, offering a suite of APIs to facilitate a range of efficient federated learning algorithms. 
For instance, Flower \cite{flower2023whisper, beutel2020flower} supports federated fine-tuning of OpenAI's Whisper on Raspberry Pi 5. 
FedML \cite{he2020fedml, lin2022fednlp} introduces FedLLM, a version compatible with popular machine learning platforms such as DeepSpeed \cite{rasley2020deepspeed}. 
FATE \cite{fan2023fate} is designed to enhance the training efficiency of FedLLM using parameter-efficient fine-tuning methods. 
Federatedscope-llm \cite{kuang2023federatedscope} presents a comprehensive package for fine-tuning large FMs in federated settings, encompassing a broad spectrum of PEFT algorithms and various accelerating/resource-efficient operators.
FedIT \cite{zhang2023towards} develops a federated GPT through instruction tuning, streamlining the integration of new algorithms to capitalize on the diverse instructions available on edge devices. 
Gao et al. \cite{gao2022federated} investigated the integration of self-supervised learning with federated learning, focusing on speech representations using wav2vec 2.0 \cite{baevski2020wav2vec}. 
Woisetschl{\"a}ger et al. \cite{woisetschlager2023federated} evaluated the existing capabilities and potential of edge computing systems for FL with large FMs, outlining steps towards enhanced computational efficiency at the edge.
Zhao et al. \cite{zhao2023privacy} proposed the use of privacy-preserving technologies, including federated learning, differential privacy, and emulator-based tuning, in conjunction with parameter-efficient fine-tuning (PEFT) techniques. 
This approach aims to refine large FMs without compromising data privacy.

\subsubsection{PEFT-based Approaches}
\label{subsubsec:peft-based-methods}

Parameter-Efficient Fine-Tuning (PEFT) is a strategy designed to tailor large FMs for specific downstream tasks. This process involves freezing the backbone of large FMs and updating only a small set of additional parameters. PEFT aims to reduce both training time and communication costs, addressing a key challenge in FL. 
FedAdapter \cite{cai2023efficient} proposes progressively modifying the adapter configuration to efficiently identify the most effective setup.
FeS \cite{cai2023federated} integrates bias-only prompt learning with pseudo-labeling in a structured curriculum, achieving substantial accuracy with minimal data labeling.
FP-FL \cite{jiang2023low} introduces a soft-label enhanced federated tuning, incorporating LoRA tuning to diminish both computational and communication expenses.
FedPrompt \cite{zhao2023fedprompt} explores the additional advantages of prompt learning, particularly its potential to allow for larger differential privacy budgets.
FedPepTAO \cite{che2023federated} proposes an effective partial prompt tuning method with adaptive optimization, enhancing performance and efficiency in the presence of non-Independent and Identically Distributed (non-IID) data distributions.
Malaviya et al. \cite{malaviya2023reducing} analyzed the effectiveness of various PEFT methods under different non-IID scenarios and varying client fractions.
Petals \cite{borzunov2022petals} facilitates collaborative fine-tuning of large models, enabling multiple users to combine resources and apply parameter-efficient tuning methods such as adapters or prompt fine-tuning over the Internet. 

\subsubsection{Model Decomposition}
\label{subsubsec:model-decomposition}

Decomposing a large FM into several sub-models is a straightforward yet effective approach towards practical FL. 
FedBFPT \cite{wang2023fedbfpt} adopts a strategy where only a portion of BERT's layers are trained on the client side, with the number of layers involved gradually increasing. 
FedOBD \cite{chen2022fedobd} takes an innovative approach by decomposing large-scale models into semantic blocks, enabling FL participants to selectively upload quantized blocks to the FL server for aggregation.
FedBERT \cite{tian2022fedbert} introduces a federated split learning method. In this framework, the server updates the Transformer layer for all clients, while clients locally train the embedding and head layers.
FedPerfix \cite{sun2023fedperfix} explores the specifics of partial personalization in a ViT model. They conduct empirical evaluations to gauge the sensitivity of each layer type to data distribution. 
Drawing on the insight that the self-attention layer and the classification head are the most sensitive components of a ViT, FedPerfix uses plugins to facilitate the transfer of information from the aggregated model to individual clients for personalization.
SplitLoRA~\cite{lin2024splitlora} provides an open-source benchmark for split learning-based LLM parameter-efficient fine-tuning, providing a foundation for research efforts dedicated to advancing federated LLM fine-tuning.

\subsubsection{Backprop-free Approaches}
\label{subsubsec:zeroth-order-optimization}

While back-propagation (BP) remains the standard training paradigm for large FMs, its application in FL is challenging, primarily due to the high computational cost and memory overhead on edge devices. 
Zeroth-order optimization, an alternative optimization method, operates solely on function values and does not require explicit access to BP-based gradient information. In essence, it optimizes large FMs without relying on back-propagation.
FwdLLM \cite{xu2024fwdllm} is a pioneering work that integrates zeroth-order optimization into FL. This approach utilizes forward gradients as an unbiased gradient estimate. However, obtaining precise gradient estimations for each parameter necessitates substantial perturbations. 
To address this, FwdLLM incorporates PEFT methods to update only the intrinsic dimensions of the foundation model. FwdLLM facilitates federated training of billion-sized LLMs (like LLaMA) on standard mobile devices.
FedKSeed \cite{qin2023federated} introduces a novel strategy to circumvent forward gradient transmission by using a gradient accumulator. This method stages scalar gradient information and computes the latest gradient locally based on the initial model parameters. 
Fed-BBPT \cite{lin2023efficient} adopts a black-box tuning approach to generate superior instructions locally without the necessity of storing the entire pre-trained model.
BBTv2 \cite{sun2022bbtv2}, though not specifically designed for federated learning, presents a gradient-free, divide-and-conquer algorithm that decomposes foundation models into a lower-dimensional subspace.

\subsection{Serving on Cloud}
\label{subsec:cloud-serving-sys}

Large FMs have been widely adopted in many real-world applications, such as search engines, chatbots, and programming tools. 
As these applications have attracted a large number of users with an extraordinary amount of requests, the serving systems for FMs need to be highly efficient to satisfy user requirements.
To this end, many FM serving systems have been proposed to improve the serving efficiency, without sacrificing the accuracy of the FMs.
In this section, FM mainly refers to LLM.

Different from the FM training process, FM serving process exhibits an auto-regressive pattern. 
Specifically, when serving a request, the FM sequentially generates tokens one by one until the FM generates a special EOS (end-of-sequence) token. 
Each token generation iteration takes all preceding tokens, including input tokens and previously generated output tokens, as input.

To reduce the redundant computation between iterations, LightSeq~\cite{wang2020lightseq} proposes KV cache to cache the intermediate states of FMs for the redundant computation reduction.
Specifically, LightSeq caches key-value pairs of attention layers in the FM.
As a result, the generation process of the FM has two phases: prefill phase and decoding phase.
The prefill phase processes all input tokens of a request and stores the intermediate states in KV cache.
The subsequent decoding phase only needs to process the computation related to the newly generated tokens and update KV cache accordingly.
We categorize the existing optimizations for FM serving systems into three categories: computation optimizations, memory optimizations, and FM serving on emerging deployment platforms.

\subsubsection{Inference Accelerating}
\label{subsubsec:inference-accel}
To accelerate the computation in a single accelerator, kernel optimization is a common approach.
FlashAttention~\cite{dao2022flashattention} and FlashAttention-2~\cite{dao2023flashattention} design for FM training can be simply used to accelerate the prefill phase.
However, due to the unique characteristics of the decoding phase, Flash-Decoding~\cite{flashdecoding} proposes a specific NVIDIA CUDA kernel to accelerate the decoding phase.
FlashDecoding++~\cite{flashdecoding++} further improves the performance of Flash-Decoding by optimizing the softmax operation and flat GEMM operation in the decoding phase and provides additional AMD GPU support.
DeepSpeed-Inference~\cite{deepspeed-inference}, ByteTransformer~\cite{zhai2023bytetransformer}, and Google's PaLM serving system~\cite{pope2023efficiently} also optimize GPU/TPU optimizations for small batch size scenarios, which is common in FM serving but rare in FM training.

When scaling FM inference to numerous GPUs at a large scale, many works~\cite{deepspeed-inference,pope2023efficiently} exploit combinations of various parallelism strategies, such as data parallelism, pipeline parallelism, tensor parallelism, and expert parallelism
These works efficiently serve FM inference on multiple modern accelerators, such as GPUs/TPUs.

Request batching and scheduling constitute another set of methods to enhance the computational efficiency of request processing. Given the auto-regressive nature of FMs, various requests may feature distinct lengths of input tokens and output tokens. Merely batching requests with different lengths together by padding them to the same length can result in the FM serving system expending computation resources on the padded tokens. 
To address this issue, Orca~\cite{yu2022orca} proposes selective batching and iteration-level scheduling to batch requests of different lengths at the granularity of iterations to increase the maximum batch size. 
With these techniques, early finished requests can return to users without waiting for the late finished requests, and newly arrived requests can be executed as soon as one iteration is finished.
Many works also improve the scheduling of request batching. Punica~\cite{chen2023punica} further proposes a grouped GEMM kernel to batch requests destined for different LoRA models. 
FlexGen~\cite{sheng2023flexgen} proposes a request scheduling algorithm to mitigate the impact of offloading on the performance of latency-insensitive FM serving in a single GPU. 
FastServe~\cite{wu2023fast} proposes an iteration-level preemptive scheduling and proactive KV cache swapping to mitigate the impact of head-of-line blocking on the performance of distributed FM serving. 
SARATHI~\cite{agrawal2023sarathi} and DeepSpeed-FastGen~\cite{deepspeed-fastgen} split the computation of the prefill phase into small chunks and schedule these chunks with the decoding phase to mitigate the impact of the prefill phase on the performance of large FMs serving. 
Splitwise~\cite{patel2023splitwise} splits the prefill phase and the decoding phase onto different machines according to their different computation and memory requirements. 

\subsubsection{Memory Saving}
\label{subsubsec:mem-saving}
A FM consumes a large amount of memory during the serving process. 
To reduce the memory consumption of FM serving, many works propose various memory management techniques. 
As for FMs's parameters and activations, DeepSpeed-Inference~\cite{deepspeed-inference} and FlexGen~\cite{sheng2023flexgen} offload activations or model parameters to the DRAM or NVMe memories when the GPU memory is insufficient.

KV cache is another important memory component in FM serving. 
To reduce the memory consumption of KV cache, vLLM~\cite{kwon2023efficient} adopts a block-level on-demand memory allocation mechanism, which only allocates memory to intermediate states when needed.
vLLM also proposes a new operator, Paged Attention, to support attention operation when using this memory allocation mechanism. 
S-LoRA~\cite{sheng2023s} extends this idea to Unified Paging to manage multiple LoRA adapters at the same time. 
SGLang~\cite{zheng2023efficiently} further exposes prompt programming primitives to users to enable more complex KV cache management among all requests with the help of RadixAttention.

\subsubsection{Emerging Platforms}
\label{subsubsec:emerging-platforms}
Typical FM serving systems are usually deployed on data centers equipped with plenty of homogeneous high-performance servers. 
Due to the scarcity and cost of these high-performance servers, there are also some FM serving systems specifically designed for other deployment platforms. 
SpotServe~\cite{miao2023spotserve} tries to serve FMs on spot instances, which are low-cost but unreliable cloud instances. 
SpotServe dynamically adjusts its parallelism strategy to accommodate the impact of spot instance preemption.
As for FM serving on heterogeneous GPUs, HexGen~\cite{jiang2023hexgen} uses an evolutionary algorithm to search for high-performance FM placement and parallelism strategy on heterogeneous GPUs. 

\subsection{Serving on Edge}
\label{subsec:edge-serving-sys}
With ever-increasing data privacy concerns and the stringent response latency requirement, running large FM on mobile devices locally (i.e., on-device inference) has recently attracted attention from both academia and industry.
Thereby, many on-device inference optimization techniques have been introduced. 

\noindent \textbf{Edge-cloud collaboration.}
A common strategy to tackle the scarce resources on mobile devices is to speed up the intensive inference with a powerful edge/cloud server collaboration.
For instance, EdgeFM~\cite{yang2023edgefm} queries and adapts the large FMs to the specific edge models with customized knowledge and architectures so that the dynamic edge model can ensure both low latency and close accuracy to the original large FMs.

\noindent \textbf{Edge-only.} 
\update{
Another main research direction is to directly optimize on-device inference.
There is a recent study focusing on on-device language models~\cite{xu2024device}.
We summarize primary techniques in the following.
}

\begin{itemize}[leftmargin=*]
    \item \textit{On-device MoE} models are proposed to only execute in routed sparse parameters during inference, which can decrease computation (detailed in $\S$\ref{subsec:dynamic-neural-network}).
    EdgeMoe~\cite{yi2023edgemoe} identifies the problem that experts have to be dynamically loaded into memory during inference. 
    To tackle this issue, this approach proposes expert-wise bit-width adaptation to reduce the size of expert parameters with acceptable accuracy loss, saving parameters loading time.
    The expert management can predict which expert shall be activated in the future so that EdgeMoe can preload it to reduce I/O overhead.
    PC-MoE~\cite{kong2023serving} is based on a crucial observation that expert activations are subject to temporal locality.
    Based on this observation, PC-MoE proposes Parameter Committee, which intelligently maintains a subset of crucial experts in use to reduce resource consumption.
    
\item \textit{Memory optimization.}
Since large FMs often rely on large parameter sizes and on-device memory resources are scarce (e.g., 8GB), inferring large FMs on devices face the challenge of ``memory wall''.
To tackle this issue, LLMCad~\cite{xu2023llmcad} utilizes speculative decoding~\cite{Leviathan2022FastIF} which can offload most workloads to a smaller memory-resident draft model.
LLMCad further proposes token tree generation and verification, a self-adaptive fallback strategy, and a speculative generation pipeline to reduce verification times and exploit idle processors during the verification process to enhance generation speed.
PowerInfer~\cite{song2023powerinfer} relies on large FMs runtime sparsity, i.e., only hot neurons are consistently activated across inputs.
To that end, PowerInfer preloads hot-activated neurons onto the GPU for fast access, while cold-activated neurons are computed on the CPU, thus significantly reducing GPU memory demands and CPU-GPU data transfers.
\update{
PowerInfer-2~\cite{xue2024powerinfer} focuses on large FMs on smartphones.
It leverages neuron cluster-based computations and segmented neuron caching, with hot neurons preloaded into memory for fast access, while cold neurons are loaded on-demand, therefore reducing memory usage and minimizing I/O bottlenecks.
}

\item \textit{I/O optimization.}
As parameter size increasing speed is larger than edge devices' memory increasing speed, dynamically loading parameters from disks to memory is avoidable.
STI~\cite{guo2023sti} identifies that loading parameters time is highly longer than computation time.
To address this problem, STI proposes dynamically adapting weights bit-width during the loading procedure according to parameters importance, minimizing loading overhead under maximum inference accuracy. Large FMs in a flash~\cite{alizadeh2023llm} solves this problem by fine-grained management of flash storage to reduce the volume of data transferred from flash to memory as well as reading data in larger, more contiguous chunks. 
    
\item \textit{Kernel optimization.}
Computing resources are also crucial while limited resources on the devices.
Prior study~\cite{zhang2023practical} implements the first 32-bit integer-based edge kernel for vision transformers with post-training integer-only quantization to speed up the inference process.
This method also introduces a range-constrained quantization technique for activation and normalization operators in transformers to trade off data range and inference accuracy.
mllm-NPU~\cite{xu2024empowering} enables fast LLM prefilling by leveraging the mobile NPUs.
It tackles a few inherent semantic gaps between LLM inference and NPU design.

\update{
\item \textit{LLM as a system service.}
Recently there is a trend of deploying a single yet powerful LLM (or foundation model) into the OS of mobile devices, namely LLM-as-a-Service (LLMaaS).
This paradigm is entirely viable due to the world knowledge and in-context learning capabilities of LLMs.
Since just one instance of the LLM weights is required in memory, it prevents the device's memory from being depleted by app-specific models.
With a unified model architecture and operator set, the LLM service can take greater advantage of system-level scheduling enhancements (such as batching or priority queues) and hardware accelerators (like GPUs, DSPs, or NPUs).
An industry example is Google’s Android AICore, an on-device LLM service integrated within the Android OS, utilized by apps such as GBoard smart reply and the Pixel voice recorder. Comparable paradigms can also be observed in offerings from Apple and Intel.
LMS~\cite{yin2024llmservicemobiledevices} presents such an LLMaaS on mobile devices with concrete system designs like app-level APIs and stateful invocation methods.
ELMS~\cite{yin2024elmselasticizedlargelanguage} further identifies a key challenge in this paradigm.
It elasticizes the single LLM to meet diverse SLOs that are required by third-party apps.
}
\end{itemize}


    

	\section{CONCLUSIONS AND FUTURE DIRECTIONS}\label{sec:conclusions}

This survey provides a holistic, systematic overview of recent literature towards resource-efficient large foundation models.
We first present the preliminary background and cost analysis of the popular foundation models, including large, vision, and multimodal.
We then dive into the model architecture, algorithm, and system designs to enable more resource-efficient large foundation model lifecycle.
In the future, the research of this domain will continue to be (or even more) crucial since the scaling law guarantees a promising future of more powerful AI with larger and larger models.
Such research is also highly interdisciplinary, involving various CS communities such as machine learning, NLP/CV/Speech, networking, cloud computing, edge computing, etc.

The research opportunity of resource-efficient large foundation model is extremely large, notably:

(1) \textbf{Cloud-edge hybrid deployment.} To enable ubiquitous, privacy-preserving, and highly available general intelligence, many foundation models will ultimately sink to near-user devices~\cite{xu2018deepcache,xu2019first,xu2018deeptype,zhang2022comprehensive}.
Preliminary efforts have been already conducted to bring LLaMA-7B to smartphones and PCs.
The killer applications include personal assistants/agents~\cite{li2024personal_llm_agents,wen2023autodroid}, multimodal information retrieval~\cite{li2023multimodal}, etc.
In the future, at what size and speed the foundation models can run on devices will become a key competitive force in the business model of hardware vendors.

(2) \textbf{Exploiting the model sparsity.}
With model being larger, the activated ratio of model will go smaller for a given task.
Recent literature~\cite{liu2023deja} finds that even a densely trained non-MoE model exhibits runtime activation sparsity, which can be exploited to reduce inference time and memory footprint.
We believe that exploiting the model and activation sparsity will be a promising direction towards sustainable model size scaling.
More efficient sparse architectures other than MoE could emerge.

(3) \textbf{Large foundation model as a service.}
On both clouds and devices, large foundation models are unifying the DNN ecosystem~\cite{yuan2023rethinking}.
Ultimately, it becomes a universal service to be invoked just as today's Web and Database.
On the one hand, it opens the opportunity for highly hardware-algorithm co-design and optimizations;
meanwhile, it poses new challenges in system and infrastructure design for scheduling, load balancing, and security\&isolation.

(4) \textbf{Agent as a holistic system to optimize.}
In the future, foundation models especially LLMs will be used as a key building block for establishing agents~\cite{li2024personal_llm_agents,wen2023autodroid}.
Its efficiency shall not be considered as in a standalone LLM service; instead, the algorithm and system designs need to cater to the specific agent workflow.
For example, an agent system might require multiple foundation models to cooperate, where there exists inherent logic dependency.
In this process, the design space of selecting the proper foundation models for each task and scheduling them on a given set of hardware resources to maximize the agent performance is huge.

(5) \textbf{Practical privacy-preserving FM.}
As the volume of user data uploaded to the cloud for FM processing continues to increase, the severity of privacy concerns correspondingly escalates.
Existing methods include federated learning, homomorphic encryption, and disentanglement learning.
While being theoretically sound, those methods still confront significant performance challenges, hindering their large-scale in-the-wild deployment.
A promising direction involves the development of innovative privacy-preserving techniques specifically designed for large FMs, or the refinement of existing methods, to effectively balance privacy with performance.

(6) \textbf{Understanding the scaling law.}
The scaling law drives the success of large FMs, while it also seems like a foundamental limitation for lightweight FMs - small-scale models are unlikely to possess higher levels of intelligence than larger models.
Understanding the mechanisms and theories behind the scaling law would be beneficial to explain (and hopefully break) the limitation.
Meanwhile, designing novel model architectures with better or even optimal scaling performance would be a direction that deserves extensive investigation.

	\bibliographystyle{plain}
	\bibliography{ref-mwx, ref-cdq, ref-wqp, ref-zyh, ref-yws, ref-yc, ref-yrj, ref-wby, ref-xdl, ref-wsh, ref-lyc, ref-lzy, ref-zl}
\end{document}